\pdfoutput=1

\documentclass[11pt]{article}

\usepackage{acl}

\usepackage{times}
\usepackage{latexsym}

\usepackage[T1]{fontenc}

\usepackage[utf8]{inputenc}

\usepackage{microtype}

\usepackage{inconsolata}

\usepackage{graphicx}
\graphicspath{ {./images/} }
\usepackage{amsmath,amssymb}
\usepackage{multirow}
\usepackage{caption}
\usepackage{subcaption}
\usepackage{algorithm} 
\usepackage[noend]{algpseudocode}
\usepackage{arydshln}
\usepackage{booktabs}
\usepackage{multirow}
\usepackage{fdsymbol}
\usepackage{siunitx}
\usepackage{dirtytalk}

\usepackage{bm}

\def\eqref#1{equation~\ref{#1}}

\def\1{\bm{1}}

\def\vb{{\bm{b}}}

\def\vg{{\bm{g}}}
\def\vh{{\bm{h}}}

\def\vv{{\bm{v}}}

\def\vz{{\bm{z}}}

\def\mW{{\bm{W}}}

\DeclareMathAlphabet{\mathsfit}{\encodingdefault}{\sfdefault}{m}{sl}
\SetMathAlphabet{\mathsfit}{bold}{\encodingdefault}{\sfdefault}{bx}{n}

\usepackage{array}
\usepackage{xspace}

\newcolumntype{L}[1]{>{\raggedright\let\newline\\\arraybackslash\hspace{0pt}}m{#1}}
\newcolumntype{C}[1]{>{\centering\let\newline\\\arraybackslash\hspace{0pt}}m{#1}}
\newcolumntype{R}[1]{>{\raggedleft\let\newline\\\arraybackslash\hspace{0pt}}m{#1}}

\makeatletter
\def\adl@drawiv#1#2#3{%
        \hskip.5\tabcolsep
        \xleaders#3{#2.5\@tempdimb #1{1}#2.5\@tempdimb}%
                #2\z@ plus1fil minus1fil\relax
        \hskip.5\tabcolsep}
\newcommand{\cdashlinelr}[1]{%
  \noalign{\vskip\aboverulesep
           \global\let\@dashdrawstore\adl@draw
           \global\let\adl@draw\adl@drawiv}
  \cdashline{#1}
  \noalign{\global\let\adl@draw\@dashdrawstore
           \vskip\belowrulesep}}
\makeatother

\makeatletter
\DeclareRobustCommand\onedot{\futurelet\@let@token\@onedot}
\def\@onedot{\ifx\@let@token.\else.\null\fi\xspace}

\def\eg{e.g\onedot,\xspace} 

\makeatother

\newcommand{\modelours}{\textsc{ConGater}\xspace}

\newcommand{\ttsig}{\texttt{t-sigmoid}\xspace}

\newcommand{\modelbert}{\textsc{BERT-Base}\xspace}
\newcommand{\modelmini}{\textsc{BERT-Mini}\xspace}
\newcommand{\modelroberta}{\textsc{Roberta-Base}\xspace}
\newcommand{\modelfine}{\textsc{Ft}\xspace}
\newcommand{\modelfinedeb}{\textsc{FtAdv}\xspace}

\newcommand{\modeladapter}{\textsc{Adp}\xspace}
\newcommand{\modeladapterdeb}{\textsc{AdpAdv}\xspace}

\title{Effective Controllable Bias Mitigation for Classification and Retrieval using Gate Adapters}

\author{Shahed Masoudian$^{1,2}$~~~Cornelia Volaucnik$^1$~~~Markus Schedl$^{1,2}$~~~Navid Rekabsaz$^{1,2}$\\
  $^1$Johannes Kepler University Linz, Institute of Computational Perception, Austria\\
  $^2$Linz Institute of Technology, AI Lab, Austria\\
\texttt{\{shahed.masoudian,markus.schedl,navid.rekabsaz\}@jku.at}\\
}

\begin{document}
\maketitle

\begin{abstract}

Bias mitigation of Language Models has been the topic of many studies with a recent focus on learning separate modules like adapters for on-demand debiasing. Besides optimizing for a modularized debiased model, it is often critical in practice to control the degree of bias reduction at inference time, e.g., in order to tune for a desired performance-fairness trade-off in search results or to control the strength of debiasing in classification tasks. In this paper, we introduce \emph{Controllable Gate Adapter (\modelours)}, a novel modular gating mechanism with adjustable sensitivity parameters,
which allows for a gradual transition from the biased state of the model to the fully debiased version at inference time. We demonstrate \modelours performance by (1) conducting adversarial debiasing experiments with three different models on three classification tasks with four protected attributes, and (2) reducing the bias of search results through fairness list-wise regularization to enable adjusting a trade-off between performance and fairness metrics. Our experiments on the classification tasks show that compared to baselines of the same caliber, \modelours can maintain higher task performance while containing less information regarding the attributes. Our results on the retrieval task show that the fully debiased \modelours can achieve the same fairness performance while maintaining more than twice as high task performance than recent strong baselines. Overall, besides strong performance \modelours enables the continuous transitioning between biased and debiased states of models, enhancing personalization of use and interpretability through controllability. \footnote{training and evaluation code for our experiments is available at \href{https://github.com/ShawMask/DebiasingConGater}{https://github.com/ShawMask/DebiasingConGater}}

\end{abstract}

Pre-trained Language models (LMs) have shown impressive ability in learning effective representations and diverse aspects of language, including harmful biases and stereotypes~\cite{zhao2019gender,sheng2019woman,blodgett2020language,rekabsaz_2020_neural,stanovsky2019evaluating}. A common bias mitigation category, referred to as representational fairness~\cite{elazar_2018_adv}, aims at minimizing the information regarding a specific attribute to make the models' decision blind to the attribute. 
This is realized in classification scenarios to make the model invariant to given protected attributes, and also in information retrieval (IR) tasks to opt for the neutrality of search results. A common in-processing approach to mitigate biases is to extend model optimization with various bias mitigation criteria (e.g., adversarial optimization or regularization) and update the whole model's parameters to a debiased state~\cite{elazar_2018_adv,colombo_2021_novel,zerveas_2022_mitigating}. New studies focus on modularizing this process by introducing new modules such as adapters~\cite{pfeiffer2021adapterfusion,houlsby_2019_param_efficient}, and sparse masking networks~\cite{zhang_2021_disentangling,hauzenberger2023parameter,zhao_2020_masking}. 

Besides effectiveness in reducing bias, it is often important in practice to be able to control the degree of imposing the debiasing criteria at inference time. This is beneficial particularly to apply possible fairness-performance trade-offs, specific preferences of each user, or the particular needs in processing each given input.\footnote{Regarding the last point, see for instance \citet{krieg2022grep} and \citet{hauzenberger2023parameter} about the need to control for the gender information in the processing of bias-sensitive inputs (like \emph{how to become CEO?}) versus the ``normal'' ones (like \emph{earliest pregnancy symptoms}).} Debiasing controllability enables to set the desired degree of a bias constraint's contribution at inference time, while 
in the current paradigm, one needs to train and deploy multiple parallel models or modules with various mitigation degrees~\cite{kumar2023parameter,hauzenberger2023parameter,zerveas_2022_mitigating}, imposing an untenable burden in practice.

In this paper, we address debiasing controllability by introducing \emph{Controllable Gate Adapter} (\modelours). The proposed module is based on a novel gating mechanism, that learns to reduce protected attribute information from the embedding while allowing information necessary for the task to pass through the model. The \modelours is equipped with a novel activation function \emph{Trajectory Sigmoid (\ttsig)}, used to form the gate vectors. \modelours is agnostic to debiasing optimization and can be trained with any gradient descent-based signal that removes attributes or increases fairness. During training, \ttsig has the same shape as a (standard) sigmoid function. At inference time, however, the form of \ttsig can flatten by decreasing the sensitivity parameter, transitioning from the sigmoid function (full gate intervention) to the constant function (no influence) creating a nonlinear interpolation effect. This transition can be viewed as traversing the trajectory of embeddings from the state of the original (biased) model to its fully debiased version, resulting in adjustable attribute removal qualities in the model's outputs and internal embeddings (\S\ref{sec:method}).

We demonstrate the functionality of \modelours by doing two sets of experiments: (1) adversarial bias mitigation with three models on three real-world classification datasets namely, occupation prediction from biographies with gender as protected attribute~\cite{de:bios}, hate speech detection with dialect-based race as protected attribute~\cite{founta:hatespeech}, and mention prediction with two attributes: gender and age of authors~\cite{rangel:pan16} totaling 12 experiment setups. In this experiment, we show that \modelours can reduce information about the attribute in the embeddings better than baselines while mostly preserving task performance. We also show that attribute information reduction in the embeddings of the model is continuous. we also demonstrate that \modelours continuous attribute control results in higher interpretability through controllability of model behavior at inference time. (2) Fairness/Neutrality of search results with gender as a protected attribute. We conduct the experiments on a recent IR benchmark~\cite{rekabsaz_2021_societal,msmarco}, optimizing \modelours with a recently-introduced list-wise neutrality regularization term~\cite{zerveas_2022_mitigating}. We demonstrate that the fully debiased \modelours can preserve task performance more than twice as high as the baselines with the same fairness performance, and \modelours is able to control the trade-off between the biased and debiased model continuously and linearly (details in \S\ref{sec:results:experiments} and \S\ref{sec:results}).

In Summary, our contributions are as follows: 
\begin{itemize}
    \item We introduce a novel gating mechanism (\modelours) with controllable debiasing ability at inference time.
    \item Through 13 different experimental setups we demonstrate that fully active \modelours enhances attribute information removal from the embeddings while having low task performance loss. 
    \item  We also show that continuous attribute control allows users to observe and change model behavior, resulting in increased interpretability (classification scenario) and flexibility (fairness of search results).
\end{itemize}

\section{Related Work}
\label{sec:related}

\emph{Efficient modular training} introduces an alternative to fine-tuning, where a (small) network is trained for a specific objective while the core model's parameters remain unchanged~\cite{pfeiffer2023modular}. Adapters realize modular training with a non-linear feed-forward network added to each layer~\cite{rebuffi2017learning,houlsby_2019_param_efficient,pmlr-v97-stickland19a} of transformers. Several works study the various aspects of adapters, such as parameter efficiency~\cite{ruckle2021adapterdrop,han2021robust}, architectural variations~\cite{DBLP:conf/nips/MahabadiHR21}, and transfer learning capacity~\cite{pfeiffer2021adapterfusion}. Recently, \citet{Lian:SST} show that scale and shifting of embedding is sufficient for effectively learning the task.

\emph{Bias mitigation.} Mitigating societal bias in LMs is explored particularly in the context of \emph{attribute erasure}. The aim of this task is to reduce the encoded information of a specific attribute from the latent embeddings and is particularly utilized in the context of mitigating empirical societal biases in LMs~\cite{mehrabi-bias, shen-representational}. Bias mitigation is approached by methods such as linearly projecting embeddings into the space with minimum correlations to protected attributes~\cite{ravfogel_2020_null,kaneko_2021_debiasing}, to achieve empirical fairness, through using a distribution alignment loss~\cite{guo_2022_auto}, or by applying adversarial training to learn representations agnostic to protected attributes~\cite{elazar_2018_adv, barrett_2019_adversarial,han_2021_diverse,wang_2021_dynamically,rekabsaz_2021_societal,ganhoer2022mitigating}.

\emph{Controllability.} Decoder LMs have been extensively studied for controllability. Researchers mostly cover tasks such as attribute manipulation (like positive/negative sentiment), and imposing predefined syntactic/semantic structure to text generation~\cite{zhang2022survey,st:alexis:semantic-control,kumar-etal-2022-gradient,qin2022cold,st:shen:educating-autoencoder}. As examples, \citet{st:nishant:steering-vector} introduces steering vectors as a way of changing the semantics of text generation, while \citet{yu2021attribute} learn an alignment function to force the text generation in the direction of a specific target concept. 
More recently, \citet{decoder:hallinan:detoxifying} used likelihood between expert and anti-expert models to detoxify text generation. 

\emph{Information Retrieval.} In IR tasks, many bias mitigation methods have been proposed. These methods mostly use list-wise optimization~\cite{oosterhuis:ir:fairness,morik:ir:fairness}. \citet{rekabsaz_2021_societal} proposed integrating adversarial training in deep-ranking models to improve bias mitigation. \citet{zerveas_2022_mitigating} introduced bias-aware optimization method using CODER~\cite{zerveas:coder} with TAS-B~\cite{hofstatter:ir:fairness}. We will use their method as the training strategy for our IR fairness task~\ref{sec:method:training}. 

Few recent studies explore modularized adversarial bias mitigation of encoder language models. \citet{lauscher_2021_sustainable} use a stack of adapters, while \citet{kumar2023parameter} first learns separate adapters for tasks and debiasing attributes and then combine them on-demand using the fusion network. Utilizing masking methods, \citet{zhang_2021_disentangling} learns binary masks applied to the initial network to erase the concept of interest, and \citet{hauzenberger2023parameter} train sparse weight-difference subnetworks, one for each attribute, which can be added to the core model on-demand. Our work extends this line of research by introducing a modularized, graded (non-binary), and controllable approach evaluated on continuous concept erasure.

Finally, the \emph{gating mechanism} has been used in various architectures to learn via scaling. As examples, \citet{ramachandran:swish} propose the self-gate activation function with trainable parameters, while \citet{papernot:tempered_sigmoid} introduce the tempered \textit{sigmoid} activation with a bias and scaling factor.~\citet{hu:squeeze} introduced Squeeze and Excitation networks which use bottle-neck networks added after each convolutional layer followed by a sigmoid activation function. Compared to Squeeze and excitation which captures global attention to the channels, our model uses a second training signal to isolate and filter out protected attributes. Another difference between our method and Squeeze and Excitation is the usage of a new activation function instead of a sigmoid activation function which gives us the benefit of controllability at inference.

\begin{figure}[t]
\center
\begin{subfigure}{0.45\textwidth}
\centering
    \includegraphics[width=0.7\columnwidth]{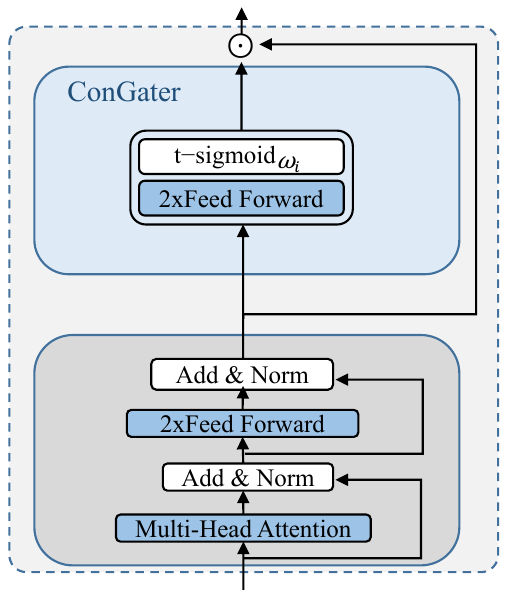} 
    \caption{\modelours}
    \label{fig:congator}
\end{subfigure}
\begin{subfigure}{0.45\textwidth}
\center
\includegraphics[width=0.7\columnwidth,]{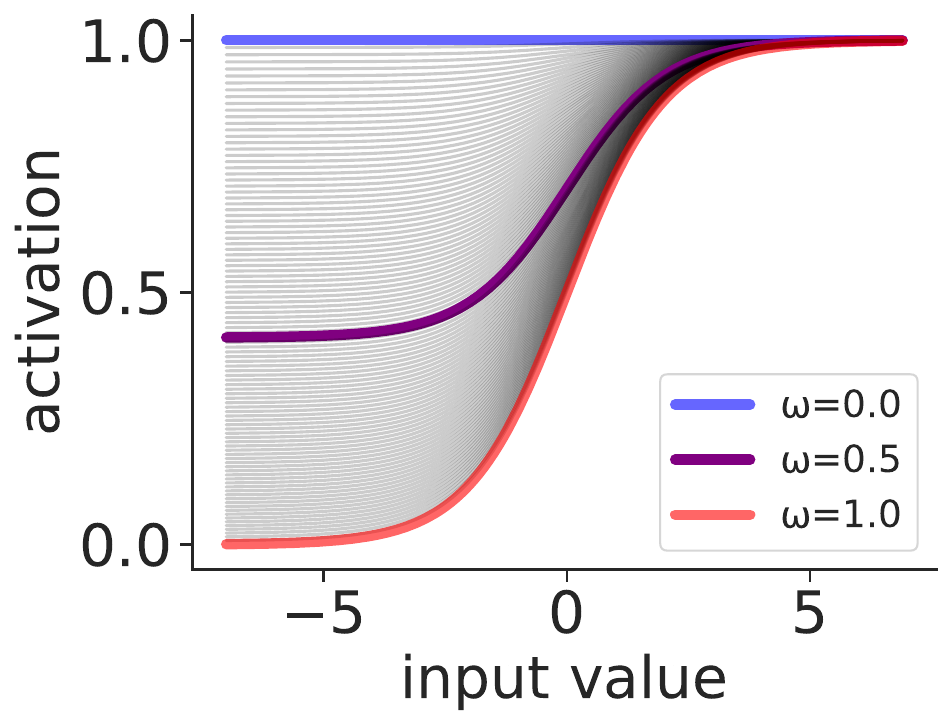}
\caption{\ttsig}
\label{fig:ttg}
\end{subfigure}
\caption{
(a) The overall architecture of \modelours as an adjustable self-gate adapter network. (b) Effect of $\omega$ parameter on \ttsig. Increasing $\omega$ results in a transition from the constant function $y=1$ (open gate) to the sigmoid function (full functional gate).
}
\vspace{-4mm}
\end{figure}

\section{Model and Training}
\label{sec:method}

In this section, we first introduce the proposed \emph{Controllable Gate Adapter (\modelours)} and \emph{trajectory-sigmoid (\ttsig)} activation function. We then explain the parallel and post-hoc training regimes, and how the gating sensitivity parameter of \ttsig can be adjusted at inference time to control the effectiveness of the gates.

\subsection{\modelours Architecture}
The \modelours module follows the principle of adapters~\cite{houlsby_2019_param_efficient} by dedicating a small network, added after each transformer block~\cite{vaswani_attention_NIPS2017} of an LM. Figure~\ref{fig:congator} depicts the architecture of a \modelours module inside a transformer block, responsible for controlling one attribute. In short, \modelours applies a gating mechanism for each layer, where the gate vector is defined via bottleneck followed by the \ttsig activation function. Concretely, for the $i$\textsuperscript{th} target attribute, we first define the gate vector $\vg_{i}$ formulated as: 
\begin{equation}
\vg_{i} = \text{\ttsig}_{\omega_i}(\vv_{i})
\label{eq:gate_vectors}
\end{equation}
\begin{equation}
    \vv_{i} = \mW^{2}_{i} \tanh(\mW^{1}_{i} \vh + \vb^{1}_{i}) + \vb^{2}_{i}
\end{equation}

where $\vh$ is the input vector (output of the transformer block), and $\mW_{i}$ and $\vb_i$ are weight and bias parameters, respectively. \ttsig is a generalized form of the sigmoid function, enhanced with the gating sensitivity variable $\omega_i$. This gating sensitivity parameter can be set to values in the range of $[0,1]$, which changes the shape of \ttsig, as illustrated for several values of $\omega$ in Figure~\ref{fig:ttg}. In particular when $\omega_i=1$, \ttsig is equivalent to the sigmoid function $\sigma$ and hence: $\vg_{i} = \sigma(\vv_{i})$. On the other end, setting $\omega_i=0$ changes \ttsig to the constant function $y=1$ resulting in $\vg_{i}=1$. Regardless of the $\omega_i$'s value, the output of \ttsig and hence each value of gating vector $\vg_{i}$ is bounded to $[0,1]$, indicating the range of the gate mechanism from fully closed to the fully open (more details below).

The transformation of the \modelours module is defined as the self-gate of the input using element-wise multiplication, defined below:
\begin{equation}
\label{eq:CRAG}
    output = \vh\odot\vg_{i}
\end{equation}

This transformation downscales each value of $\vh$ by its corresponding gate value, except for the cases with a corresponding open gate (gating value of~1), to which no change is made. Overall, a \modelours has the same number of parameters as a standard adapter network~\cite{pfeiffer2021adapterfusion} with the negligible computation overheads of Eq.~\ref{eq:gate_vectors} and~\ref{eq:CRAG}.

We now formulate the \ttsig activation function and discuss how it can be used to control the behavior of the model. The \ttsig function is formulated below:
\begin{equation}
    \ttsig_{\omega}(x) = 1-\frac{\log_2{(\omega + 1)}}{1+e^{x}},\quad \omega \in [0,1]
\label{eq:ttsigmoid}
\end{equation}

The gate sensitivity parameter $\omega$ is not trainable and can be set manually to change the shape of the activation function. 
If $\omega=0$, the \ttsig becomes the constant function $\ttsig(\vv_{i})~=~1$, meaning that the whole \modelours module turns into an identity function that simply outputs the given input. By increasing the value of $\omega$, \ttsig gradually transforms to the shape of a sigmoid function. The gradual transformation of \ttsig has the following characteristics: (1) For a specific value of $\omega$, the output of \ttsig monotonically increases with increasing input value $x$; (2) Given $\omega_2 > \omega_1$, the resulting outputs of the same input value $x$ is $\ttsig_{\omega_2}(x)\le\ttsig_{\omega_1}(x)$ (stronger gate); (3) Throughout the spectrum of $\omega$, the shape of \ttsig gradually changes, avoiding drastic alterations. These characteristics allow a smooth change in the effect of the gating mechanism, nonlinear interpolation, and hence continuous controllability of the information flow for the respective attribute.\footnote{While we focus on removing one attribute with \modelours, this definition can be extended to multiple attributes. We propose one possible multi-attribute \modelours architecture in Appendix~\ref{sec:appendix:multi-attribute} by element-wise multiplication of gates, and report preliminary results for two attributes Appendix~\ref{sec:appendix:results:multi-attribute}.}

\subsection{Training and Inference}
\label{sec:method:training}

Training \modelours to control an attribute requires two distinct training signals. The first training signal comes from the loss function of the main task, denoted by $\mathcal{L}_{task}$. Th second loss is dedicated to each attribute $i$, denoted by $\mathcal{L}_{\rho_{i}}$. \modelours is agnostic to the choice of the training signal for $\mathcal{L}_{task}$ and $\mathcal{L}_{\rho_{i}}$ as we show by deliberately choosing different training signals for our experiments. Depending on the task and dataset, common approaches to realizing a defined signal are through utilizing the provided labels in a dataset, or by leveraging the indicators specific to an attribute, particularly in the case of lack of reliable supervised data~\cite{lauscher_2021_sustainable,romanov2019s}. Equation~\ref{eq:total_loss} shows the overall loss of each attribute where $\lambda$ is the scaling factor to influence the strength of attribute loss.
\begin{equation}
    \mathcal{L}_{total_{i}} = \mathcal{L}_{task} + \lambda\mathcal{L}_{\rho_{i}}
    \label{eq:total_loss}
\end{equation}

\begin{algorithm}[t]
	\caption{\modelours Training} 

	\begin{algorithmic}[1]
        \State \textbf{Input:} Task-related parameters $\Theta$, parameters of $i$\textsuperscript{th} \modelours $\theta_i$
        \If {Parallel-Training}
            \While {training}
                \State Set $\omega_i=0$ 
                \State Update $\Theta$ using $\mathcal{L}_{task}$
                \State Set $\omega_i=1$ and freeze $\Theta$
                \State Update $\theta_i$ using $\mathcal{L}_{task}+\mathcal{L}_{\rho_i}$
            \EndWhile
        \ElsIf{Posthoc-Training}
            \State Set $\omega_i=0$
            \While {training}
                \State Update $\Theta$ using $\mathcal{L}_{task}$
            \EndWhile
            \State Set $\omega_i=1$ and freeze $\Theta$
            \While {training}
                \State Update $\theta_i$ with $\mathcal{L}_{task}+\mathcal{L}_{\rho_i}$
            \EndWhile
        \EndIf
	\end{algorithmic} 
 \label{alg:training}
\end{algorithm}

Depending on the task and mitigation objective, these loss terms can be defined differently. We explain two realizations of these loss functions later in this section, which we later utilize in our classification and IR experiments. Regardless of the choice of the loss, we first define training procedures for \modelours as follows. For parallel training, the \modelours modules are trained simultaneously with the task, and in post-hoc training, \modelours is added to a fully-trained model in order to learn attribute-specific information. Algorithm~\ref{alg:training} shows the pseudocode of these training strategies. The task-related parameters are denoted with $\Theta$, which can be the whole parameters of an LM, or the ones of an additional task-specific modular network such as a task adapter. In each training cycle, regardless of the loss function, we first deactivate the \modelours by setting $\omega_i = 0$ and use the task loss ($\mathcal{L}_{task}$) to train $\Theta$. We then activate the \modelours module by setting $\omega_i=1$, and update its parameters using equation~\ref{eq:total_loss} to encapsulate the information of the target attribute into the respective \modelours while maintaining task performance. While parallel training enables higher flexibility in optimization and exposes the task head indirectly to bias mitigation loss, post-hoc training offers the practical benefits of adding controllable gates to an existing trained model. As described, the \modelours's parameters are trained only with full engagement ($\omega_i=1$), and the model is never exposed to the settings with partial engagement ($0<\omega_i<1$). This makes the training of \modelours efficient and comparable to the training model for each attribute individually (\eg using adapters).

At inference time, \ttsig reshaping characteristics indicate how much the target attribute should affect the embeddings, by setting $\omega_i$ to any value in $[0,1]$. In particular, when $\omega_i = 0$, the model works at its original (initial) state with no effect from the \modelours module. By increasing $\omega_i$ and changing the shape of the \ttsig, the effect of the gate increases, and a stronger transformation is applied to the embeddings. With $\omega_i = 1$, \modelours reaches its full transformation capacity by applying the sigmoid activation function. We examine this continuous controllability in the following sections. 

In what follows, we explain the realizations of the $\mathcal{L}_{task}$ and $\mathcal{L}_{\rho_{i}}$ in the classification and and search bias mitigation scenarios.

\textbf{Classification loss} uses cross-entropy loss between task labels $y$ and model's output $f(\vz)$:
\begin{equation}
     \mathcal{L}_{task}= \text{CE}(f(\vz),y) \nonumber
\end{equation}
where $\vz$ is the encoded output embedding, and $f$ the classification head. The disentanglement loss $\mathcal{L}_{\rho_{i}}$ can be realized by various methods such as mutual information reduction methods~\cite{colombo_2021_novel}, adversarial training~\cite{elazar_2018_adv} or other loss related to representational/empirical fairness~\cite{ravfogel_2020_null}. We use adversarial loss~\cite{kumar2023parameter,lauscher_2021_sustainable,zhang_2021_disentangling,hauzenberger2023parameter} by defining a classification head $h_{\rho_{i}}$ for the attribute $i$ to predict the corresponding label $y_{\rho_i}$. The adversarial loss follows a min-max optimization, aiming to reduce predictability of the attribute while increasing task performance. 
Following previous studies, we use gradient reversal layer and turn min-max optimization into a minimization problem, formulated below.
\begin{equation*}
\mathcal{L}_{\rho_i} = 
\text{CE}(h_{\rho_i}(\vz),y_{\rho_i})     \label{eq:classification:adversarial_loss}
\end{equation*}

\textbf{Search bias regularization loss}  Following \citet{zerveas_2022_mitigating}, we utilize the CODER framework~\cite{zerveas:coder} to optimize both task and bias mitigation in a list-wise fashion using the ListNet loss~\citep{cao_learning_2007}. For the main task, $\mathcal{L}_{task}$ is the KL-divergence between the distributions of ground-truth relevance labels ($y$) and the predicted scores ($\hat{s}$), defined over $N$ candidate documents. Denoting ground-truth labels and predicted scores as $y$ and $\hat{s}$, accordingly, $\mathcal{L}_{task}$ is formulated as:
\begin{equation*}
     \mathcal{L}_{task} = \text{D}_{\text{KL}}(\phi(y)||\phi(\hat{s}))=-\sum_{j=1}^{N}\phi(y)_j\log\frac{\phi(\hat{s})_j}{\phi(y)_j} 
     \label{eq:IR:task_loss}
\end{equation*}
where $\phi$ refers to the softmax function applied to the values. We enforce the neutrality of retrieved documents, with a list-wise regularization term added to the task loss. The fairness term is similarly formulated with KL-divergence, namely between the distribution of the neutrality scores of the target labels ($y_{\rho_i}$), and the one of the predicted scores $\hat{s}$, formulated below:
\begin{equation*}
     \mathcal{L}_{\rho_i} = \text{D}_{\text{KL}}(\phi(\hat{s})||\phi(\rho))=-\sum_{j=1}^{C}\phi(\hat{s})_j\log\frac{\phi(y_{\rho_i})_j}{\phi(\hat{s})_j}
     \label{eq:IR:reg_loss}
\end{equation*}
As indicated in Eq.~\ref{eq:total_loss}, in both classification and IR scenarios the bias mitigation loss is scaled with hyperparameter $\lambda$ and added to the corresponding task loss.

\section{Experiment Setup}
\label{sec:results:experiments}

\paragraph{Datasets}

We conduct our classification experiments on three datasets: The \textbf{BIOS}~\cite{de:bios} dataset which contains short biographies used to predict a person's job. The name and any indication of the person's gender in the biography are omitted. The dataset labels are 28 occupations for the task, and two protected attribute classes (female/male). The second dataset is \textbf{FDCL18}~\cite{founta_2018_twitter} for hate speech detection, containing a set of tweets each classified as \emph{hateful}, \emph{abusive}, \emph{spam}, or \emph{normal}. Following previous studies~\cite{sap_2019_risk,ravfogel_2020_null}, we assign race dialect labels of \emph{African American} and \emph{White American} to FDCL18 using the probabilistic model developed by \citet{blodgett_2016_demographic}. The third dataset is \textbf{PAN16}~\cite{rangel:pan16} containing a set of tweets accompanied by the labels of gender and age of the authors. The task's objective is to predict whether another user is mentioned in a tweet. PAN16 provides the binary task classes of \emph{mention}, and \emph{no mention}, two gender labels, and five age groups. For the IR task, we use the fairness-sensitive queries dataset \textbf{MSMARCO}$_{Fair}$~\cite{rekabsaz_2021_societal} which contains  215 queries. The queries are from the MSMARCO Passage Retrieval collection~\cite{msmarco}, with 8,841,822 passages. We use 158 words related to each of the two protected attribute classes (female/male) following~\cite{zerveas_2022_mitigating}, to 
calculate the neutrality of each document following~\cite{rekabsaz_2021_societal}. The details of the neutrality criteria can be found in appendix~\ref{appendix:sec:nfairr}.

\paragraph{LMs and Training.} For the classification benchmarks, we conduct the experiments on three LMs namely, BERT-Base~\cite{devlin:bert}, BERT-Mini\cite{turc_2019_bert_small} and RoBERTa~\cite{liu_2019_roberta}. In all experiments, $\mathcal{L}_{task}$ is realized by cross-entropy and binary cross-entropy loss for binary classes and adversarial training to remove attributes. We train our models with a parallel strategy. The adversarial head consists of an ensemble of 5 networks for each attribute, and each network consists of two fully connected layers with Tanh activation in between. The overall loss of the adversarial is scaled by $\lambda=1$. For the IR task, following~\citet{zerveas:coder} we conduct the experiments on DistilBERT~\cite{sanh:distilbert}. As explained in Section~\ref{sec:method:training}, The loss function ($\mathcal{L}_{task}$) is realized by ListNet~\cite{cao_learning_2007}, and fairness is achieved through the fairness regularization loss ($\lambda\mathcal{L}_{\rho}$)~\cite{zerveas_2022_mitigating}. Each baseline model is trained with several values of the regularization terms $\lambda$, while \modelours is trained only once with $\lambda=20$.

\begin{table*}[ht]
\centering
\caption{Results of BERT-Base, RoBERTa-Base and BERT-Mini models on three datasets and four attributes. The \modelours sensitivity parameter is set to fully debiasing ($\omega=1$) to have comparable results with the baselines}
\small
\scalebox{1}{
\begin{tabular}{l l ll | ll | ll | ll}
\toprule
\multirow{2}{*}{Model} &\multirow{2}{*}{Type} &\multicolumn{2}{c}{\textbf{BIOS}} & \multicolumn{2}{c}{\textbf{FDCL18}}& \multicolumn{2}{c}{\textbf{PAN16-Gender}}& \multicolumn{2}{c}{\textbf{PAN16-Age}} \\
& & \textbf{Task}$\uparrow$ & \textbf{Probe}$\downarrow$ & \textbf{Task}$\uparrow$ & \textbf{Probe}$\downarrow$&  \textbf{Task}$\uparrow$ & \textbf{Probe}$\downarrow$&  \textbf{Task}$\uparrow$ & \textbf{Probe}$\downarrow$ \\\midrule

\multirow{5}{*}{\modelbert}
&\modelfine &$84.6_{0.4}$ &$67.3_{0.8}$ & $81.0_{1.0}$ & $92.9_{1.8}$ &$93.6_{1.8}$ &$69.6_{0.8}$ &$93.6_{1.8}$ &$42.3_{0.9}$ \\
&\modeladapter &$84.3_{0.1}$ &$67.0_{0.1}$ &$80.0_{0.1}$  &$93.3_{0.4}$ &$92.4_{0.1}$ &$70.7_{0.1}$ &$92.4_{0.1}$ &$42.4_{0.}$ \\
&\modelfinedeb &$84.0_{0.3}$ &$60.8_{0.2}$ & $81.0_{1.0}$ &$84.4_{4.0}$ &$92.4_{0.8}$ &$59.8_{0.7}$ &$92.4_{0.8}$ &$31.3_{1.1}$ \\
&\modeladapterdeb  &$84.2_{0.1}$ & $61.9_{0.5}$&$79.8_{0.3}$  &$75.6_{0.5}$ &$92.2_{0.1}$ &\textbf{54.2}$_{0.4}$ &$92.1_{0.1}$ &$21.7_{0.1}$ \\
&\modelours  &\textbf{85.0}$_{0.1}$  &\textbf{58.7}$_{0.4}$ &$81.0_{0.2}$  & \textbf{67.5}$_{0.6}$ &\textbf{93.8}$_{0.1}$ &$55.3_{0.5}$ &\textbf{93.8}$_{0.1}$ &\textbf{21.3}$_{0.6}$ \\
\midrule

\multirow{5}{*}{\modelroberta}
&\modelfine &$84.5_{0.4}$ &$66.2_{0.7}$ &$80.6_{0.4}$ &$93.2_{1.2}$ &\textbf{98.5}$_{0.1}$ &$63.6_{0.4}$ & \textbf{98.5}$_{0.1}$ &$22.7_{0.8}$ \\
&\modeladapter &$84.3_{0.1}$ &$67.3_{0.7}$ &$80.0_{0.6}$ &$94.0_{0.6}$ &$98.2_{0.1}$ &$62.8_{0.4}$ &$98.1_{0.1}$ & $31.9_{0.1}$ \\
&\modelfinedeb &$84.1_{0.3}$ &$61.6_{0.3}$ &$80.5_{1.0}$ &$83.6_{1.9}$ &$98.2_{0.1}$ &$52.0_{0.9}$ &$98.2_{0.1}$ & $24.1_{1.4}$ \\
&\modeladapterdeb  &$84.0_{0.1}$ &$62.9_{0.1}$ &$80.0_{0.5}$ &$79.7_{0.3}$ &$98.1_{0.1}$ &\textbf{53.7}$_{0.7}$ & $98.0_{0.1}$ &$22.3_{1.0}$  \\
&\modelours & \textbf{84.8}$_{0.1}$ &\textbf{61.4}$_{1.0}$ &\textbf{81.4}$_{0.2}$ &\textbf{73.9}$_{0.8}$ &$98.4_{0.1}$ &$55.4_{0.8}$ &$98.4_{0.1}$ &\textbf{22.2}$_{0.1}$\\
\midrule
    
\multirow{5}{*}{\modelmini}
&\modelfine &\textbf{82.4}$_{0.1}$ &$65.7_{0.2}$ &$79.9_{1.0}$ &$92.4_{0.6}$ &\textbf{91.5}$_{0.1}$ &$65.4_{0.8}$ &\textbf{91.5}$_{0.1}$ &$40.9_{0.4}$ \\
&\modeladapter &$82.1_{0.1}$ &$65.2_{0.4}$ &$81.3_{0.1}$ &$93.5_{0.8}$ &$81.5_{0.1}$  &$65.5_{0.4}$  &$81.6_{0.1}$  & $37.4_{1.4}$  \\
&\modelfinedeb &$81.7_{0.1}$ & $60.4_{0.4}$ &$79.3_{0.8}$ &$81.4_{2.1}$ &$90.3_{0.4}$ &$58.6_{0.8}$ &$90.3_{0.4}$ &$27.9_{1.8}$ \\
&\modeladapterdeb  &$82.1_{0.2}$ &$61.4_{0.6}$ &$80.7_{0.1}$ &$75.9_{1.2}$ &$81.3_{0.1}$  &\textbf{53.1}$_{0.1}$  &$81.1_{0.1}$  & \textbf{21.7}$_{0.3}$  \\
&\modelours  &$81.7_{0.4}$ &\textbf{59.2}$_{0.1}$ &\textbf{81.9}$_{0.1}$ &\textbf{62.5}$_{0.5}$ &$90.2_{0.1}$  &$56.4_{0.1}$  &$89.9_{0.2}$  &$21.8_{0.2}$ \\
\bottomrule

\end{tabular}
}

\label{tab:results:all_models}
\vspace{2mm}
\end{table*}

\vspace{-1mm}
\paragraph{Models} Considering that parameter-wise \modelours is the same as adapters we choose our baselines as follows: \textbf{\modelfine} finetunes all parameters of the LM on the task with no debiasing objective for classification. \textbf{\modelfinedeb} also finetunes all parameters of
the LM on the task with debiasing objective. \textbf{\modeladapter} uses a standard adapter network and trains the adapter only on the task. \textbf{\modeladapterdeb} uses the same adapter architecture but trains it with task and attribute removal objectives simultaneously.  The complete details of our hyperparameters setting and training procedure are explained in Appendix~\ref{sec:appendix:experiment}

\begin{figure*}
     \centering
     \begin{subfigure}[b]{0.51\columnwidth}
         \centering
         \includegraphics[width=\textwidth]{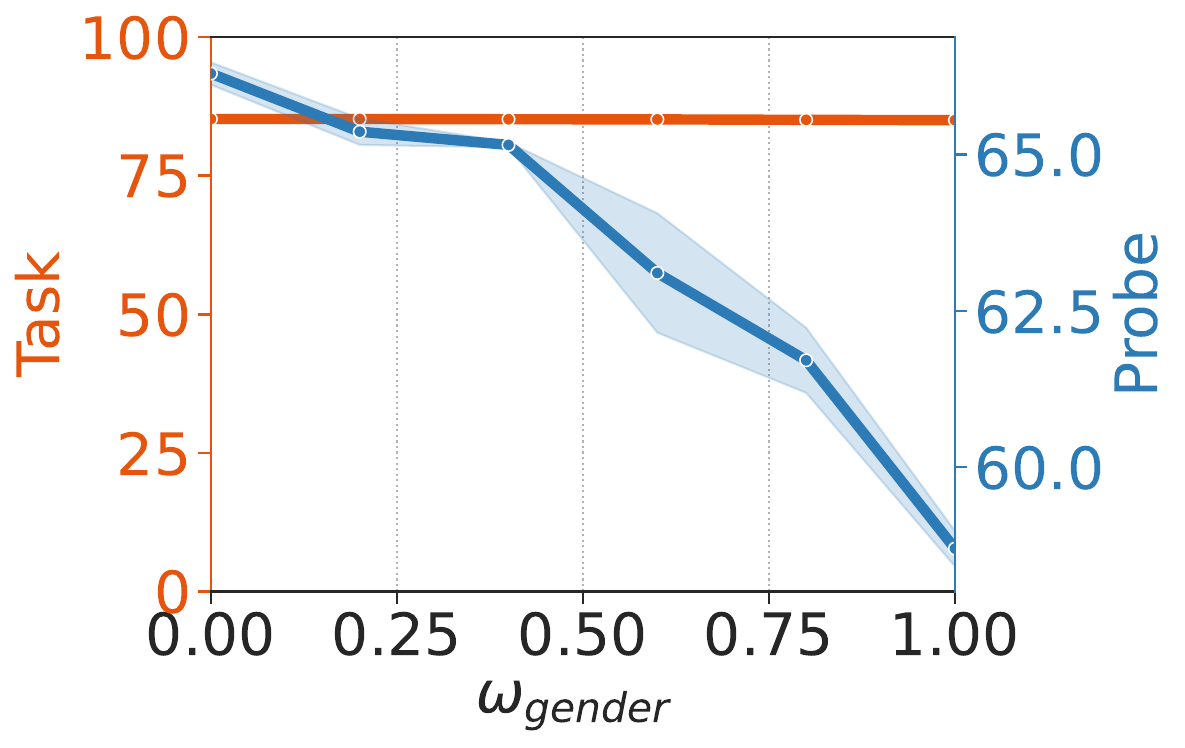}
         \caption{BIOS (Gender)}
         \label{fig:results:probes:bios}
     \end{subfigure}
     \hfill
     \begin{subfigure}[b]{0.51\columnwidth}
         \centering
         \includegraphics[width=\textwidth]{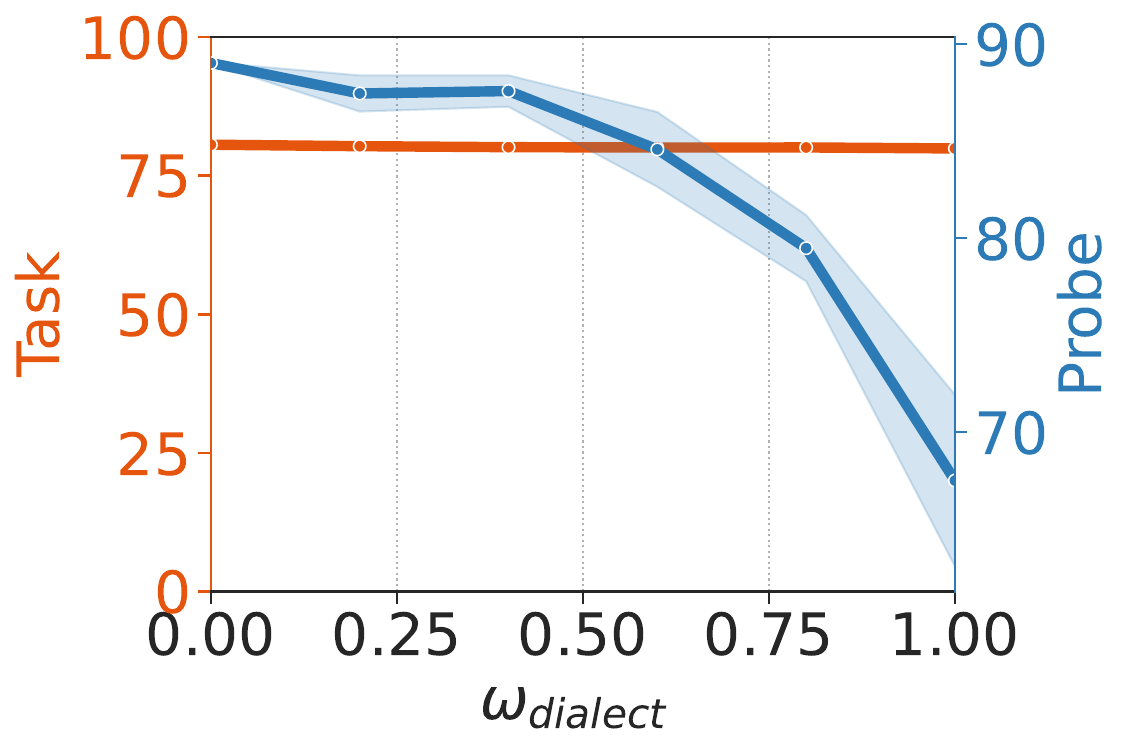}
         \caption{FCDL18 (Dialect-race)}
         \label{fig:results:probes:fcdl}
     \end{subfigure}
     \hfill
     \begin{subfigure}[b]{0.51\columnwidth}
         \centering
         \includegraphics[width=\textwidth]{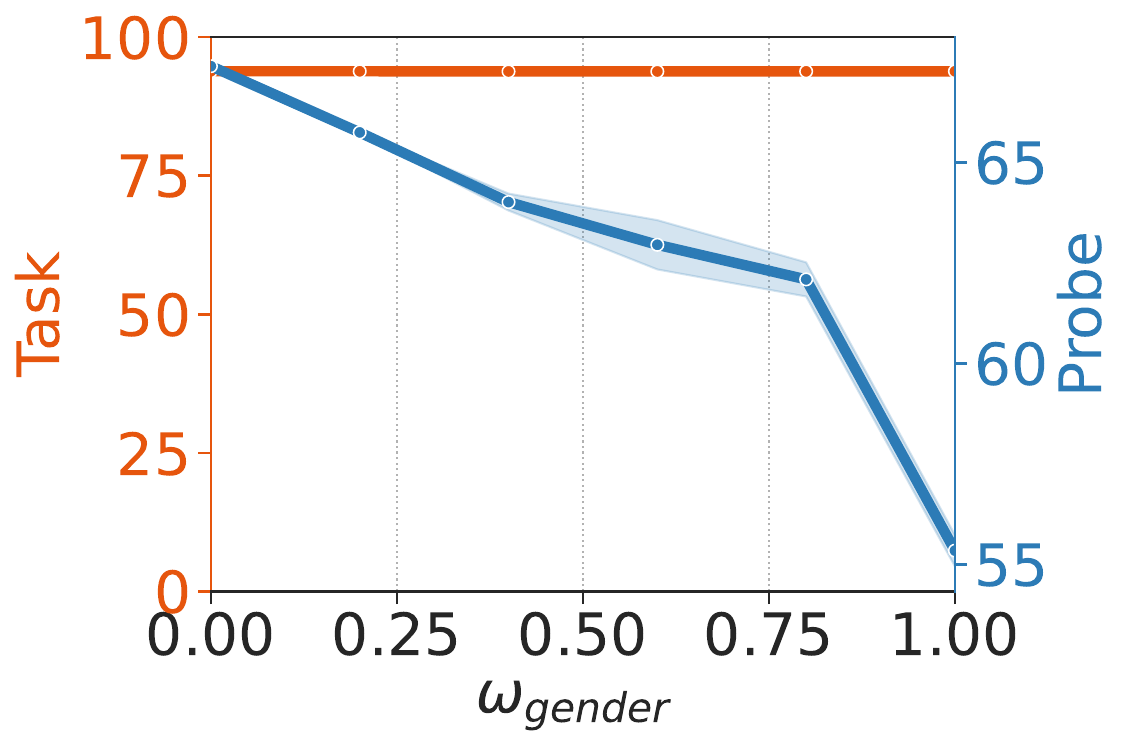}
         \caption{PAN16 - Gender}
         \label{fig:results:probes:pangender}
     \end{subfigure}
     \hfill
     \begin{subfigure}[b]{0.51\columnwidth}
         \centering
         \includegraphics[width=\textwidth]{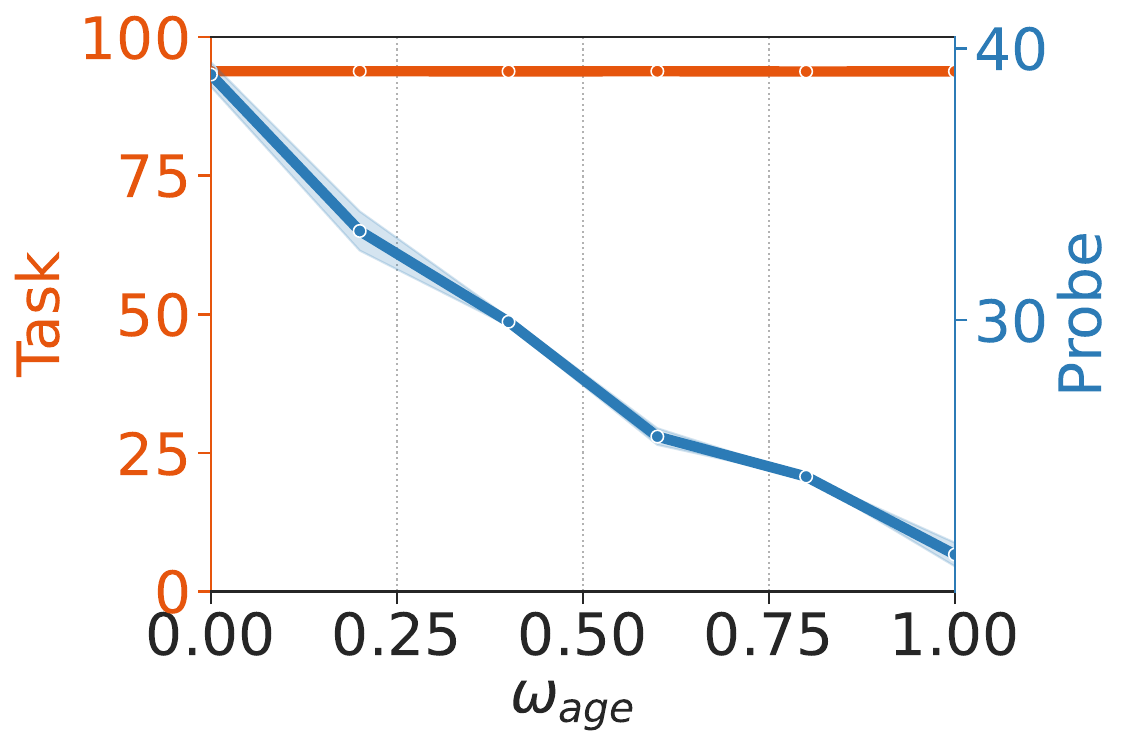}
         \caption{PAN16 - Age}
         \label{fig:results:probes:panage}
     \end{subfigure}
     \caption{Results of the \modelours models using BERT-Base when increasing the gating sensitivity $\omega$ from 0 (no effect) to 1 (full effect). Each trained model is evaluated multiple times on the various $\omega$ values adjusted at inference time. The left/right y-axis corresponds to the task performance and attribute probing results, respectively. The results show the continuous reduction in the information presence of the target concept, as $\omega$ increases. 
     }
    \label{fig:results:probes}    
     \vspace{-2mm}
\end{figure*}

\vspace{-1mm}
\paragraph{Metrics}

We evaluate the performance of the classification models on the core task using the accuracy metric. Following the previous works of~\citet{kumar2023parameter} and~\citet{hauzenberger2023parameter} we measure attribute information using strong probing networks. For each model, we train 5 independent classification heads (two-layer feed-forward layer with a $\text{Tanh}$ activation) for 30 epochs to extract and predict the target attribute. We report the average performance of the probes in terms of balanced/macro accuracy (average of per-class accuracy scores). This evaluation measures how much information about a given attribute still exist in the model and can be recovered. Balanced accuracy has the benefit of better reflecting the performance of the methods when considering minority groups, particularly given the unbalanced distributions over protected labels in the datasets. For the IR task, we use the Mean Reciprocal Rank (MRR@10) of the top 10 retrieved documents as a metric for the core task evaluation and Normalized Fairness of Retrieval Results (NFaiRR@10) from the top 10 retrieved documents as the fairness metric of the model~\cite{rekabsaz_2021_societal}. NFaiRR metric normalizes the pre-query FaiRR score over the ideal FaiRR achieved from a background set of documents allowing comparable results across queries (detail in appendix~\ref{appendix:sec:nfairr}). To account for possible variations, we report our results as the average of 3 independently-trained models, and the report mean and standard deviation of the results.

\begin{figure}[t]
     \centering
     \includegraphics[width=1\columnwidth]{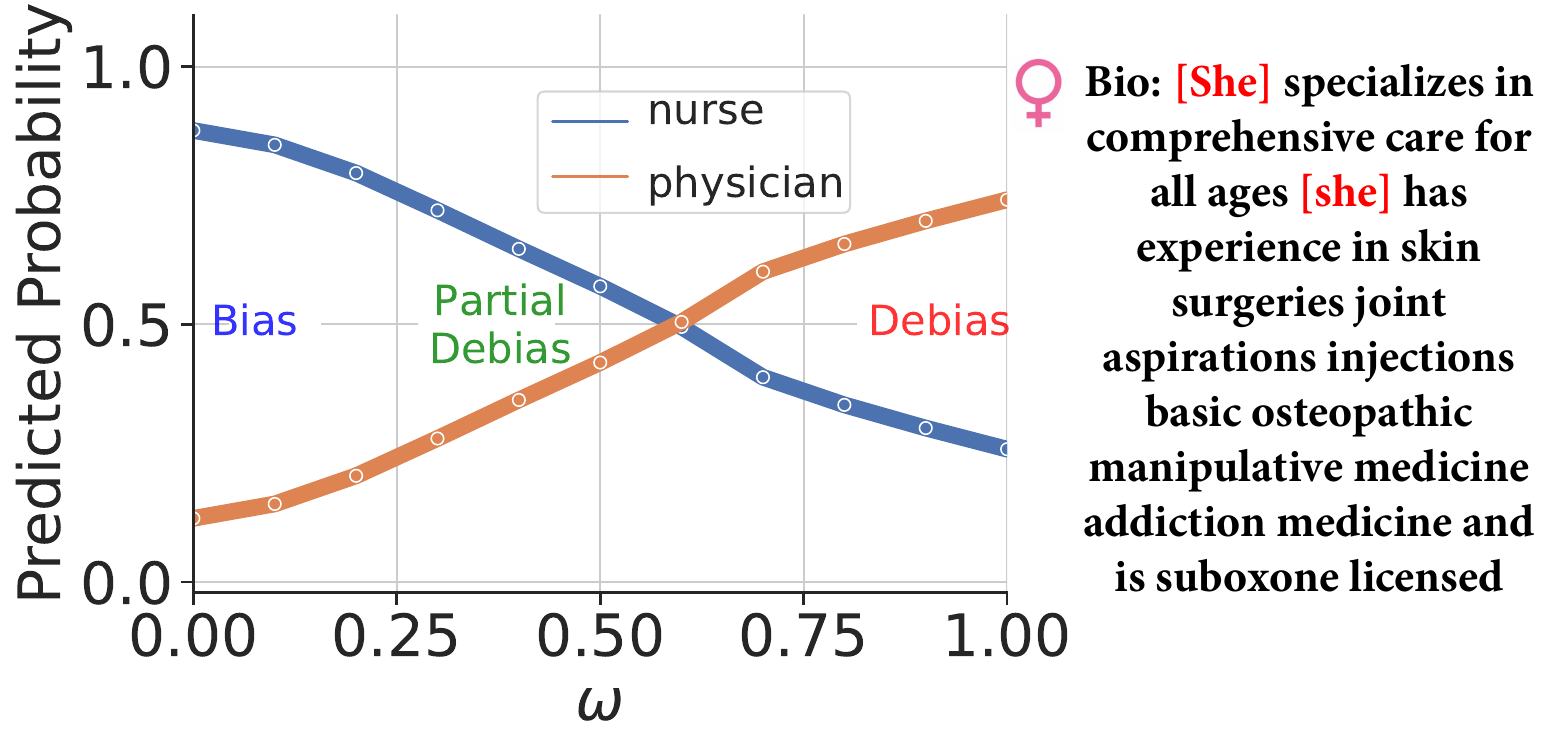}
     \caption{Prediction probabilities of \modelours when gradually increasing $\omega$, for a female physician's biography, incorrectly classified as a nurse in the initial state. The figure illustrates how changing the strength of gender removal affects the model's decision, providing a higher degree of interpretability through controllability.}
    \label{fig:prob_nurse}
     \vspace{-4mm}
\end{figure}

\section{Results and Discussion}
\label{sec:results}

We start our discussion by going through results of the classification tasks, and continue discussing our observations on the search result experiment. 

\subsection{Classification Tasks}
\label{sec:result:classification}

Table~\ref{tab:results:all_models} reports the results of the baselines, as well as \modelours at full bias mitigation power ($\omega=1$) for 3 models, 3 datasets, and 4 attributes. As shown, the fully activated \modelours models contain less information about the attribute in comparison to fine-tuned baselines (\modelfine\!*) on all datasets across all models while having lower task performance only on BERT-Mini model (Pan16, BIOS) and slightly worse on ROBERTA-Base (Pan16 dataset). We observe a higher task performance drop on BERT-Mini which we hypothesize is due to the small size of the network. In BERT-Mini the information about the attribute (gender/age) and task might be intertwined, which makes it difficult for \modelours to filter out attribute information without losing task information. In comparison to adapter-based models, on 11 out of 12 classification experiments \modelours perform on par or better on the task and remove more information on 3 out of 4 attributes. On the BERT-mini and Pan16 dataset, even though adapter-based models are better than \modelours in terms of attribute information removal but (\modeladapter\!*) are not able to achieve satisfactory task performance(81.1\%) which we assume is due to a lack of enough learning capacity of adapters for this task. Overall our experiments indicate that \modelours is able to perform the task better than baselines while enhancing information removal from the embeddings of the model.

\begin{table}[t]
\centering
\caption{DistilBERT-Base Results on \textbf{MSMARCO}$_{Fair}$ benchmark.}
\label{tab:results:IR}
\small
\scalebox{0.9}{
\begin{tabular}{l l | ll}

\toprule
\multirow{2}{*}{Type} &\multirow{2}{*}{$\lambda$} & \multicolumn{2}{c}{\textbf{MSMARCO}}\\
& & \textbf{MRR@10}$\uparrow$ & \textbf{NFaiRR@10}$\uparrow$ \\\midrule

\multirow{4}{*}{\modelfine}
 &0.0 &0.234$_{0.002}$ &0.904$_{0.001}$ \\
 &5.0 &0.183$_{0.003}$ &0.943$_{0.002}$ \\
 &10 &0.142$_{0.002}$ &0.955$_{0.001}$ \\
 &20 &0.079$_{0.003}$ &0.972$_{0.000}$ \\
 \midrule
\multirow{4}{*}{\modeladapter}
 &0.0  &0.230$_{0.002}$ &0.898$_{0.001}$ \\ 	
 &5.0 &0.147$_{0.003}$ &0.933$_{0.000}$ \\
 &10 &0.082$_{0.000}$ &0.949$_{0.001}$ \\
 &20 &0.023$_{0.002}$ &0.965$_{0.001}$ \\
\cdashlinelr{1-4}
\multicolumn{4}{l}{\modelours}\\ 
$\omega=0.0$& 0.0  &0.234$_{0.003}$ & 0.903$_{0.001}$  \\
$\omega=0.4$& - &0.227$_{0.002}$  &  0.917$_{0.000}$\\
$\omega=0.8$& - &0.208$_{0.001}$  & 0.942$_{0.000}$\\
$\omega=1.0$ &20.0 &0.168$_{0.007}$  &  0.970$_{0.000}$\\
\bottomrule
\end{tabular}
}

\vspace{-4mm}

\end{table}

We also investigate how much information about each attribute exists in the model when changing $\omega$ and its influence on the task performance. We investigate by changing the $\omega$ and retraining the probes. Figure~\ref{fig:results:probes} shows the task performance and probing results for the \modelours for the BERT-Base model when increasing $\omega$ parameters. The reported result for each probing (each attribute with a specific $\omega$) is the mean and standard deviation of 5 probes applied to 3 independently trained models. The task performance and probing results are shown in orange (left y-axis) and blue (right y-axis), respectively. The results of the RoBERTa and BERT-Mini are reported in Appendix~\ref{sec:appendix:results:mini-roberta}.

Consistent across all datasets and attributes, we observe that increasing $\omega$ leads to a continuous decrease in the presence of the corresponding attribute, until reaching the lowest probing balanced accuracy at $\omega=1$. This continuous attribute removal is achieved while maintaining task performance. On the whole, our results point to the ability of \modelours to impose graded control over an attribute at inference time.

To exemplify how continuous controllability at inference time enhances interpretability, and provides a higher level of model transparency to end-users, figure~\ref{fig:prob_nurse} depicts how changes in $\omega_{gender}$ can influence models' decision probability about a female physician who is labeled as nurse by the biased model $\omega=0$. Figures~\ref{fig:appendix:sample:positive:1}-\ref{fig:appendix:sample:positive:5} in Appendix~\ref{sec:appendix:results} provide more examples of such positive/negative effects from the studied datasets.

In Appendix~\ref{sec:appendix:results:multi-attribute}, we investigate the simultaneous bias control of the gender and age attributes of PAN16 using the multi-concept \modelours 
and the changes in the behavior of the model by investigating the alterations in uncertainty (\ref{sec:appendix:results:uncertainty}) and prediction labels (\ref{sec:appendix:result:flips}). We also study the effect of \modelours adversarial bias mitigation effect on empirical fairness metrics in Appendix~\ref{sec:appendix:results:fairness}.

\begin{figure}[t]
\centering
\includegraphics[width=0.9\columnwidth]{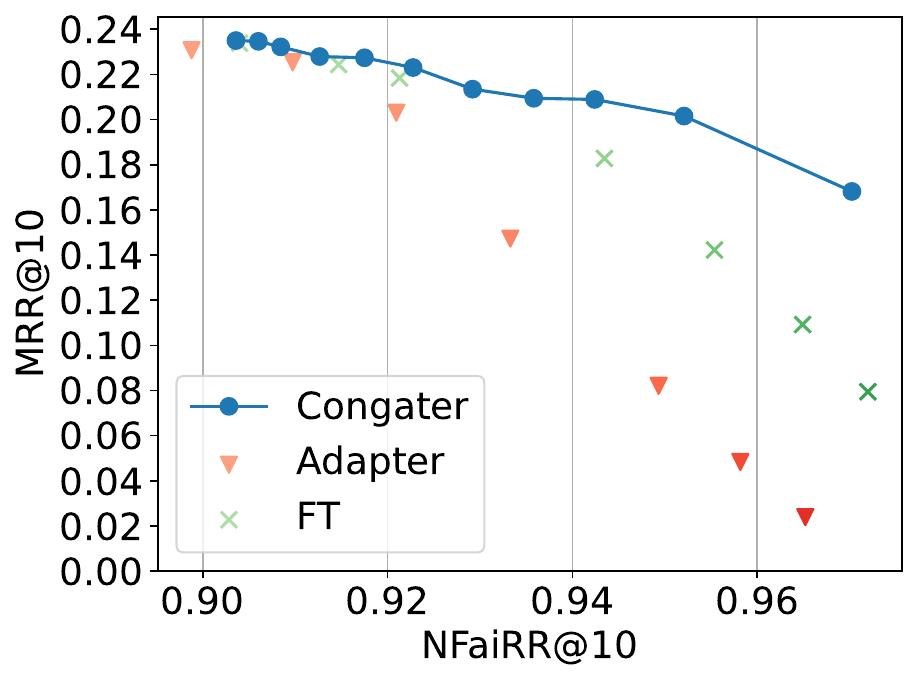}
\caption{Fairness-performance trade-off between \modelfine, \modeladapter, and \modelours. For baselines, each point refer to a new model training with color intensities indicating the degree of the regularization coefficient $\lambda$. \modelours is trained only once, and each point indicates the evaluation according to an $\omega$ value. 
} 
\label{fig:results:IR}
\end{figure}

\subsection{Search Result Bias Mitigation Benchmark}
\label{sec:result:IR}

Table~\ref{tab:results:IR} shows the performance and search results fairness results of the baseline models trained with different regularization coefficients ($\lambda$), as well as the \modelours model trained once with $\lambda= 0$ and $\lambda=20$ according to the post-hoc strategy. 
The same results with more data points are illustrated in Figure~\ref{fig:results:IR}. To achieve a higher degree of fairness in  baseline models, we increase $\lambda$ and retrain the models, while in \modelours, we simply increase the value of $\omega$ at test time.

Consistent with previous studies~\citet{zerveas_2022_mitigating,rekabsaz_2021_societal}, increasing $\lambda$ leads to a decrease in task performance (fairness-performance trade-off) in all models.  However, interestingly the \modelours with highest degree of fairness ($\omega=1$) achieves  $\text{MRR}@10=0.168$ performance, which is around 2.1 times higher than the \modelfine with the same level of fairness (at $\lambda=20$). It is noticeable that the performance of \modeladapter -- as the modularized approach -- radically deteriorates for high values of $\lambda$. In addition to higher performance perseverance, \modelours is able to monotonically navigate between the states (from the one with the least to the most fairness scores), enabling an effective control over the fairness-performance trade-off at inference time. 
This functionality of \modelours can be leveraged in practice by designing a personalized control \emph{knob} in the hands of the end-users and practitioners, which empowers them to adjust the level of fairness/neutrality in search results.

\section{Conclusion}
We introduce \modelours, a gated module enhanced with a novel controllable activation function, which enables continuous adjustment for the information flow of an attribute at inference time. We conduct 13 experiments spanning 2 tasks (classification and information retrieval) 4 datasets, and 4 models to remove attribute information (classification) and enhance fairness of the model (IR). Our results show that \modelours can successfully isolate and filter out attribute information with the least harm to task performance. In addition, we showed that \modelours is able to continuously traverse between the biased and debiased states, enhancing personalization and interpretability through controllability.

\section{Ethical Considerations and Limitations}
A limitation of our work concerns the lack of completeness in the definition of the concept and attributes provided by the datasets. In particular, gender in all datasets including BIOS, PAN16, and MSMARCO is limited to the binary female/male, lacking an inclusive and nuanced definition of gender. Similarly in FDCL18, we consider only two dialects of \emph{African American} and \emph{White American}, while clearly this definition is limited and non-inclusive. Furthermore, as in previous work~\cite{sap_2019_risk,ravfogel_2020_null, zhang_2021_disentangling,kumar2023parameter}, the labels of this protected attribute are assigned through a probabilistic model, and hence the dataset might not represent the nuances and traits of the real-world. 

The second limitation regards the degree of the generalization of our method with respect to various deep learning architectures (such as CNNs), as our definition and experiments are constrained to the use of transformer-based models.

The Third limitation regarding the generalization of the method is the multi-attribute setting for \modelours over any possible number of concepts or a subset of them. We conduct our experiments with a focus on only one attribute and introduce a possible fusion method and a preliminary result. Our multi-attribute experiment is only conducted on one dataset with two attributes of gender and age, particularly due to the lack of availability of suitable datasets. We note that the conclusion provided in the paper should be viewed to the extent of these experiments, and further studies (as well as more suitable datasets) are required to achieve a more comprehensive picture of this topic.

As a general limitation shared with the other related studies on in this domain, we should note that the aim of representation disentanglement optimizations is to reduce the information about a particular concept inside model embeddings with attributes based on the \emph{observed data}. These data-oriented approaches might lack effective generalization, particularly when the model is evaluated on other domains or out-of-distribution data.

What we are proposing is a single model capable of handling both biased decisions, partially debiased decisions, and unbiased decisions which creates much more flexibility for the end user. Any misuse of the proposed method in tasks where fairness for the users are high priority such as job recommendations is not the intention of the authors and is considered one of the dangers of the proposed model.  

\section{Acknowledgment}
This work received financial support by the State of Upper Austria and the Federal Ministry of Education, Science, and Research, through grants LIT-2020-9-SEE-113 and LIT-2021-YOU-215.
This work is also supported by the Austrian Science Fund (FWF): P36413, P33526, and DFH-23. 
\bibliography{reference}

\begin{thebibliography}{67}
\expandafter\ifx\csname natexlab\endcsname\relax\def\natexlab#1{#1}\fi

\bibitem[{Barrett et~al.(2019)Barrett, Kementchedjhieva, Elazar, Elliott, and
  S{\o}gaard}]{barrett_2019_adversarial}
Maria Barrett, Yova Kementchedjhieva, Yanai Elazar, Desmond Elliott, and Anders
  S{\o}gaard. 2019.
\newblock Adversarial removal of demographic attributes revisited.
\newblock In \emph{Proceedings of the Conference on Empirical Methods in
  Natural Language Processing and the 9th International Joint Conference on
  Natural Language Processing (EMNLP-IJCNLP)}, pages 6331--6336.

\bibitem[{Blodgett et~al.(2020)Blodgett, Barocas, Daum{\'e}~III, and
  Wallach}]{blodgett2020language}
Su~Lin Blodgett, Solon Barocas, Hal Daum{\'e}~III, and Hanna Wallach. 2020.
\newblock Language (technology) is power: A critical survey of “bias” in
  nlp.
\newblock In \emph{Proceedings of the 58th Annual Meeting of the Association
  for Computational Linguistics}, pages 5454--5476.

\bibitem[{Blodgett et~al.(2016)Blodgett, Green, and
  O{'}Connor}]{blodgett_2016_demographic}
Su~Lin Blodgett, Lisa Green, and Brendan O{'}Connor. 2016.
\newblock \href {https://doi.org/10.18653/v1/D16-1120} {Demographic dialectal
  variation in social media: A case study of {A}frican-{A}merican {E}nglish}.
\newblock In \emph{Proceedings of the 2016 Conference on Empirical Methods in
  Natural Language Processing}, pages 1119--1130, Austin, Texas. Association
  for Computational Linguistics.

\bibitem[{Cao et~al.(2007)Cao, Qin, Liu, Tsai, and Li}]{cao_learning_2007}
Zhe Cao, Tao Qin, Tie-Yan Liu, Ming-Feng Tsai, and Hang Li. 2007.
\newblock \href {https://doi.org/10.1145/1273496.1273513} {Learning to rank:
  from pairwise approach to listwise approach}.
\newblock In \emph{Proceedings of the 24th international conference on
  {Machine} learning}, {ICML} '07, pages 129--136, New York, NY, USA.
  Association for Computing Machinery.

\bibitem[{Colombo et~al.(2021)Colombo, Piantanida, and
  Clavel}]{colombo_2021_novel}
Pierre Colombo, Pablo Piantanida, and Chlo{\'e} Clavel. 2021.
\newblock A novel estimator of mutual information for learning to disentangle
  textual representations.
\newblock In \emph{Proceedings of the 59th Annual Meeting of the Association
  for Computational Linguistics and the 11th International Joint Conference on
  Natural Language Processing (Volume 1: Long Papers)}, pages 6539--6550.

\bibitem[{De-Arteaga et~al.(2019)De-Arteaga, Romanov, Wallach, Chayes, Borgs,
  Chouldechova, Geyik, Kenthapadi, and Kalai}]{de:bios}
Maria De-Arteaga, Alexey Romanov, Hanna Wallach, Jennifer Chayes, Christian
  Borgs, Alexandra Chouldechova, Sahin Geyik, Krishnaram Kenthapadi, and
  Adam~Tauman Kalai. 2019.
\newblock Bias in bios: A case study of semantic representation bias in a
  high-stakes setting.
\newblock In \emph{proceedings of the Conference on Fairness, Accountability,
  and Transparency}, pages 120--128.

\bibitem[{Devlin et~al.(2018)Devlin, Chang, Lee, and Toutanova}]{devlin:bert}
Jacob Devlin, Ming-Wei Chang, Kenton Lee, and Kristina Toutanova. 2018.
\newblock Bert: Pre-training of deep bidirectional transformers for language
  understanding.
\newblock \emph{arXiv preprint arXiv:1810.04805}.

\bibitem[{Elazar and Goldberg(2018)}]{elazar_2018_adv}
Yanai Elazar and Yoav Goldberg. 2018.
\newblock \href {https://doi.org/10.18653/v1/D18-1002} {Adversarial removal of
  demographic attributes from text data}.
\newblock In \emph{Proceedings of the 2018 Conference on Empirical Methods in
  Natural Language Processing}, pages 11--21. Association for Computational
  Linguistics.

\bibitem[{Fabris et~al.(2020)Fabris, Purpura, Silvello, and
  Susto}]{fabris:fairness}
Alessandro Fabris, Alberto Purpura, Gianmaria Silvello, and Gian~Antonio Susto.
  2020.
\newblock \href {https://doi.org/10.1016/j.ipm.2020.102377} {Gender stereotype
  reinforcement: Measuring the gender bias conveyed by ranking algorithms}.
\newblock \emph{Inf. Process. Manag.}, 57(6):102377.

\bibitem[{Founta et~al.(2018{\natexlab{a}})Founta, Djouvas, Chatzakou,
  Leontiadis, Blackburn, Stringhini, Vakali, Sirivianos, and
  Kourtellis}]{founta:hatespeech}
Antigoni~Maria Founta, Constantinos Djouvas, Despoina Chatzakou, Ilias
  Leontiadis, Jeremy Blackburn, Gianluca Stringhini, Athena Vakali, Michael
  Sirivianos, and Nicolas Kourtellis. 2018{\natexlab{a}}.
\newblock Large scale crowdsourcing and characterization of twitter abusive
  behavior.
\newblock In \emph{Twelfth International AAAI Conference on Web and Social
  Media}.

\bibitem[{Founta et~al.(2018{\natexlab{b}})Founta, Djouvas, Chatzakou,
  Leontiadis, Blackburn, Stringhini, Vakali, Sirivianos, and
  Kourtellis}]{founta_2018_twitter}
Antigoni~Maria Founta, Constantinos Djouvas, Despoina Chatzakou, Ilias
  Leontiadis, Jeremy Blackburn, Gianluca Stringhini, Athena Vakali, Michael
  Sirivianos, and Nicolas Kourtellis. 2018{\natexlab{b}}.
\newblock Large scale crowdsourcing and characterization of twitter abusive
  behavior.
\newblock In \emph{Twelfth International AAAI Conference on Web and Social
  Media}.

\bibitem[{Ganh\"{o}r et~al.(2022)Ganh\"{o}r, Penz, Rekabsaz, Lesota, and
  Schedl}]{ganhoer2022mitigating}
Christian Ganh\"{o}r, David Penz, Navid Rekabsaz, Oleg Lesota, and Markus
  Schedl. 2022.
\newblock \href {https://doi.org/10.1145/3477495.3531820} {Unlearning protected
  user attributes in recommendations with adversarial training}.
\newblock In \emph{Proceedings of the 45th International ACM SIGIR Conference
  on Research and Development in Information Retrieval}, SIGIR '22, page
  2142–2147, New York, NY, USA. Association for Computing Machinery.

\bibitem[{Guo et~al.(2022)Guo, Yang, and Abbasi}]{guo_2022_auto}
Yue Guo, Yi~Yang, and Ahmed Abbasi. 2022.
\newblock Auto-debias: Debiasing masked language models with automated biased
  prompts.
\newblock In \emph{Proceedings of the 60th Annual Meeting of the Association
  for Computational Linguistics (Volume 1: Long Papers)}, pages 1012--1023.

\bibitem[{Hallinan et~al.(2023)Hallinan, Liu, Choi, and
  Sap}]{decoder:hallinan:detoxifying}
Skyler Hallinan, Alisa Liu, Yejin Choi, and Maarten Sap. 2023.
\newblock \href {https://doi.org/10.18653/v1/2023.acl-short.21} {Detoxifying
  text with marco: Controllable revision with experts and anti-experts}.
\newblock In \emph{Proceedings of the 61st Annual Meeting of the Association
  for Computational Linguistics (Volume 2: Short Papers), {ACL} 2023, Toronto,
  Canada, July 9-14, 2023}, pages 228--242. Association for Computational
  Linguistics.

\bibitem[{Han et~al.(2021{\natexlab{a}})Han, Pang, and Wu}]{han2021robust}
Wenjuan Han, Bo~Pang, and Ying~Nian Wu. 2021{\natexlab{a}}.
\newblock Robust transfer learning with pretrained language models through
  adapters.
\newblock In \emph{Proceedings of the 59th Annual Meeting of the Association
  for Computational Linguistics and the 11th International Joint Conference on
  Natural Language Processing (Volume 2: Short Papers)}, pages 854--861.

\bibitem[{Han et~al.(2021{\natexlab{b}})Han, Baldwin, and
  Cohn}]{han_2021_diverse}
Xudong Han, Timothy Baldwin, and Trevor Cohn. 2021{\natexlab{b}}.
\newblock \href {https://doi.org/10.18653/v1/2021.eacl-main.239} {Diverse
  adversaries for mitigating bias in training}.
\newblock In \emph{Proceedings of the 16th Conference of the European Chapter
  of the Association for Computational Linguistics: Main Volume}, pages
  2760--2765, Online. Association for Computational Linguistics.

\bibitem[{Han et~al.(2021{\natexlab{c}})Han, Baldwin, and Cohn}]{han-bias}
Xudong Han, Timothy Baldwin, and Trevor Cohn. 2021{\natexlab{c}}.
\newblock \href {https://doi.org/10.18653/v1/2021.eacl-main.239} {Diverse
  adversaries for mitigating bias in training}.
\newblock In \emph{Proceedings of the 16th Conference of the European Chapter
  of the Association for Computational Linguistics: Main Volume, {EACL} 2021,
  Online, April 19 - 23, 2021}, pages 2760--2765. Association for Computational
  Linguistics.

\bibitem[{Hauzenberger et~al.(2023)Hauzenberger, Masoudian, Kumar, Schedl, and
  Rekabsaz}]{hauzenberger2023parameter}
Lukas Hauzenberger, Shahed Masoudian, Deepak Kumar, Markus Schedl, and Navid
  Rekabsaz. 2023.
\newblock {Modular and On-demand Bias Mitigation with Attribute-Removal
  Subnetworks}.
\newblock In \emph{Findings of the Association for Computational Linguistics:
  ACL {(Findings of ACL)}}.

\bibitem[{Hofst{\"{a}}tter et~al.(2021)Hofst{\"{a}}tter, Lin, Yang, Lin, and
  Hanbury}]{hofstatter:ir:fairness}
Sebastian Hofst{\"{a}}tter, Sheng{-}Chieh Lin, Jheng{-}Hong Yang, Jimmy Lin,
  and Allan Hanbury. 2021.
\newblock \href {https://doi.org/10.1145/3404835.3462891} {Efficiently teaching
  an effective dense retriever with balanced topic aware sampling}.
\newblock In \emph{{SIGIR} '21: The 44th International {ACM} {SIGIR} Conference
  on Research and Development in Information Retrieval, Virtual Event, Canada,
  July 11-15, 2021}, pages 113--122. {ACM}.

\bibitem[{Houlsby et~al.(2019)Houlsby, Giurgiu, Jastrzebski, Morrone,
  De~Laroussilhe, Gesmundo, Attariyan, and
  Gelly}]{houlsby_2019_param_efficient}
Neil Houlsby, Andrei Giurgiu, Stanislaw Jastrzebski, Bruna Morrone, Quentin
  De~Laroussilhe, Andrea Gesmundo, Mona Attariyan, and Sylvain Gelly. 2019.
\newblock \href {https://proceedings.mlr.press/v97/houlsby19a.html}
  {Parameter-efficient transfer learning for {NLP}}.
\newblock In \emph{International Conference on Machine Learning}, volume~97,
  pages 2790--2799. Proceedings of Machine Learning Research.

\bibitem[{Hu et~al.(2018)Hu, Shen, and Sun}]{hu:squeeze}
Jie Hu, Li~Shen, and Gang Sun. 2018.
\newblock Squeeze-and-excitation networks.
\newblock In \emph{Proceedings of the IEEE conference on computer vision and
  pattern recognition}, pages 7132--7141.

\bibitem[{Kaneko and Bollegala(2021)}]{kaneko_2021_debiasing}
Masahiro Kaneko and Danushka Bollegala. 2021.
\newblock Debiasing pre-trained contextualised embeddings.
\newblock In \emph{Proceedings of the 16th Conference of the European Chapter
  of the Association for Computational Linguistics: Main Volume}, pages
  1256--1266.

\bibitem[{Krieg et~al.(2023)Krieg, Parada-Cabaleiro, Medicus, Lesota, Schedl,
  and Rekabsaz}]{krieg2022grep}
Klara Krieg, Emilia Parada-Cabaleiro, Gertraud Medicus, Oleg Lesota, Markus
  Schedl, and Navid Rekabsaz. 2023.
\newblock Grep-biasir: A dataset for investigating gender representation-bias
  in information retrieval results.
\newblock In \emph{Proceeding of the ACM SIGIR Conference On Human Information
  Interaction And Retrieval (CHIIR)}.

\bibitem[{Kulshrestha et~al.(2017)Kulshrestha, Eslami, Messias, Zafar, Ghosh,
  Gummadi, and Karahalios}]{Kulshrestha:fairness}
Juhi Kulshrestha, Motahhare Eslami, Johnnatan Messias, Muhammad~Bilal Zafar,
  Saptarshi Ghosh, Krishna~P. Gummadi, and Karrie Karahalios. 2017.
\newblock \href {https://doi.org/10.1145/2998181.2998321} {Quantifying search
  bias: Investigating sources of bias for political searches in social media}.
\newblock In \emph{Proceedings of the 2017 {ACM} Conference on Computer
  Supported Cooperative Work and Social Computing, {CSCW} 2017, Portland, OR,
  USA, February 25 - March 1, 2017}, pages 417--432. {ACM}.

\bibitem[{Kumar et~al.(2023)Kumar, Lesota, Zerveas, Cohen, Eickhoff, Schedl,
  and Rekabsaz}]{kumar2023parameter}
Deepak Kumar, Oleg Lesota, George Zerveas, Daniel Cohen, Carsten Eickhoff,
  Markus Schedl, and Navid Rekabsaz. 2023.
\newblock \href {https://aclanthology.org/2023.eacl-main.201}
  {Parameter-efficient modularised bias mitigation via {A}dapter{F}usion}.
\newblock In \emph{Proceedings of the 17th Conference of the European Chapter
  of the Association for Computational Linguistics}, pages 2738--2751,
  Dubrovnik, Croatia. Association for Computational Linguistics.

\bibitem[{Kumar et~al.(2022)Kumar, Paria, and
  Tsvetkov}]{kumar-etal-2022-gradient}
Sachin Kumar, Biswajit Paria, and Yulia Tsvetkov. 2022.
\newblock \href {https://aclanthology.org/2022.emnlp-main.144} {Gradient-based
  constrained sampling from language models}.
\newblock In \emph{Proceedings of the 2022 Conference on Empirical Methods in
  Natural Language Processing}, pages 2251--2277, Abu Dhabi, United Arab
  Emirates. Association for Computational Linguistics.

\bibitem[{Lauscher et~al.(2021)Lauscher, Lueken, and
  Glava{\v{s}}}]{lauscher_2021_sustainable}
Anne Lauscher, Tobias Lueken, and Goran Glava{\v{s}}. 2021.
\newblock \href {https://doi.org/10.18653/v1/2021.findings-emnlp.411}
  {Sustainable modular debiasing of language models}.
\newblock In \emph{Findings of the Association for Computational Linguistics:
  EMNLP 2021}, pages 4782--4797, Punta Cana, Dominican Republic. Association
  for Computational Linguistics.

\bibitem[{Lesota et~al.(2021)Lesota, Rekabsaz, Cohen, Grasserbauer, Eickhoff,
  and Schedl}]{lesota:modern}
Oleg Lesota, Navid Rekabsaz, Daniel Cohen, Klaus~Antonius Grasserbauer, Carsten
  Eickhoff, and Markus Schedl. 2021.
\newblock A modern perspective on query likelihood with deep generative
  retrieval models.
\newblock In \emph{Proceedings of the 2021 ACM SIGIR International Conference
  on Theory of Information Retrieval}, pages 185--195.

\bibitem[{Lian et~al.(2022)Lian, Zhou, Feng, and Wang}]{Lian:SST}
Dongze Lian, Daquan Zhou, Jiashi Feng, and Xinchao Wang. 2022.
\newblock \href
  {http://papers.nips.cc/paper\_files/paper/2022/hash/00bb4e415ef117f2dee2fc3b778d806d-Abstract-Conference.html}
  {Scaling {\&} shifting your features: {A} new baseline for efficient model
  tuning}.
\newblock In \emph{NeurIPS}.

\bibitem[{Liu et~al.(2019)Liu, Ott, Goyal, Du, Joshi, Chen, Levy, Lewis,
  Zettlemoyer, and Stoyanov}]{liu_2019_roberta}
Yinhan Liu, Myle Ott, Naman Goyal, Jingfei Du, Mandar Joshi, Danqi Chen, Omer
  Levy, Mike Lewis, Luke Zettlemoyer, and Veselin Stoyanov. 2019.
\newblock Roberta: {A} robustly optimized bert pretraining approach.
\newblock \emph{ArXiv}, abs/1907.11692.

\bibitem[{Mahabadi et~al.(2021)Mahabadi, Henderson, and
  Ruder}]{DBLP:conf/nips/MahabadiHR21}
Rabeeh~Karimi Mahabadi, James Henderson, and Sebastian Ruder. 2021.
\newblock \href
  {https://proceedings.neurips.cc/paper/2021/hash/081be9fdff07f3bc808f935906ef70c0-Abstract.html}
  {Compacter: Efficient low-rank hypercomplex adapter layers}.
\newblock In \emph{Advances in Neural Information Processing Systems 34: Annual
  Conference on Neural Information Processing Systems 2021, NeurIPS 2021,
  December 6-14, 2021, virtual}, pages 1022--1035.

\bibitem[{Mehrabi et~al.(2022)Mehrabi, Morstatter, Saxena, Lerman, and
  Galstyan}]{mehrabi-bias}
Ninareh Mehrabi, Fred Morstatter, Nripsuta Saxena, Kristina Lerman, and Aram
  Galstyan. 2022.
\newblock \href {https://doi.org/10.1145/3457607} {A survey on bias and
  fairness in machine learning}.
\newblock \emph{{ACM} Comput. Surv.}, 54(6):115:1--115:35.

\bibitem[{Morik et~al.(2021)Morik, Singh, Hong, and
  Joachims}]{morik:ir:fairness}
Marco Morik, Ashudeep Singh, Jessica Hong, and Thorsten Joachims. 2021.
\newblock \href {https://doi.org/10.24963/ijcai.2021/655} {Controlling fairness
  and bias in dynamic learning-to-rank (extended abstract)}.
\newblock In \emph{Proceedings of the Thirtieth International Joint Conference
  on Artificial Intelligence, {IJCAI} 2021, Virtual Event / Montreal, Canada,
  19-27 August 2021}, pages 4804--4808. ijcai.org.

\bibitem[{Nguyen et~al.(2016)Nguyen, Rosenberg, Song, Gao, Tiwary, Majumder,
  and Deng}]{msmarco}
Tri Nguyen, Mir Rosenberg, Xia Song, Jianfeng Gao, Saurabh Tiwary, Rangan
  Majumder, and Li~Deng. 2016.
\newblock \href {https://ceur-ws.org/Vol-1773/CoCoNIPS\_2016\_paper9.pdf} {{MS}
  {MARCO:} {A} human generated machine reading comprehension dataset}.
\newblock In \emph{Proceedings of the Workshop on Cognitive Computation:
  Integrating neural and symbolic approaches 2016 co-located with the 30th
  Annual Conference on Neural Information Processing Systems {(NIPS} 2016),
  Barcelona, Spain, December 9, 2016}, volume 1773 of \emph{{CEUR} Workshop
  Proceedings}. CEUR-WS.org.

\bibitem[{Oosterhuis(2022)}]{oosterhuis:ir:fairness}
Harrie Oosterhuis. 2022.
\newblock \href {https://doi.org/10.24963/ijcai.2022/743} {Computationally
  efficient optimization of plackett-luce ranking models for relevance and
  fairness (extended abstract)}.
\newblock In \emph{Proceedings of the Thirty-First International Joint
  Conference on Artificial Intelligence, {IJCAI} 2022, Vienna, Austria, 23-29
  July 2022}, pages 5319--5323. ijcai.org.

\bibitem[{Papernot et~al.(2021)Papernot, Thakurta, Song, Chien, and
  Erlingsson}]{papernot:tempered_sigmoid}
Nicolas Papernot, Abhradeep Thakurta, Shuang Song, Steve Chien, and {\'U}lfar
  Erlingsson. 2021.
\newblock Tempered sigmoid activations for deep learning with differential
  privacy.
\newblock In \emph{Proceedings of the AAAI Conference on Artificial
  Intelligence}, volume~35, pages 9312--9321.

\bibitem[{Pfeiffer et~al.(2021)Pfeiffer, Kamath, R{\"u}ckl{\'e}, Cho, and
  Gurevych}]{pfeiffer2021adapterfusion}
Jonas Pfeiffer, Aishwarya Kamath, Andreas R{\"u}ckl{\'e}, Kyunghyun Cho, and
  Iryna Gurevych. 2021.
\newblock Adapterfusion: Non-destructive task composition for transfer
  learning.
\newblock In \emph{Proceedings of the 16th Conference of the European Chapter
  of the Association for Computational Linguistics: Main Volume}, pages
  487--503.

\bibitem[{Pfeiffer et~al.(2023)Pfeiffer, Ruder, Vuli{\'c}, and
  Ponti}]{pfeiffer2023modular}
Jonas Pfeiffer, Sebastian Ruder, Ivan Vuli{\'c}, and Edoardo~Maria Ponti. 2023.
\newblock Modular deep learning.
\newblock \emph{arXiv preprint arXiv:2302.11529}.

\bibitem[{Qin et~al.(2022)Qin, Welleck, Khashabi, and Choi}]{qin2022cold}
Lianhui Qin, Sean Welleck, Daniel Khashabi, and Yejin Choi. 2022.
\newblock Cold decoding: Energy-based constrained text generation with langevin
  dynamics.
\newblock In \emph{Advances in Neural Information Processing Systems}.

\bibitem[{Ramachandran et~al.(2017)Ramachandran, Zoph, and
  Le}]{ramachandran:swish}
Prajit Ramachandran, Barret Zoph, and Quoc~V Le. 2017.
\newblock Searching for activation functions.
\newblock \emph{arXiv preprint arXiv:1710.05941}.

\bibitem[{Rangel et~al.(2016)Rangel, Rosso, Verhoeven, Daelemans, Potthast, and
  Stein}]{rangel:pan16}
Francisco Rangel, Paolo Rosso, Ben Verhoeven, Walter Daelemans, Martin
  Potthast, and Benno Stein. 2016.
\newblock Overview of the 4th author profiling task at pan 2016: cross-genre
  evaluations.
\newblock In \emph{Working Notes Papers of the CLEF 2016 Evaluation Labs. CEUR
  Workshop Proceedings/Balog, Krisztian [edit.]; et al.}, pages 750--784.

\bibitem[{Ravfogel et~al.(2020)Ravfogel, Elazar, Gonen, Twiton, and
  Goldberg}]{ravfogel_2020_null}
Shauli Ravfogel, Yanai Elazar, Hila Gonen, Michael Twiton, and Yoav Goldberg.
  2020.
\newblock Null it out: Guarding protected attributes by iterative nullspace
  projection.
\newblock In \emph{Proceedings of the 58th Annual Meeting of the Association
  for Computational Linguistics}, pages 7237--7256.

\bibitem[{Rebuffi et~al.(2017)Rebuffi, Bilen, and
  Vedaldi}]{rebuffi2017learning}
Sylvestre-Alvise Rebuffi, Hakan Bilen, and Andrea Vedaldi. 2017.
\newblock Learning multiple visual domains with residual adapters.
\newblock In \emph{Advances in Neural Information Processing Systems
  (NeurIPS)}, volume~30.

\bibitem[{Rekabsaz et~al.(2021)Rekabsaz, Kopeinik, and
  Schedl}]{rekabsaz_2021_societal}
Navid Rekabsaz, Simone Kopeinik, and Markus Schedl. 2021.
\newblock Societal biases in retrieved contents: Measurement framework and
  adversarial mitigation of {BERT} rankers.
\newblock In \emph{Proceedings of the 44th International ACM SIGIR Conference
  on Research and Development in Information Retrieval}, pages 306--316.

\bibitem[{Rekabsaz and Schedl(2020)}]{rekabsaz_2020_neural}
Navid Rekabsaz and Markus Schedl. 2020.
\newblock Do neural ranking models intensify gender bias?
\newblock In \emph{Proceedings of the 43rd International ACM SIGIR Conference
  on Research and Development in Information Retrieval}, pages 2065--2068.

\bibitem[{Romanov et~al.(2019)Romanov, De-Arteaga, Wallach, Chayes, Borgs,
  Chouldechova, Geyik, Kenthapadi, Rumshisky, and Kalai}]{romanov2019s}
Alexey Romanov, Maria De-Arteaga, Hanna Wallach, Jennifer Chayes, Christian
  Borgs, Alexandra Chouldechova, Sahin Geyik, Krishnaram Kenthapadi, Anna
  Rumshisky, and Adam Kalai. 2019.
\newblock What’s in a name? reducing bias in bios without access to protected
  attributes.
\newblock In \emph{Proceedings of the 2019 Conference of the North American
  Chapter of the Association for Computational Linguistics: Human Language
  Technologies, Volume 1 (Long and Short Papers)}, pages 4187--4195.

\bibitem[{Ross et~al.(2022)Ross, Wu, Peng, Peters, and
  Gardner}]{st:alexis:semantic-control}
Alexis Ross, Tongshuang Wu, Hao Peng, Matthew~E. Peters, and Matt Gardner.
  2022.
\newblock \href {https://doi.org/10.18653/v1/2022.acl-long.228} {Tailor:
  Generating and perturbing text with semantic controls}.
\newblock In \emph{Proceedings of the 60th Annual Meeting of the Association
  for Computational Linguistics (Volume 1: Long Papers), {ACL} 2022, Dublin,
  Ireland, May 22-27, 2022}, pages 3194--3213. Association for Computational
  Linguistics.

\bibitem[{R{\"u}ckl{\'e} et~al.(2021)R{\"u}ckl{\'e}, Geigle, Glockner, Beck,
  Pfeiffer, Reimers, and Gurevych}]{ruckle2021adapterdrop}
Andreas R{\"u}ckl{\'e}, Gregor Geigle, Max Glockner, Tilman Beck, Jonas
  Pfeiffer, Nils Reimers, and Iryna Gurevych. 2021.
\newblock {AdapterDrop}: On the efficiency of adapters in transformers.
\newblock In \emph{Proceedings of the 2021 Conference on Empirical Methods in
  Natural Language Processing}, pages 7930--7946.

\bibitem[{Sanh et~al.(2019)Sanh, Debut, Chaumond, and Wolf}]{sanh:distilbert}
Victor Sanh, Lysandre Debut, Julien Chaumond, and Thomas Wolf. 2019.
\newblock Distilbert, a distilled version of bert: smaller, faster, cheaper and
  lighter.
\newblock \emph{arXiv preprint arXiv:1910.01108}.

\bibitem[{Sap et~al.(2019)Sap, Card, Gabriel, Choi, and Smith}]{sap_2019_risk}
Maarten Sap, Dallas Card, Saadia Gabriel, Yejin Choi, and Noah~A. Smith. 2019.
\newblock \href {https://doi.org/10.18653/v1/P19-1163} {The risk of racial bias
  in hate speech detection}.
\newblock In \emph{Proceedings of the 57th Annual Meeting of the Association
  for Computational Linguistics}, pages 1668--1678, Florence, Italy.
  Association for Computational Linguistics.

\bibitem[{Shen et~al.(2022)Shen, Han, Cohn, Baldwin, and
  Frermann}]{shen-representational}
Aili Shen, Xudong Han, Trevor Cohn, Timothy Baldwin, and Lea Frermann. 2022.
\newblock \href {https://aclanthology.org/2022.findings-aacl.8} {Does
  representational fairness imply empirical fairness?}
\newblock In \emph{Findings of the Association for Computational Linguistics:
  AACL-IJCNLP 2022}, pages 81--95, Online only. Association for Computational
  Linguistics.

\bibitem[{Shen et~al.(2020)Shen, Mueller, Barzilay, and
  Jaakkola}]{st:shen:educating-autoencoder}
Tianxiao Shen, Jonas Mueller, Regina Barzilay, and Tommi~S. Jaakkola. 2020.
\newblock \href {http://proceedings.mlr.press/v119/shen20c.html} {Educating
  text autoencoders: Latent representation guidance via denoising}.
\newblock In \emph{Proceedings of the 37th International Conference on Machine
  Learning, {ICML} 2020, 13-18 July 2020, Virtual Event}, volume 119 of
  \emph{Proceedings of Machine Learning Research}, pages 8719--8729. {PMLR}.

\bibitem[{Sheng et~al.(2019)Sheng, Chang, Natarajan, and Peng}]{sheng2019woman}
Emily Sheng, Kai-Wei Chang, Prem Natarajan, and Nanyun Peng. 2019.
\newblock The woman worked as a babysitter: On biases in language generation.
\newblock In \emph{Proceedings of the 2019 Conference on Empirical Methods in
  Natural Language Processing and the 9th International Joint Conference on
  Natural Language Processing (EMNLP-IJCNLP)}, pages 3398--3403.

\bibitem[{Stanovsky et~al.(2019)Stanovsky, Smith, and
  Zettlemoyer}]{stanovsky2019evaluating}
Gabriel Stanovsky, Noah~A Smith, and Luke Zettlemoyer. 2019.
\newblock Evaluating gender bias in machine translation.
\newblock In \emph{Proceedings of the 57th Annual Meeting of the Association
  for Computational Linguistics}, pages 1679--1684.

\bibitem[{Stickland and Murray(2019)}]{pmlr-v97-stickland19a}
Asa~Cooper Stickland and Iain Murray. 2019.
\newblock \href {https://proceedings.mlr.press/v97/stickland19a.html} {{BERT}
  and {PAL}s: Projected attention layers for efficient adaptation in multi-task
  learning}.
\newblock In \emph{Proceedings of the 36th International Conference on Machine
  Learning}, volume~97 of \emph{Proceedings of Machine Learning Research},
  pages 5986--5995. PMLR.

\bibitem[{Subramani et~al.(2022)Subramani, Suresh, and
  Peters}]{st:nishant:steering-vector}
Nishant Subramani, Nivedita Suresh, and Matthew~E. Peters. 2022.
\newblock \href {https://doi.org/10.18653/v1/2022.findings-acl.48} {Extracting
  latent steering vectors from pretrained language models}.
\newblock In \emph{Findings of the Association for Computational Linguistics:
  {ACL} 2022, Dublin, Ireland, May 22-27, 2022}, pages 566--581. Association
  for Computational Linguistics.

\bibitem[{Turc et~al.(2019)Turc, Chang, Lee, and
  Toutanova}]{turc_2019_bert_small}
Iulia Turc, Ming-Wei Chang, Kenton Lee, and Kristina Toutanova. 2019.
\newblock Well-read students learn better: On the importance of pre-training
  compact models.
\newblock \emph{arXiv preprint arXiv:1908.08962v2}.

\bibitem[{Vaswani et~al.(2017)Vaswani, Shazeer, Parmar, Uszkoreit, Jones,
  Gomez, Kaiser, and Polosukhin}]{vaswani_attention_NIPS2017}
Ashish Vaswani, Noam Shazeer, Niki Parmar, Jakob Uszkoreit, Llion Jones,
  Aidan~N Gomez, \L~ukasz Kaiser, and Illia Polosukhin. 2017.
\newblock \href
  {https://proceedings.neurips.cc/paper/2017/file/3f5ee243547dee91fbd053c1c4a845aa-Paper.pdf}
  {Attention is all you need}.
\newblock In \emph{Advances in Neural Information Processing Systems},
  volume~30, pages 6000--–6010. Curran Associates, Inc.

\bibitem[{Wang et~al.(2021)Wang, Yan, He, Wu, and Xu}]{wang_2021_dynamically}
Liwen Wang, Yuanmeng Yan, Keqing He, Yanan Wu, and Weiran Xu. 2021.
\newblock Dynamically disentangling social bias from task-oriented
  representations with adversarial attack.
\newblock In \emph{Proceedings of the 2021 Conference of the North American
  Chapter of the Association for Computational Linguistics: Human Language
  Technologies}, pages 3740--3750.

\bibitem[{Xu et~al.(2020)Xu, Desai, and Durrett}]{xu2020understanding}
Jiacheng Xu, Shrey Desai, and Greg Durrett. 2020.
\newblock Understanding neural abstractive summarization models via
  uncertainty.
\newblock In \emph{Proceedings of the 2020 Conference on Empirical Methods in
  Natural Language Processing (EMNLP)}, pages 6275--6281.

\bibitem[{Yu et~al.(2021)Yu, Yu, and Sagae}]{yu2021attribute}
Dian Yu, Zhou Yu, and Kenji Sagae. 2021.
\newblock Attribute alignment: Controlling text generation from pre-trained
  language models.
\newblock In \emph{Findings of the Association for Computational Linguistics:
  EMNLP 2021}, pages 2251--2268.

\bibitem[{Zerveas et~al.(2022{\natexlab{a}})Zerveas, Rekabsaz, Cohen, and
  Eickhoff}]{zerveas:coder}
George Zerveas, Navid Rekabsaz, Daniel Cohen, and Carsten Eickhoff.
  2022{\natexlab{a}}.
\newblock Coder: An efficient framework for improving retrieval through
  contextual document embedding reranking.
\newblock In \emph{Proceedings of the 2022 Conference on Empirical Methods in
  Natural Language Processing}, pages 10626--10644.

\bibitem[{Zerveas et~al.(2022{\natexlab{b}})Zerveas, Rekabsaz, Cohen, and
  Eickhoff}]{zerveas_2022_mitigating}
George Zerveas, Navid Rekabsaz, Daniel Cohen, and Carsten Eickhoff.
  2022{\natexlab{b}}.
\newblock \href {https://doi.org/10.1145/3477495.3531891} {Mitigating bias in
  search results through contextual document reranking and neutrality
  regularization}.
\newblock In \emph{Proceedings of the 45th International ACM SIGIR Conference
  on Research and Development in Information Retrieval}, SIGIR '22, page
  2532–2538, New York, NY, USA. Association for Computing Machinery.

\bibitem[{Zhang et~al.(2022)Zhang, Song, Li, Zhou, and Song}]{zhang2022survey}
Hanqing Zhang, Haolin Song, Shaoyu Li, Ming Zhou, and Dawei Song. 2022.
\newblock A survey of controllable text generation using transformer-based
  pre-trained language models.
\newblock \emph{arXiv preprint arXiv:2201.05337}.

\bibitem[{Zhang et~al.(2021)Zhang, van~de Meent, and
  Wallace}]{zhang_2021_disentangling}
Xiongyi Zhang, Jan-Willem van~de Meent, and Byron Wallace. 2021.
\newblock \href {https://doi.org/10.18653/v1/2021.emnlp-main.60} {Disentangling
  representations of text by masking transformers}.
\newblock In \emph{Proceedings of the 2021 Conference on Empirical Methods in
  Natural Language Processing}, pages 778--791, Online and Punta Cana,
  Dominican Republic. Association for Computational Linguistics.

\bibitem[{Zhao et~al.(2019)Zhao, Wang, Yatskar, Cotterell, Ordonez, and
  Chang}]{zhao2019gender}
Jieyu Zhao, Tianlu Wang, Mark Yatskar, Ryan Cotterell, Vicente Ordonez, and
  Kai-Wei Chang. 2019.
\newblock Gender bias in contextualized word embeddings.
\newblock In \emph{Proceedings of the Conference of the North American Chapter
  of the Association for Computational Linguistics: Human Language
  Technologies}, pages 629--634.

\bibitem[{Zhao et~al.(2020)Zhao, Lin, Mi, Jaggi, and
  Sch{\"u}tze}]{zhao_2020_masking}
Mengjie Zhao, Tao Lin, Fei Mi, Martin Jaggi, and Hinrich Sch{\"u}tze. 2020.
\newblock \href {https://doi.org/10.18653/v1/2020.emnlp-main.174} {Masking as
  an efficient alternative to finetuning for pretrained language models}.
\newblock In \emph{Empirical Methods in Natural Language Processing}, pages
  2226--2241. Association for Computational Linguistics.

\end{thebibliography}

\clearpage

\appendix

\section{Multi-Concept \modelours}
\label{sec:appendix:multi-attribute}

\begin{figure}[t]
\center
\includegraphics[width=\columnwidth]{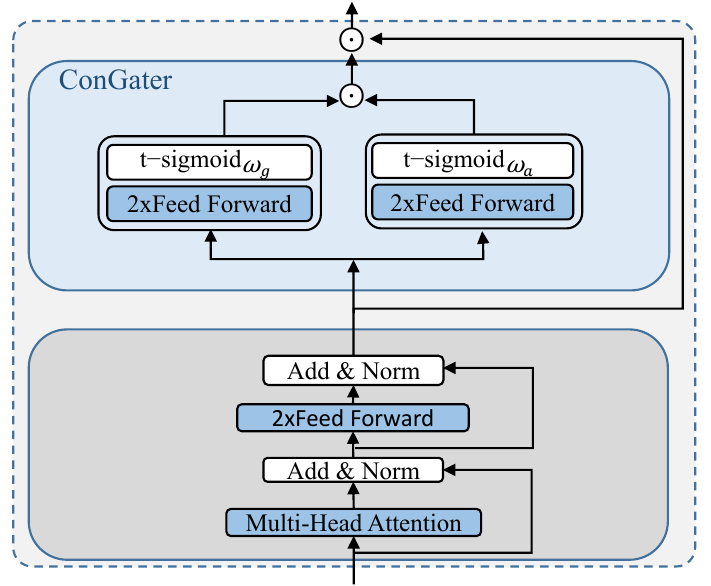} 
\caption{The simple proposed \modelours for multi-attributes. The fusion gate is defined as the element-wise multiplication of the individual gates
}
\label{fig:appendix:congater_multi-attribute}
\end{figure}

In this section, we investigate a version of \modelours with multi-attribute. Figure~\ref{fig:appendix:congater_multi-attribute} depicts this variation, where the individual gates for age and gender are combined with element-wise multiplication to form a multi-attribute gating vector. This vector is then used for the self-gating mechanism of \modelours in Eq.~\ref{eq:CRAG}. The training procedure of the multi-attribute setting is exactly the same as the single-attribute one. During inference, the gating sensitivity of each attribute can be changed independently. Our experiment results on a two-attribute setting are provided in Appendix~\ref{sec:appendix:results:multi-attribute}.

\begin{table*}[ht]
    \centering
    \caption{Summary of the datasets and their protected attribute(s)}
    \small
    \begin{tabular}{lccrrr}
        \toprule
         Dataset & Classes & Attribute & Train & Validation & Test  \\\midrule
         FCDL18 & 4 & Dialect & 52,352 & 4,736 & 5,888 \\
         BIOS & 28 & Gender & 294,784 & 38,016 & 88,640 \\
         PAN16 & 2 & Gender\&age & 160,000 & 10,048 & 30,016  \\
         \bottomrule
    \end{tabular}%
    
    \label{tab:datasets}
\end{table*}

\begin{table*}[ht]
\centering
\caption{\modeladapter and \modelours Hyperparameters used to fine tune and remove information from the network on different datasets}
\small
\begin{tabular}{lllll}
\toprule
\multicolumn{5}{c}{Dataset}   \\
\midrule
 & FCDL18 & PAN16 & BIOS & MSMARCO \\
\midrule
\multicolumn{5}{c}{Embedding} \\
\midrule
batch size & 64 & 64 & 64 & 64 \\
number of workers & 8 & 8 & 8 & 5 \\
Max document/query length & 40 & 40 & 120 & 32 \\
Padding & max length & max length & max length & max length \\
\midrule
 \multicolumn{5}{c}{Model \& Training} \\ 
\midrule
Model Type & Base/Mini/Roberta & Base/Mini/Roberta & Base/Mini/Roberta & DistillBERT \\
\modelours Bottleneck factor &  8 & 12 & 2 & 4\\ 
$\lambda$ & 1 & 1 & 1  & max = 20\\
$\lambda$ warm-up scheduler & 3 & 3 & 3 & - \\
loss function & Cross Entropy & Cross Entropy & Cross Entropy & ListNet loss\\
Optimizer & AdamW & AdamW & AdamW & RAdam \\
task lr & $2\times10^{-5}$ & $2\times10^{-5}$ & $2\times10^{-5}$ &$1.7\times10^{-6}$ \\
weight decay & 0.01 & 0.01 & 0.01 \\
adv lr & $1\times10^{-4}$ & $1\times10^{-4}$ & $1\times10^{-4}$ & - \\
probe lr & $1\times10^{-4}$ & $1\times10^{-4}$ & $1\times10^{-4}$& - \\
task/adv dropout & 0.1 & 0.1 & 0.1 & 0.0\\
lr scheduler & Cosine Decay & Cosine Decay & Cosine Decay & - \\
train epochs & 15 & 15 & 15 & 10 \\
adv epochs & 15 & 15 & 15 & 10 \\
probe epochs & 30 & 30 & 30 & - \\
task head layer & 1 & 1 & 1 & - \\
adv head layer & 2 & 2 & 2 & -  \\
probe head & 2 & 2 & 2 & - \\
adv/probe activation & Tanh & Tanh & Tanh & -\\
\bottomrule
\end{tabular}%

\label{tbl:hyperparameters}
\end{table*}

\begin{table*}[t!]
\small
\centering
\caption{BERT-base number of parameters specification for each method}
\begin{tabular}{lccc}
\toprule
Parameter Count & \modelfine & \modeladapter & \modelours \\ 
\midrule
Total Number of Parameter & 109,485,316 & 116,577,028 & 116,577,028  \\
Attribute(single) Parameters & 109,485,316 & 7,094,788 & 7,094,788 \\ 
Adversarial Training Module (\%) & 100 & 6.0 & 6.0 \\
\bottomrule
\end{tabular}

\label{tbl:parameters}
\end{table*}

\section{Additional Experiment Setup}
\label{sec:appendix:experiment}

\begin{figure*}
     \centering
     \begin{subfigure}[b]{0.49\columnwidth}
         \centering
         \includegraphics[width=\textwidth]{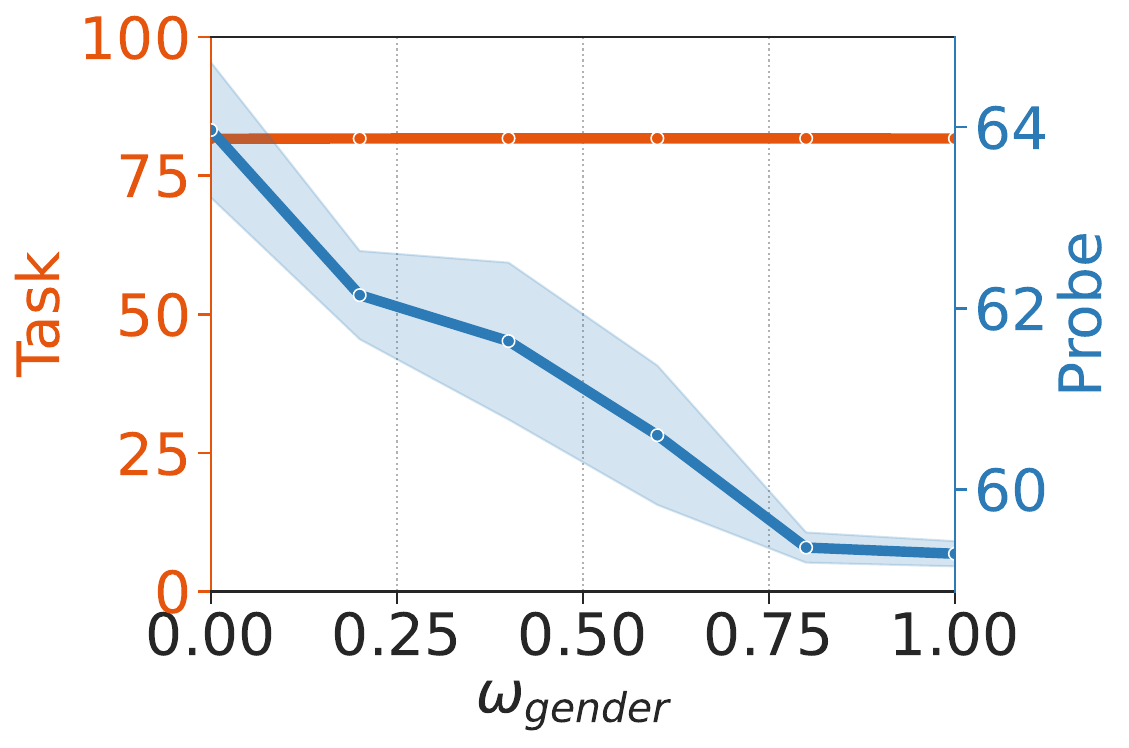}
         \caption{BIOS (Gender)}
     \end{subfigure}
     \hfill
     \begin{subfigure}[b]{0.49\columnwidth}
         \centering
         \includegraphics[width=\textwidth]{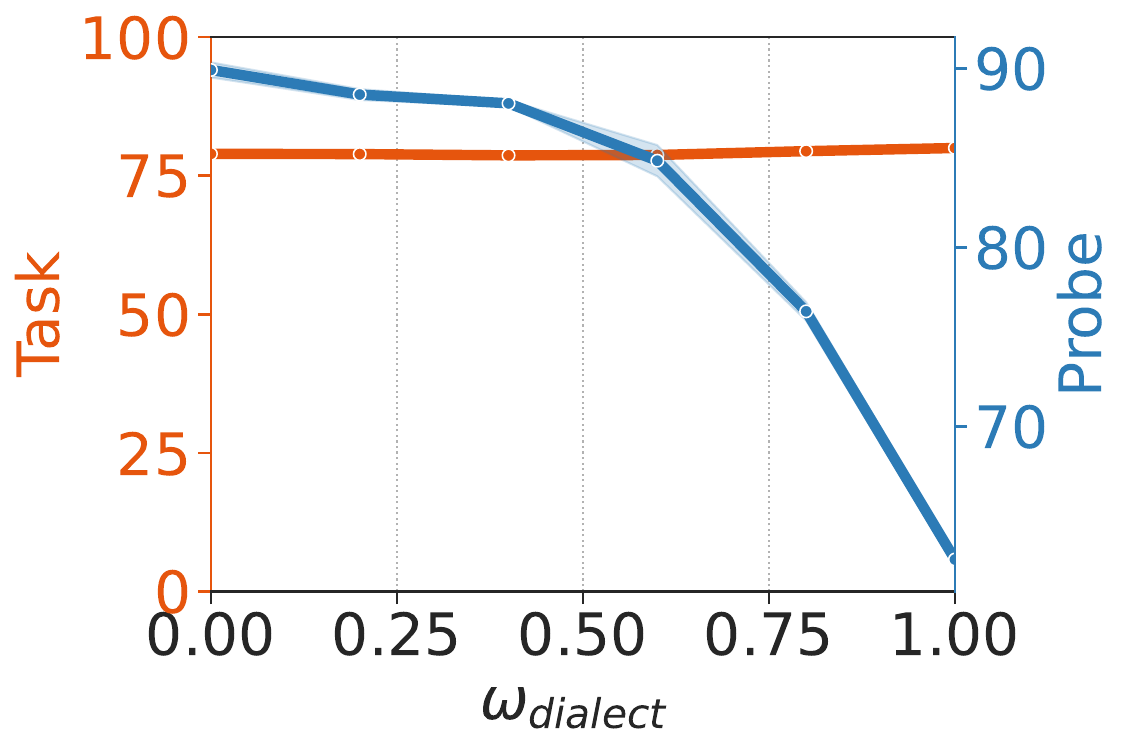}
         \caption{FCDL18 (Dialect-race)}
     \end{subfigure}
     \hfill
     \begin{subfigure}[b]{0.49\columnwidth}
         \centering
         \includegraphics[width=\textwidth]{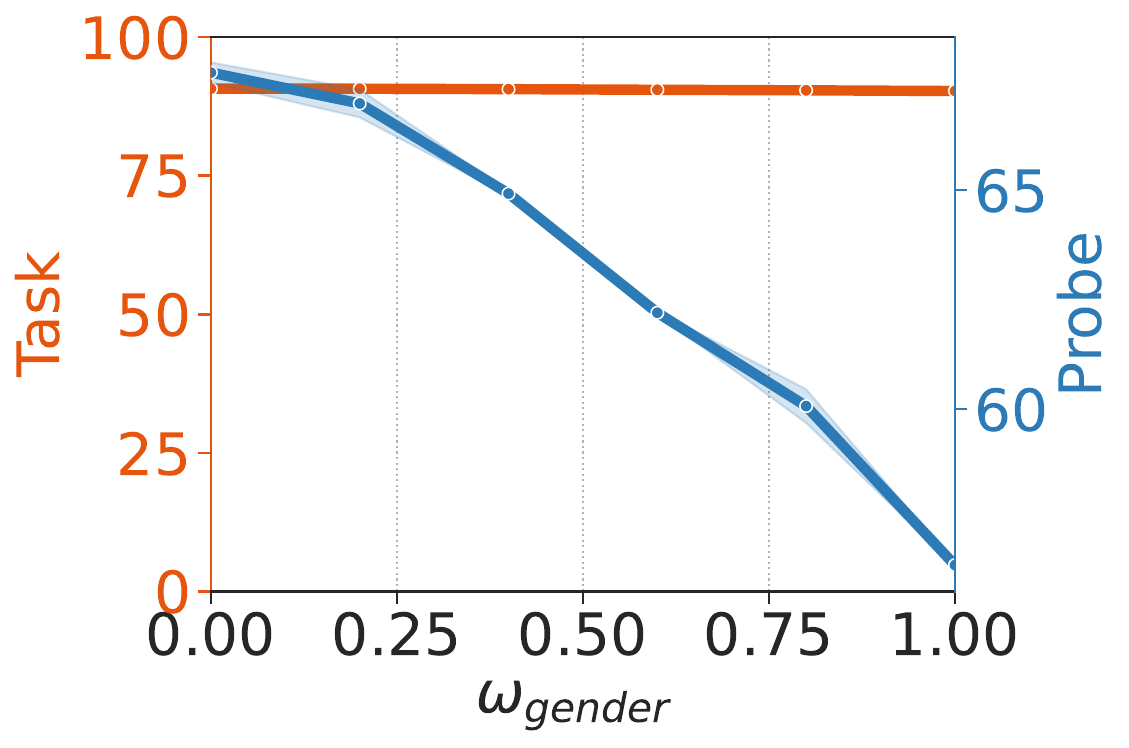}
         \caption{PAN16 - Gender}
     \end{subfigure}
     \hfill
     \begin{subfigure}[b]{0.49\columnwidth}
         \centering
         \includegraphics[width=\textwidth]{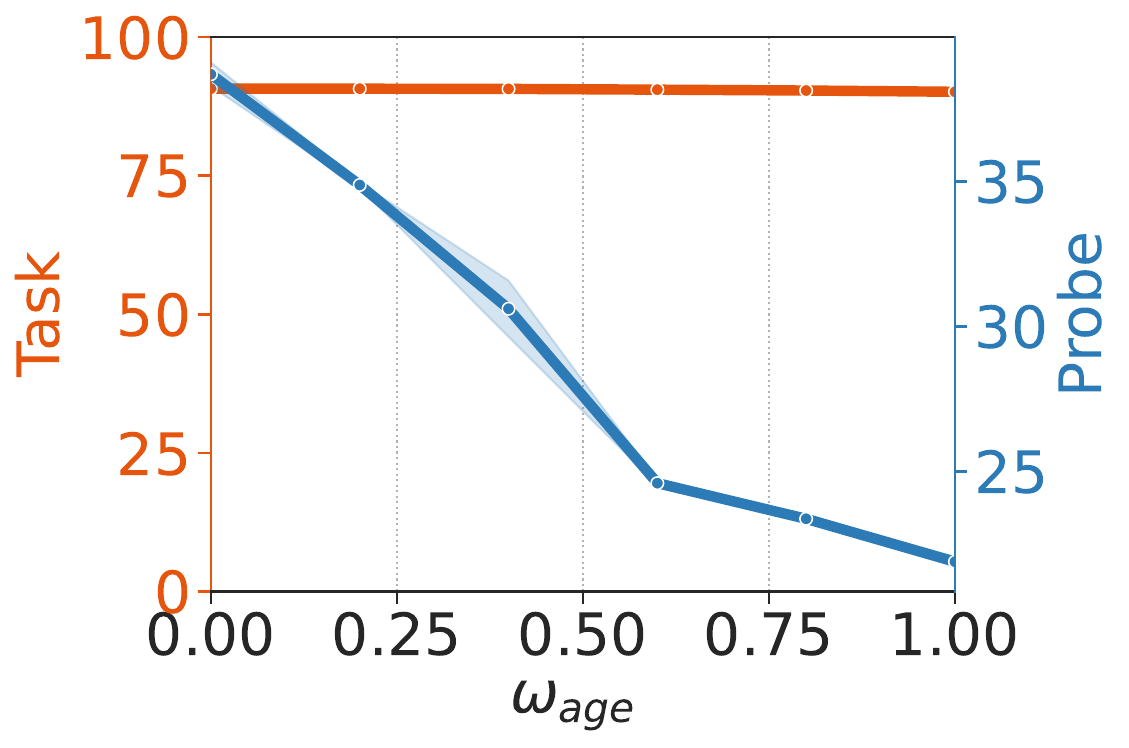}
         \caption{PAN16 - Age}
     \end{subfigure}
     \vspace{-3mm}
     \caption{Results of the \modelours models using BERT-Mini. See Figure~\ref{fig:results:probes} for full explanation.
     }
    \label{fig:appendix:results:probes-mini}    
\end{figure*}

\begin{figure*}
     \centering
     \begin{subfigure}[b]{0.49\columnwidth}
         \centering
         \includegraphics[width=\textwidth]{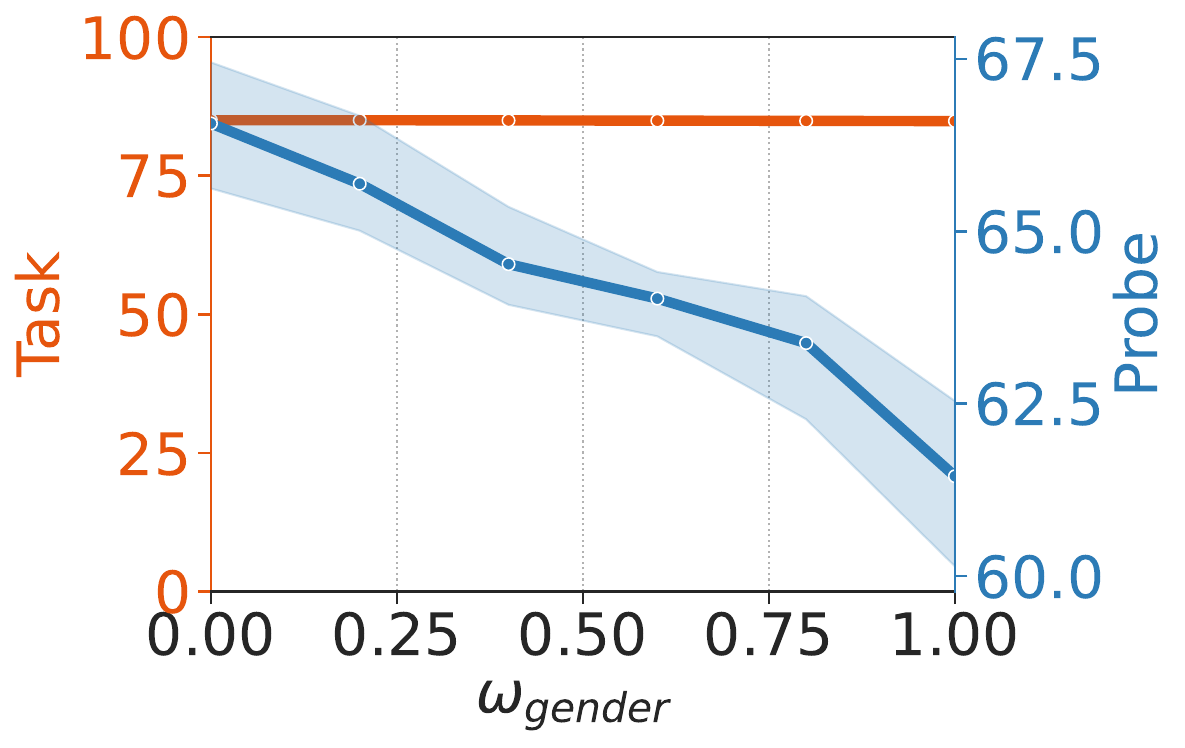}
         \caption{BIOS (Gender)}
     \end{subfigure}
     \hfill
     \begin{subfigure}[b]{0.49\columnwidth}
         \centering
         \includegraphics[width=\textwidth]{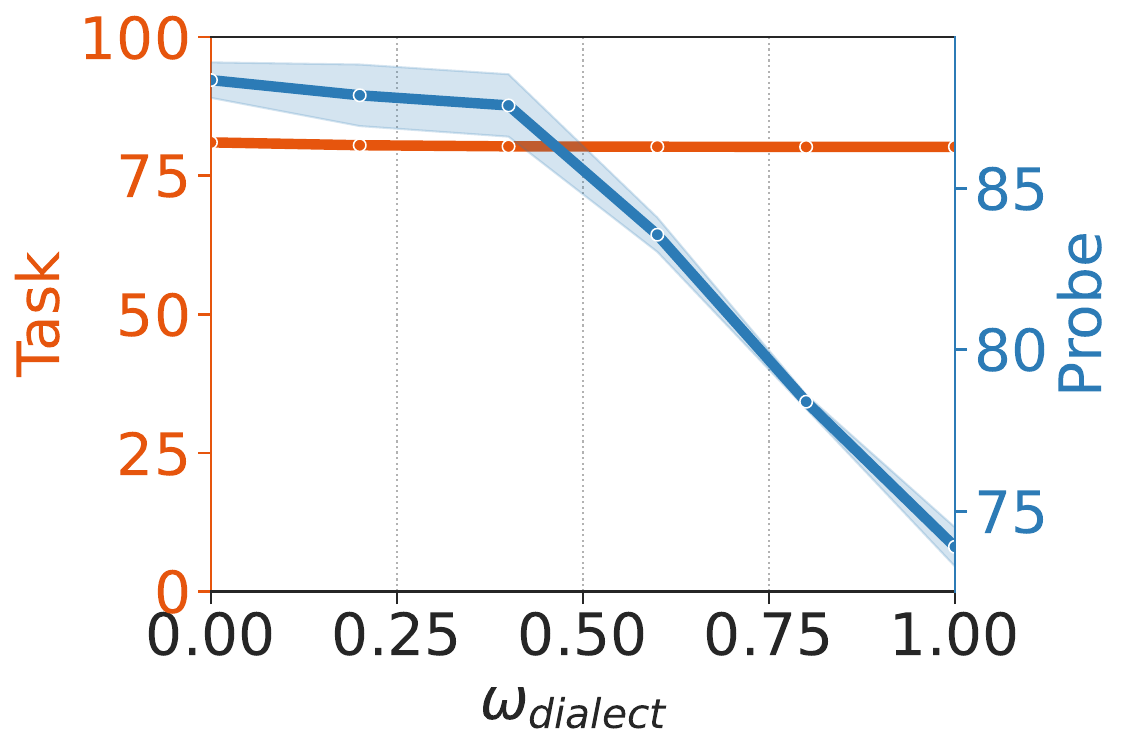}
         \caption{FCDL18 (Dialect-race)}
     \end{subfigure}
     \hfill
     \begin{subfigure}[b]{0.49\columnwidth}
         \centering
         \includegraphics[width=\textwidth]{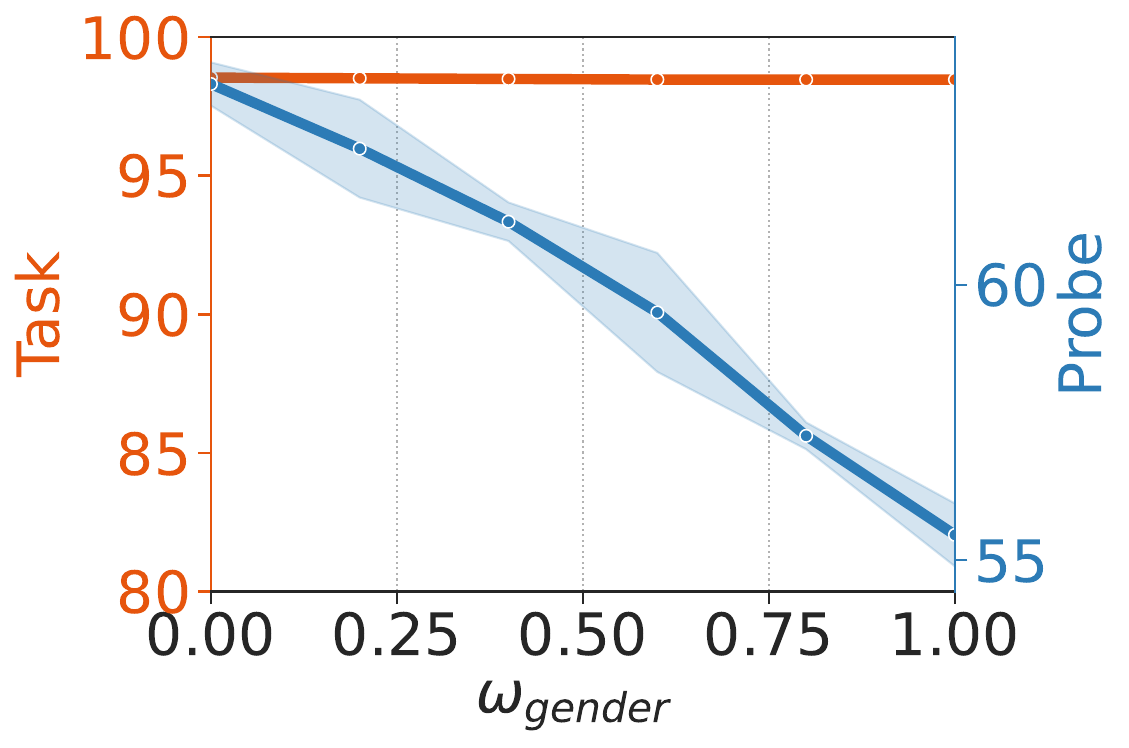}
         \caption{PAN16 - Gender}
     \end{subfigure}
     \hfill
     \begin{subfigure}[b]{0.49\columnwidth}
         \centering
         \includegraphics[width=\textwidth]{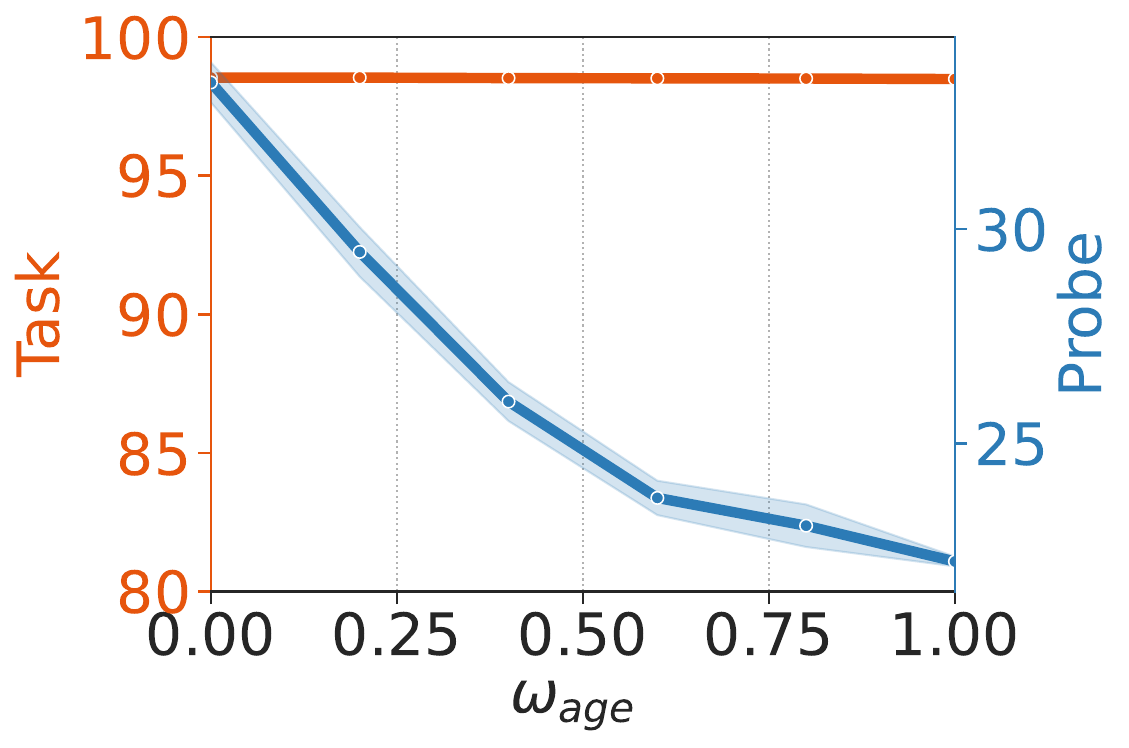}
         \caption{PAN16 - Age}
     \end{subfigure}
     \vspace{-3mm}
     \caption{Results of the \modelours models using RoBERTa-Base. See Figure~\ref{fig:results:probes} for full explanation.}
    \label{fig:appendix:results:probes-roberta}    
\end{figure*}

\begin{figure*}[t]
\begin{subfigure}[t]{0.24\textwidth}
    \includegraphics[width=\columnwidth]{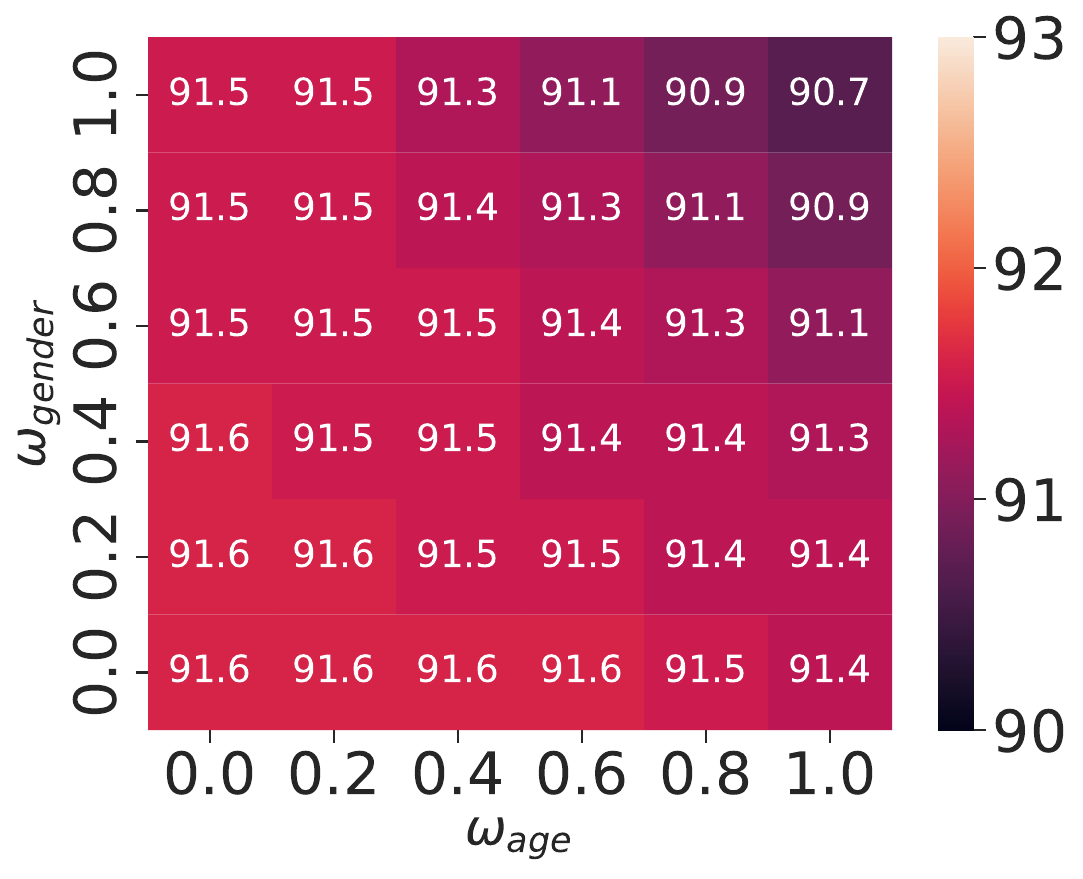}
    \caption{Task Accuracy}
\end{subfigure}
\hfill
\begin{subfigure}[t]{0.25\textwidth}
    \includegraphics[width=\columnwidth]{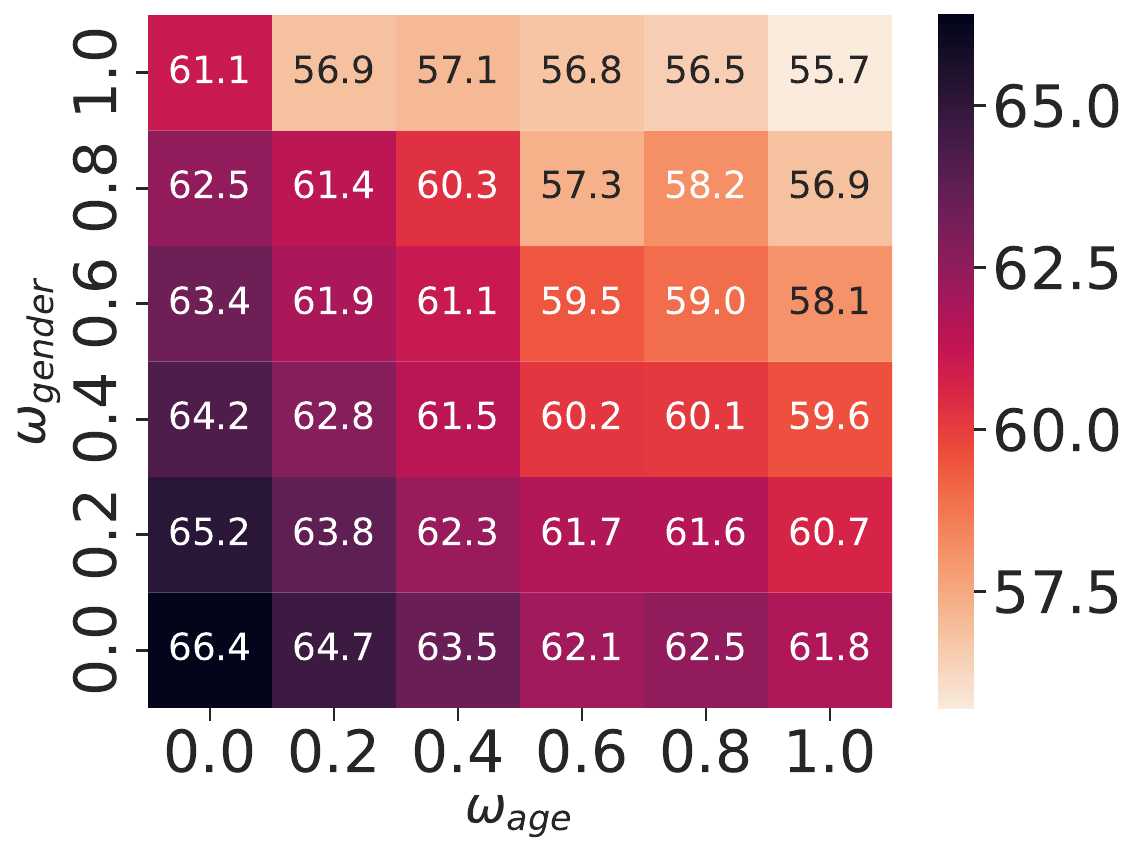}
    \caption{Gender Probes}
\end{subfigure}
\hfill
\begin{subfigure}[t]{0.25\textwidth}
    \includegraphics[width=\columnwidth]{{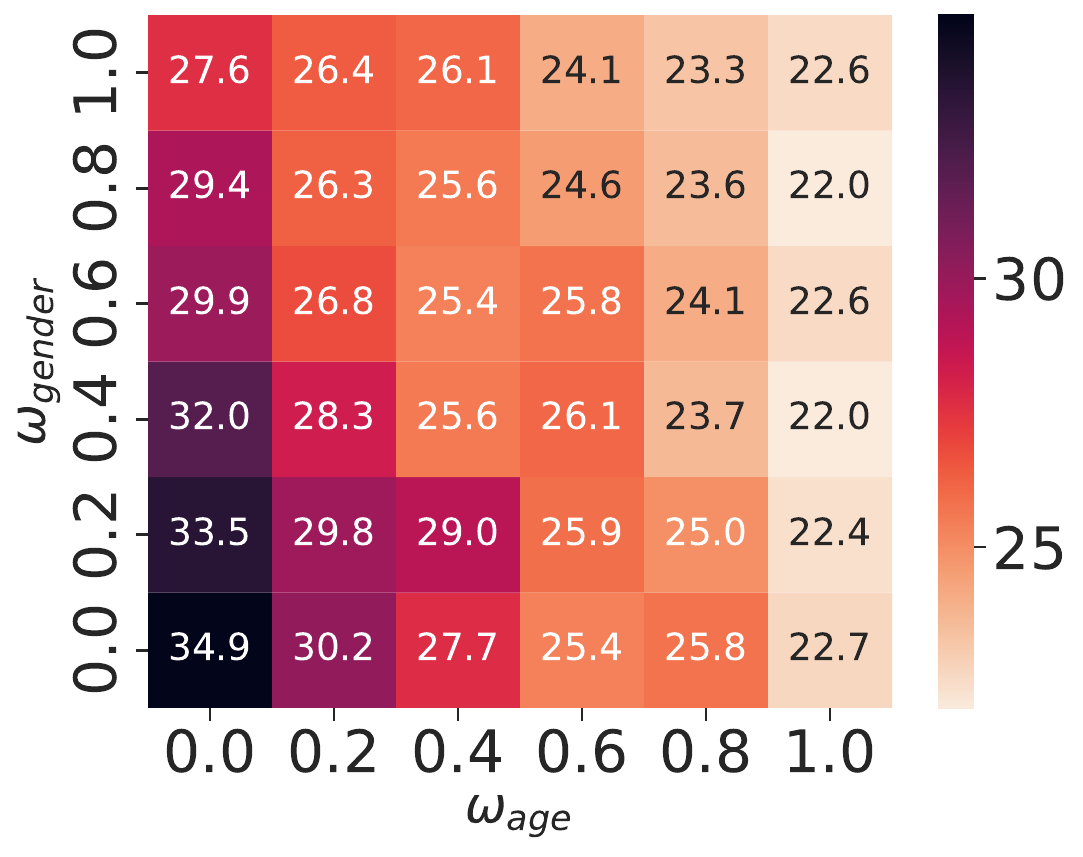}}
    \caption{Age Probes}
\end{subfigure}
\caption{Results of the multi-concept \modelours on the two attributes of PAN16 task using BERT-Mini, as changing the gender and age $\omega$ simultaneously. (a) Task performance with accuracy (\%). (b) Balanced accuracy (\%) of gender probes. (c) Balanced accuracy (\%) of age probes.}
\label{fig:appendix:multi-attribute}
\end{figure*}

In the FDCL18 dataset, we use the TwitterAAE model~\cite{blodgett_2016_demographic} to assign racial dialect classes. The TwitterAAE model predicts four racial classes, \emph{African American}, \emph{White American}, \emph{Hispanic}, and \emph{Others}. We labeled a tweet as \emph{African American} or \emph{White American} if the prediction score was greater than $0.5$. For the PAN16 dataset, following \cite{sap_2019_risk} we balanced the task labels and sampled 200K data. The age groups of this dataset are 18-24, 25-34, 35-49, 50-64, and 65+. Table~\ref{tab:datasets} gives a summary of the whole dataset, training, validation and test data which was used during the training and evaluation of the network.

We select train, validation, and test sets randomly from the pool of data with the proportions 63:12:15 for BIOS, 63:12:15 for FDCL18, and 80:5:15 for PAN16. We use the validation set for hyperparameter tuning, and the best result on the validation set is evaluated on the test set for the final results. The validation and test sets in all datasets follow the same distribution as the whole dataset. To address the unbalanced dataset and the potential problems in adversarial training, we apply upsampling only on the \emph{training sets} of BIOS and FDCL18 datasets, to balance the protected attribute labels within each task label. For instance, genders are balanced in the dentist class by repeating the data items of the minority subgroup.

Models are trained on the task for 15 epochs and for the post-hoc models an additional 15 epochs of adversarial training. Details of hyperparameters are reported in Table~\ref{tbl:hyperparameters} and the number of parameters of the models is reported in Table~\ref{tbl:parameters}.

Each layer that is mentioned in the hyperparameter section has the same width as the original Bert-Base model which in our experiment is (768). In our experiment we trained all of the transformer blocks but in general, any training method on the task completely depends on the designer's will and what \modelours offers is an extension to the original model with additional training which leads to controllability.

\begin{figure*}[t]
\begin{subfigure}[t]{0.3\textwidth}
    \includegraphics[width=\columnwidth]{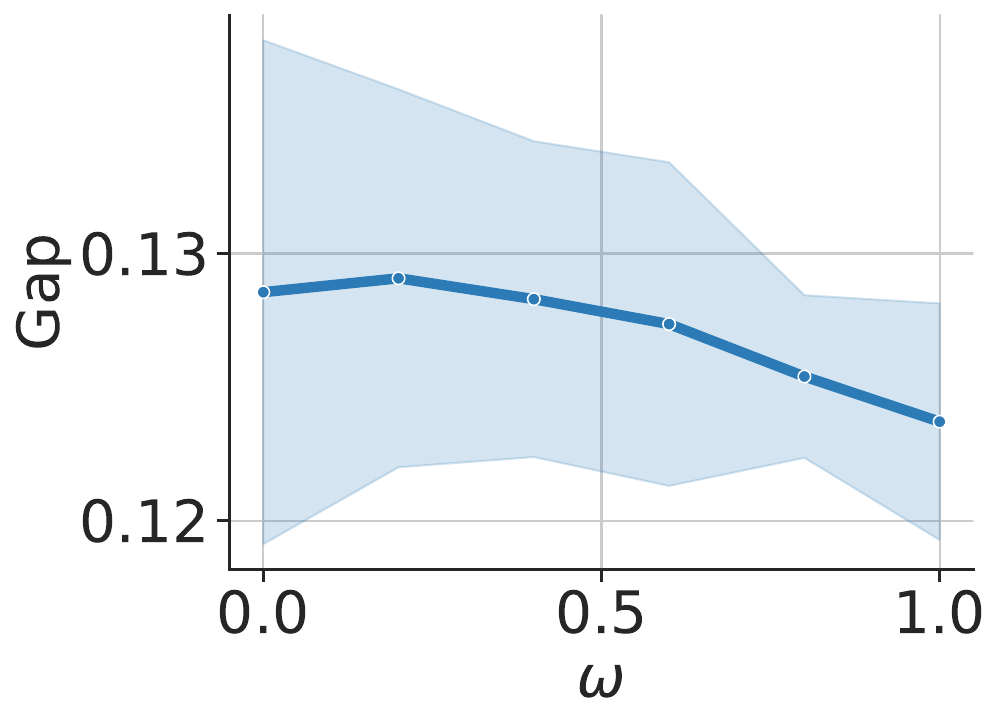}
    \caption{BIOS}
\end{subfigure}
\hfill
\begin{subfigure}[t]{0.3\textwidth}
    \includegraphics[width=\columnwidth]{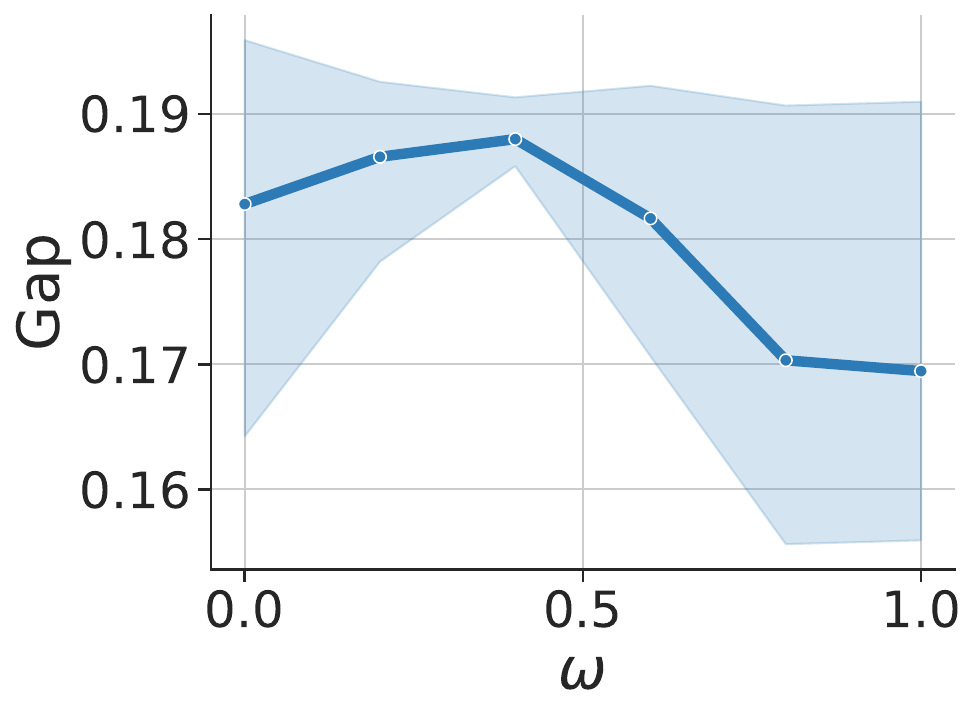}
    \caption{FCDL18}
\end{subfigure}
\hfill
\begin{subfigure}[t]{0.3\textwidth}
    \includegraphics[width=\columnwidth]{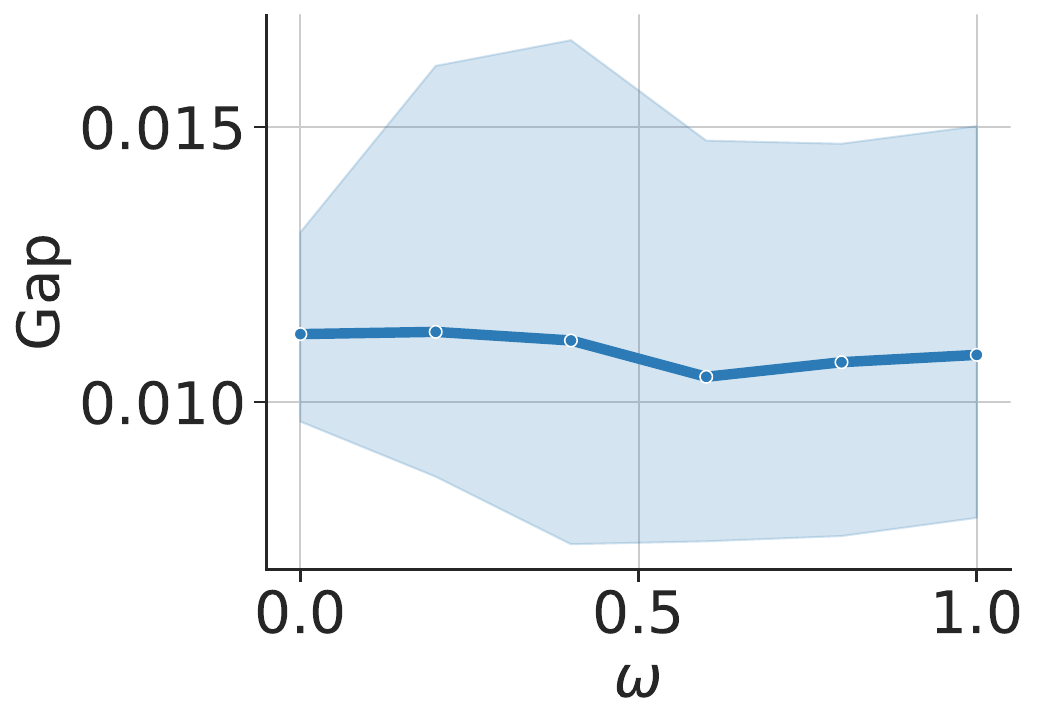}
    \caption{PAN16}
\end{subfigure}
\caption{Mean and standard deviation of Gap metric for Three independent BERT-base models with parallel training method}
\label{fig:appendix:gap}
\end{figure*}

\section{Additional Results}
\label{sec:appendix:results}

\subsection{Other LMs}
\label{sec:appendix:results:mini-roberta}

Figures~\ref{fig:appendix:results:probes-mini} and ~\ref{fig:appendix:results:probes-roberta} show the results using BERT-Mini and RoBERTa-Base LMs, respectively. We observe the same control capability of attribute as we discussed in the main paper for the other to LMs as well.

\subsection{Multi-Attribute Results}
\label{sec:appendix:results:multi-attribute}

To examine the ability of multi-concept \modelours (introduced in Appendix~\ref{sec:appendix:multi-attribute}), we train the model on the two attributes of PAN16, where each concept has its gating sensitivity parameter ($\omega_{gender}$ and $\omega_{age}$). The evaluation results on task performance, gender probing, and age probing are reported in Figure~\ref{fig:appendix:multi-attribute} for a specific combination of $\omega_{gender}$ and $\omega_{age}$. As shown, the multi-concept \modelours model can maintain the task performance, as $\omega$ values change, while the presence of concept information gradually decreases. We also observe that changing one $\omega$ has an influence on the probing results of the other concept, indicating the probable correlations between the concepts. By simultaneously increasing both $\omega$ there exist combinations where information about both gender and age has the minimum value (\eg $\omega_{age}=1$ and $\omega_{gender}=0.1$). This initial experiment shows the benefits of \modelours for multi-attribute control, as well as the challenges in this area, suggesting further investigations for future work.

\begin{figure*}[t]
     \centering
     \begin{subfigure}[t]{0.49\columnwidth}
         \centering
         \includegraphics[width=\textwidth]{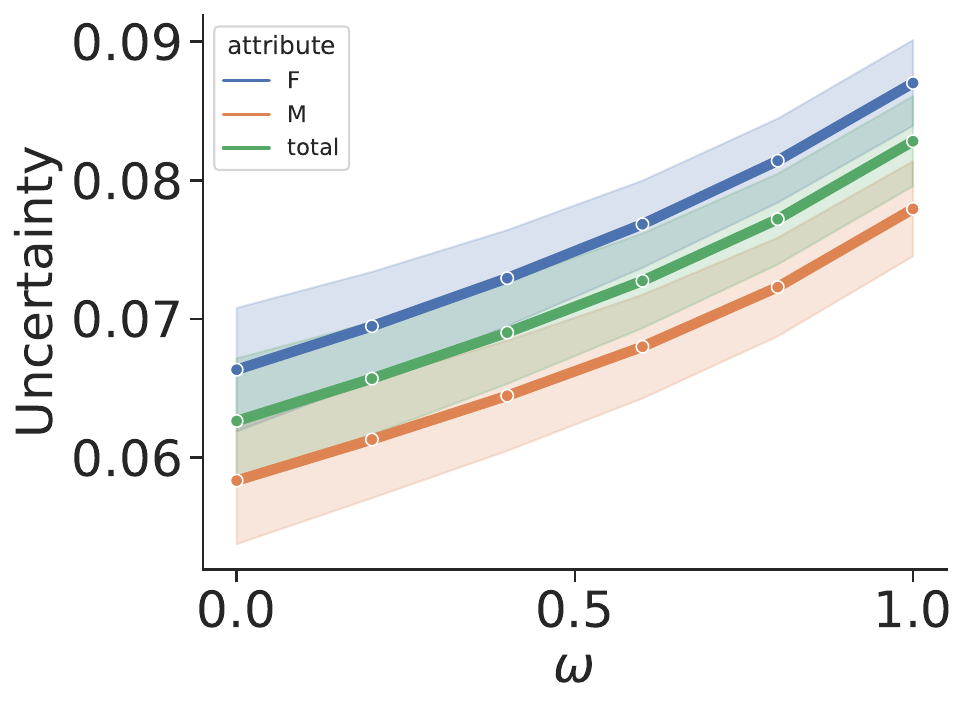}
         \caption{BIOS}
     \end{subfigure}
     \hfill
     \begin{subfigure}[t]{0.49\columnwidth}
         \centering
         \includegraphics[width=\textwidth]{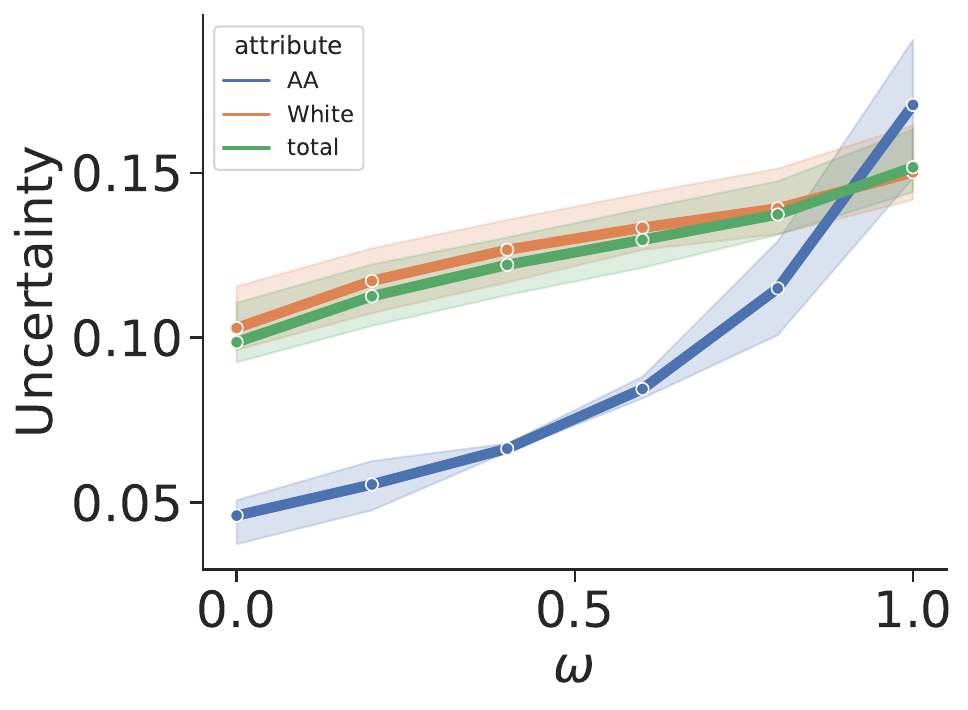}
         \caption{FCDL18}
     \end{subfigure}
     \hfill
     \begin{subfigure}[t]{0.49\columnwidth}
         \centering
         \includegraphics[width=\textwidth]{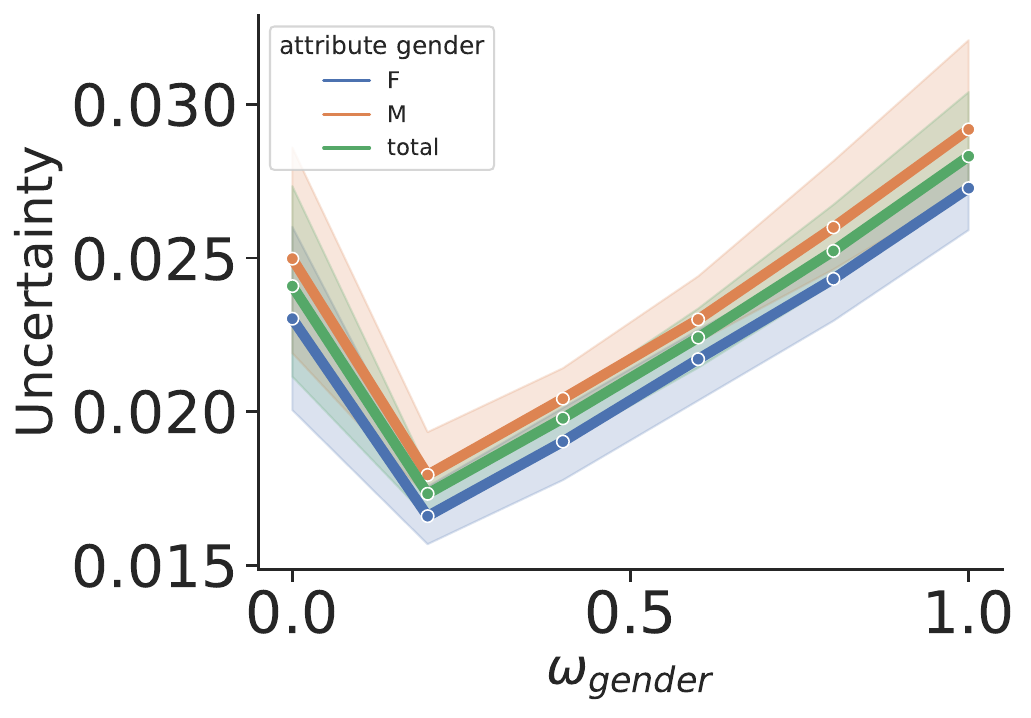}
         \caption{PAN16-Gender}
     \end{subfigure}
     \hfill
     \begin{subfigure}[t]{0.49\columnwidth}
         \centering
         \includegraphics[width=\textwidth]{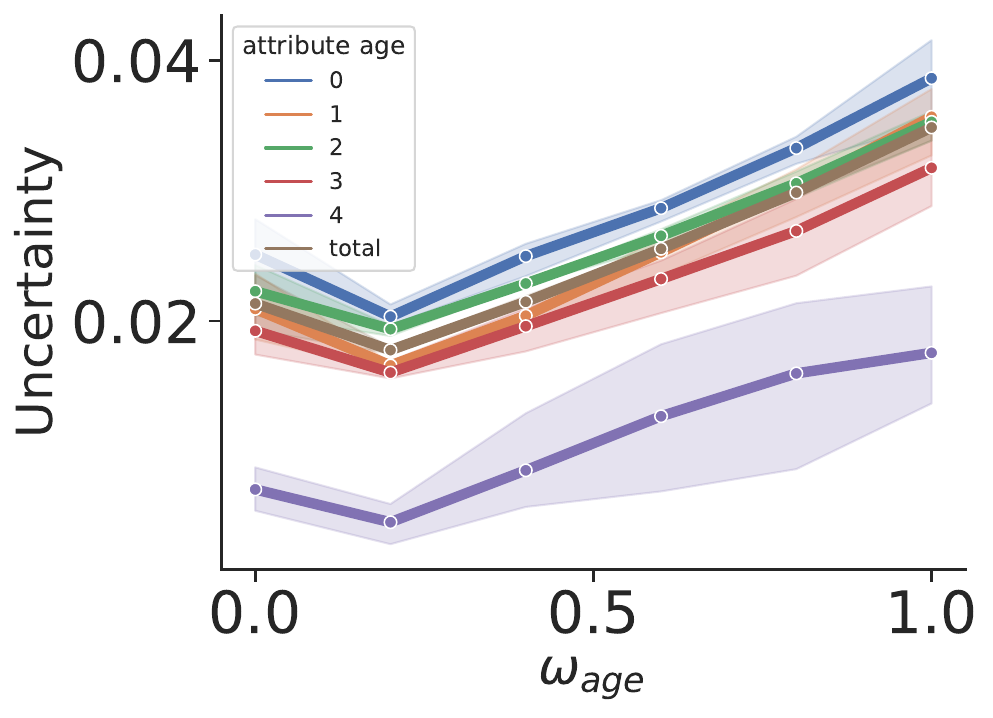}
         \caption{Pan16-Age}
     \end{subfigure}
     \caption{Mean of the test data points' uncertainty values, defined as the entropy of the predicted probability distributions provided by the \modelours models. Increasing sensitivity parameter(s) results in continuous changes in model uncertainties.}
    \label{fig:appendix:results:uncertainty}
    \vspace{-4mm}
\end{figure*}

\subsection{Study of Fairness}
\label{sec:appendix:results:fairness}

Removing attributes has been the focus of studies for several purposes such as bias mitigation, privacy preservation, and fairness improvement. Researchers focused on removing harmful information such as gender or race from the network as the main cause of societal biases to improve fairness with regard to minority groups~\cite{han-bias,mehrabi-bias,shen-representational}. In this section, we investigate the effect of \modelours adversarial training with regard to empirical fairness metrics. In particular, we utilize GAP~\cite{shen-representational} as the evaluation metric of empirical fairness. The GAP metric for a binary attribute is defined as:
\begin{equation}
    GAP = \sqrt{\frac{1}{|Y|}\sum_{y \in Y}(GAP_{a,y}^{TPR})^2}
\label{eq:appendix:gap}
\end{equation}
where $GAP_{a,y}^{TPR}$ is the difference between the True Positive Rate (TPR) for each class $a$, and GAP is the normalized difference between binary sub-populations $Y$. 

Figure~\ref{fig:appendix:gap} shows the mean and standard deviation of the GAP results for three independent-trained runs of the \modelours models using the BERT-Base. The results are overall consistent with the core message in previous studies~\cite{shen-representational}, indicating that removing concept information (known as representational fairness) is not necessarily correlated with GAP (empirical fairness). We however observe that the average GAP for BIOS and FCDL18 slightly decrease, where similar to the probing results, the changes appear continuously between the initial to the target state. Overall, the continuous controllability of \modelours allows for the choice of the state with the desired representational or empirical fairness, given the context of the task and/or a user's preference.

\subsection{Shift in Model Uncertainty}
\label{sec:appendix:results:uncertainty}

Model uncertainty is another core aspect in models, providing additional information about the model's decision behavior during prediction. We investigate how uncertainty changes during changing $\omega$. Following previous studies~\cite{lesota:modern,xu2020understanding}, we measure model/prediction uncertainty as the entropy of the predicted probability distribution, namely:
\begin{equation}
    \text{Uncertainty}(X) = - \sum_{j=1}^{\left|X\right|}p(X_j)\log{p(X_j)}
    \label{eq:entropy}
\end{equation}
where $X$ is the predicted probability distribution of a single data point provided by a model, and defined over the categorical space of $\left|X\right|$ classes. For each state of the \modelours models, we calculate this measure of uncertainty over the task's predictions for each data point in the respective test set and average over the results. 

As depicted in Figure~\ref{fig:appendix:results:uncertainty}, the overall uncertainty of the BERT-Base model constantly changes (increases). We observe the same pattern in the uncertainty values when calculated on sub-populations of each dataset. Looking at the result of BIOS, we can see that the models behave more deterministic (less uncertainty) when it comes to Males and this uncertainty increases as we increase the sensitivity parameter. As for the FCDL18, we can see that for \emph{African American} sub-population the model has much lower uncertainty at $\omega=0$ compare to $\omega=1$ where the uncertainty of \emph{African American} is much closer to White. On the other hand in PAN16, we observe that model uncertainty is lower for the Female sub-population and at the beginning, the uncertainty for gender decreases then increases for higher values of $\omega$. As for the age, we observe the same pattern as we increased the $\omega$ value of age but the $\omega_{gender} = 0$

\subsection{Prediction Flips}
\label{sec:appendix:result:flips}

\begin{figure*}[t]
     \centering
     \begin{subfigure}[b]{0.33\textwidth}
         \centering
         \includegraphics[width=\columnwidth]{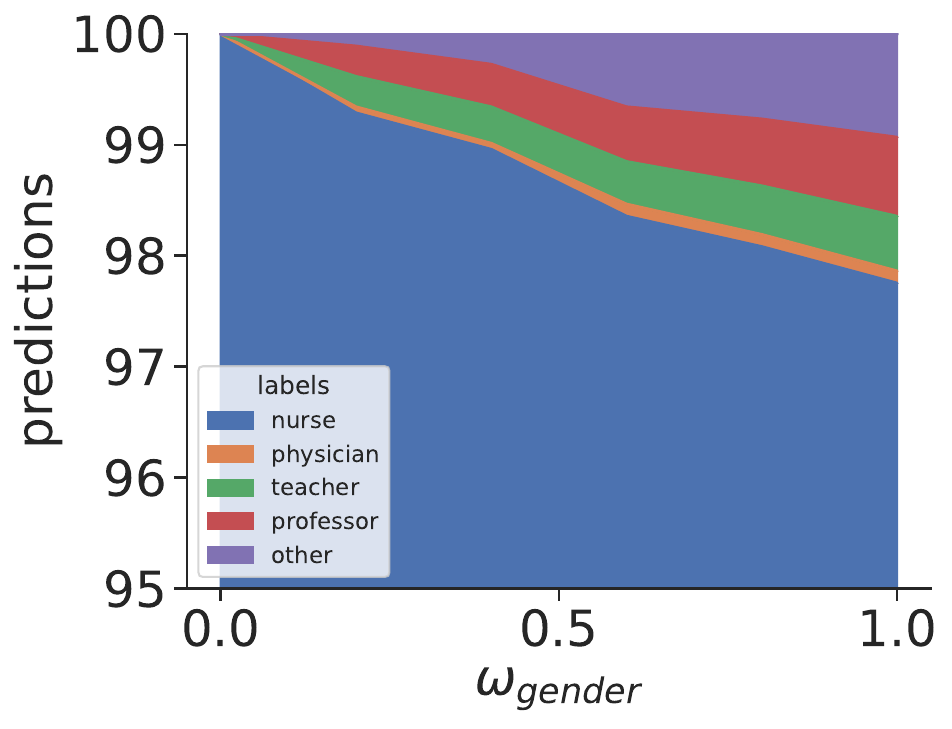}
         \caption{\textit{Nurse} in BIOS}
         \label{fig:results:labelflips:bios}
     \end{subfigure}
     \hfill
     \begin{subfigure}[b]{0.33\textwidth}
         \centering
         \includegraphics[width=\columnwidth]{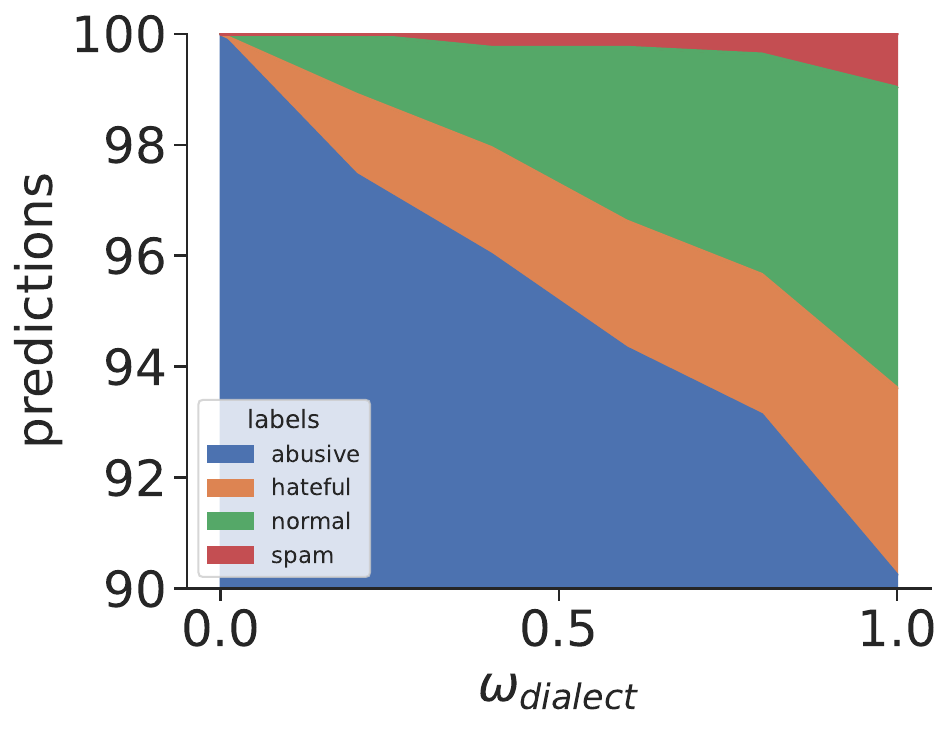}
         \caption{\textit{Abusive} in FCDL18}
         \label{fig:results:labelflips:fcdl}
     \end{subfigure}
     \caption{Percentage of the data points with the predicted label in the original ($\omega = 0$) model that remain on the prediction, when increasing $\omega$.
     }
    \label{fig:results:labelflips}
\end{figure*}


Despite the fact that average classification accuracy  -- as shown in the previous experiments -- only marginally fluctuates during (binary or continuous) concept erasure, we observe that the behavior of the model in terms of the predicted probabilities considerably changes among data points. In some cases, the changes in predicted probabilities become so pronounced that, changing a concept's degree of presence completely alters the model's decision (predicted class). In what follows, we investigate the effect of partial concept erasure on the prediction behavior of the model as we change the gating sensitivity parameter $\omega$.

\begin{figure*}[t]
\begin{subfigure}[t]{0.33\textwidth}
    \includegraphics[width=\columnwidth]{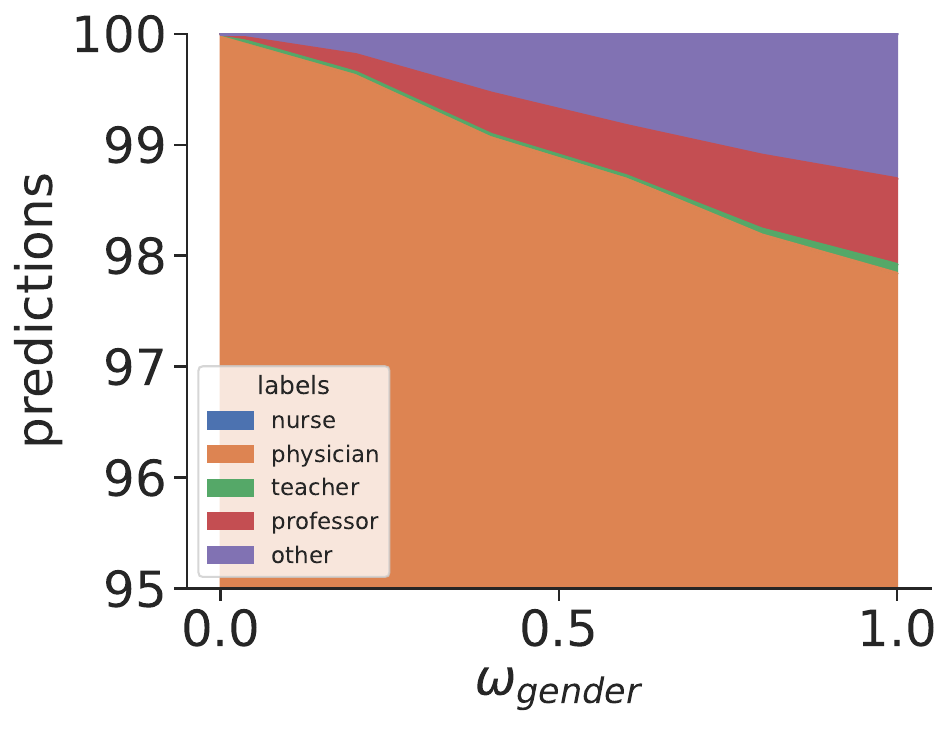}
    \caption{Physician}
\end{subfigure}
\begin{subfigure}[t]{0.33\textwidth}
    \includegraphics[width=\columnwidth]{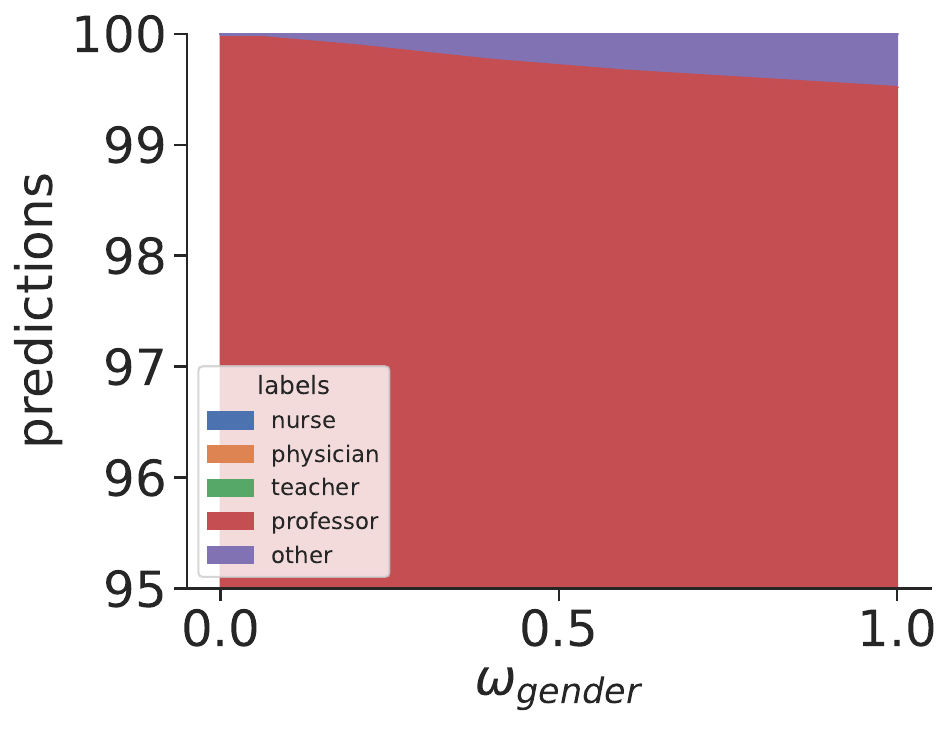}
    \caption{Professor}
\end{subfigure}
\begin{subfigure}[t]{0.33\textwidth}
    \includegraphics[width=\columnwidth]{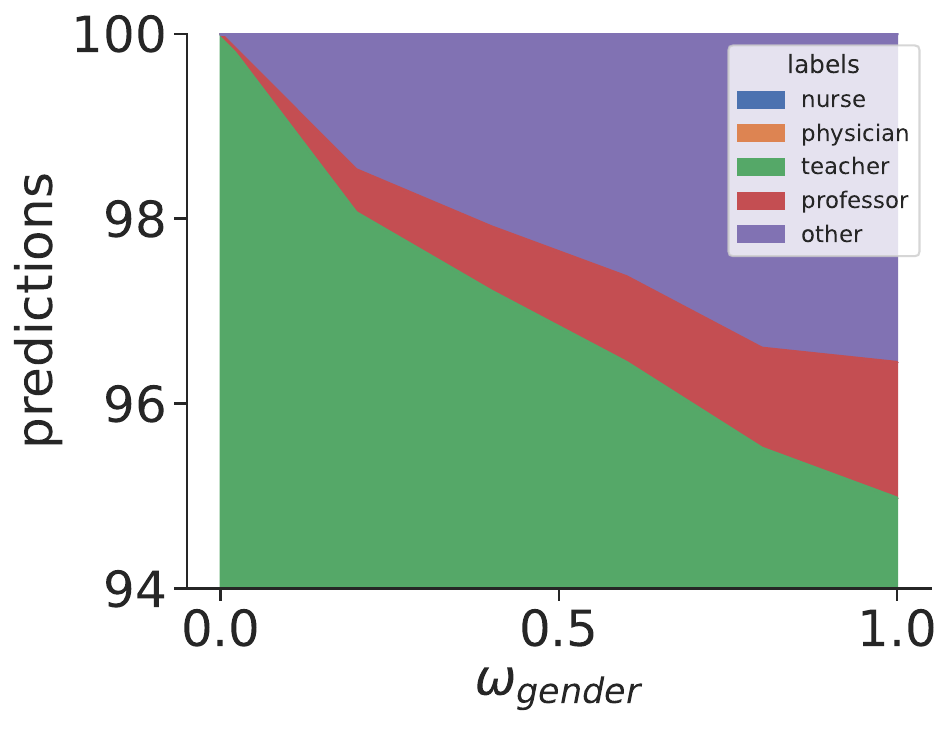}
    \caption{Teacher}
\end{subfigure}
\caption{Percentage of the predicted labels among the data points that are initially predicted as the mentioned label by the initial model during the partial concept removal of the BIOS dataset using \modelours.}
\label{fig:appendix:flips:bios}
\end{figure*}

\begin{figure*}[t]
\begin{subfigure}[t]{0.33\textwidth}
  \centering
  \includegraphics[width=\columnwidth]{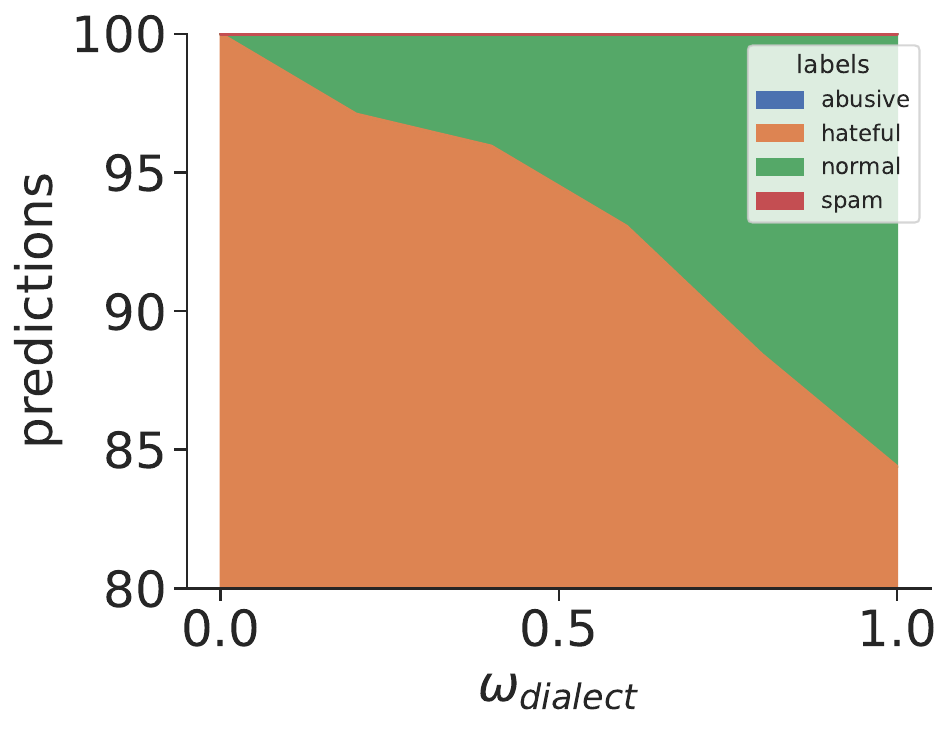}
  \caption{Hateful}
\end{subfigure}
\begin{subfigure}[t]{0.33\textwidth}
    \includegraphics[width=\columnwidth]{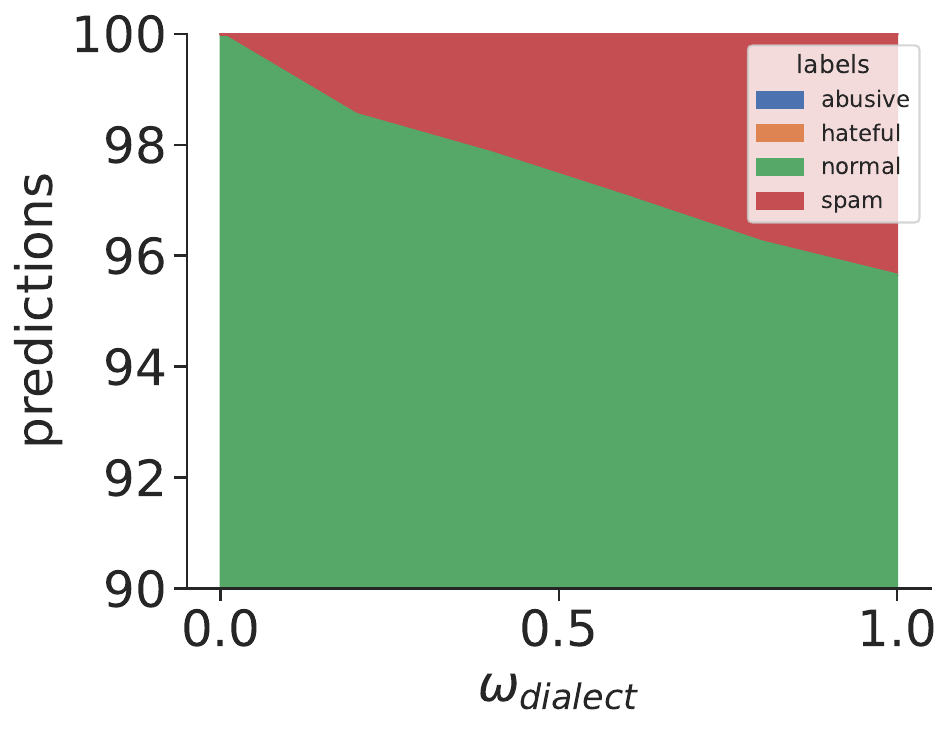}
    \caption{Normal}
\end{subfigure}
\begin{subfigure}[t]{0.33\textwidth}
    \includegraphics[width=\columnwidth]{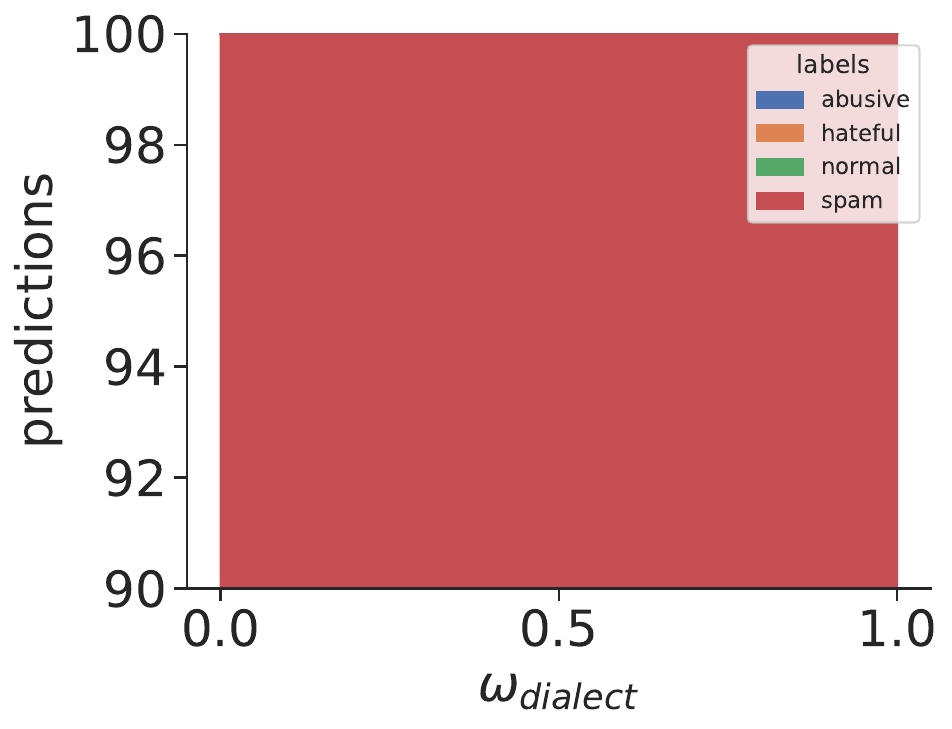}
    \caption{Spam}
\end{subfigure}
\caption{Percentage of the predicted labels among the data points that are initially predicted as the mentioned label by the initial model during the partial concept removal of the FDCL18 dataset using \modelours.}
\label{fig:appendix:flips:hatespeech}
\end{figure*}

\begin{figure*}[t]
\begin{subfigure}[t]{0.33\textwidth}
  \centering
  \includegraphics[width=\columnwidth]{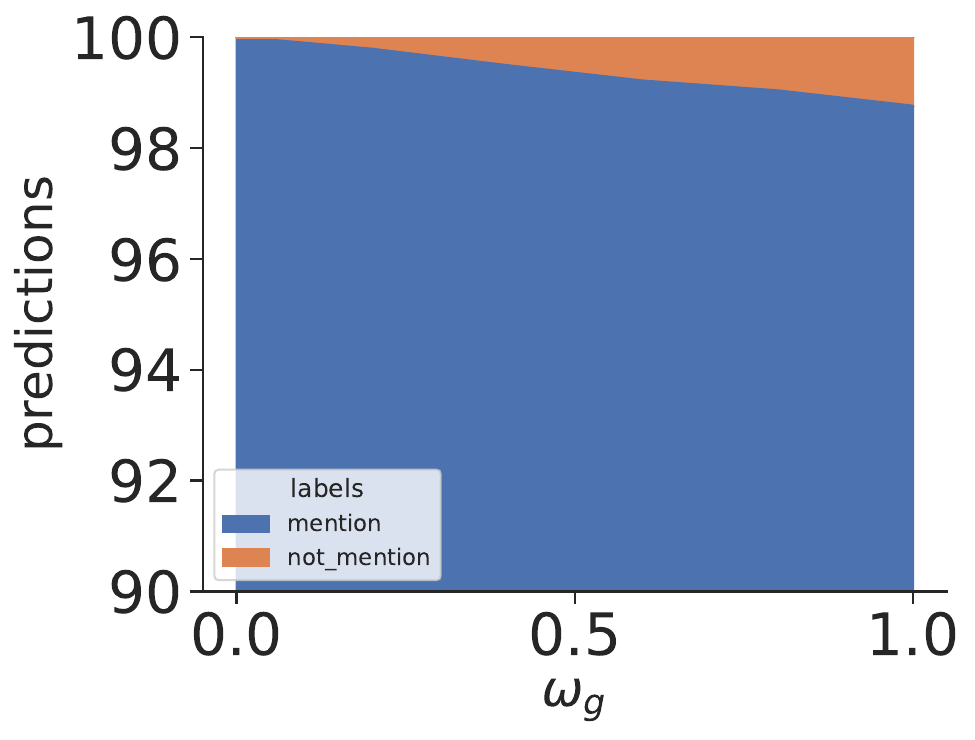}
  \caption{Gender}
\end{subfigure}
\hfill
\begin{subfigure}[t]{0.33\textwidth}
    \includegraphics[width=\columnwidth]{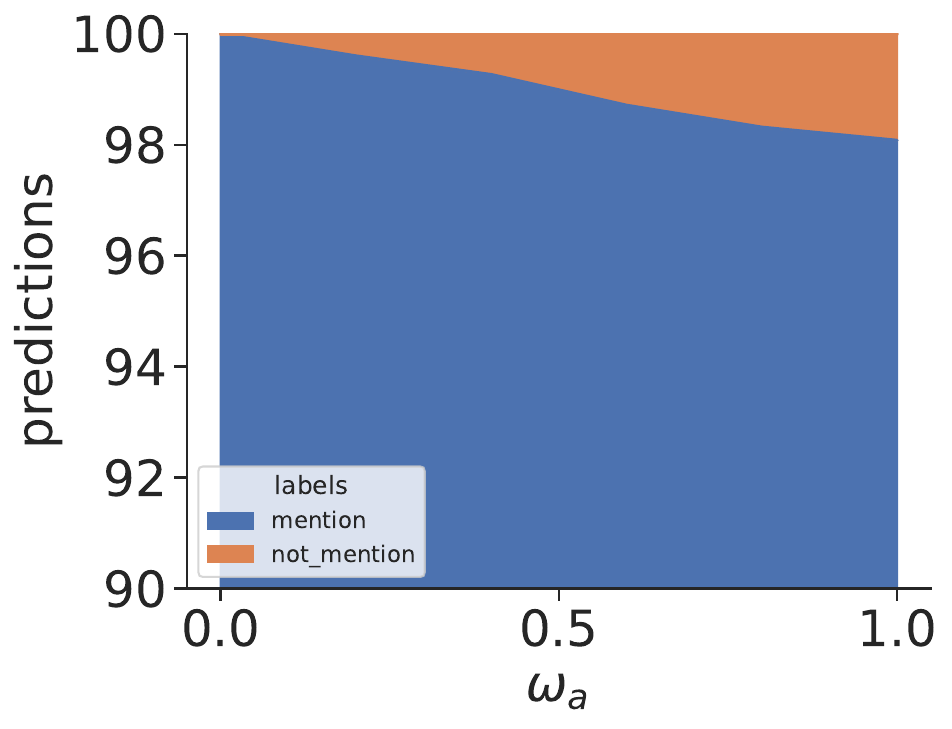}
    \caption{Age}
\end{subfigure}
\caption{Changes in the Mention prediction as we change  $\omega$ of the gender or age concept in the respective model.}
\label{fig:appendix:flips:pan16}
\end{figure*}

Figure~\ref{fig:results:labelflips} shows the statistics of the changes in the predicted classes by the \modelours model when increasing the corresponding $\omega$ parameter(s). Figure~\ref{fig:results:labelflips:bios} tracks the percentage of the predicted labels for the data points in the BIOS test set, predicted as \textit{Nurse} by the base model ($\omega=0$). Figure~\ref{fig:results:labelflips:fcdl} reports the same on FCDL18 for the \textit{Abusive} predictions. Figures~\ref{fig:appendix:flips:bios}, \ref{fig:appendix:flips:hatespeech} \ref{fig:appendix:flips:pan16} depict the results for more labels on these datasets, as well as the ones for PAN16. As shown, the changes in the predicted labels continuously increase as we increase the $\omega$ value, indicating that the decision making of the model continuously changes and more predicted labels from the initial state flip. This continuous change is consistent with concept removal results across the three datasets, demonstrating the capability of \modelours in gradually changing its predictions when moving from the initial state to the target state. To gain a more fine-grained view of this topic, we further investigate the changes in model uncertainty in Appendix~\ref{sec:appendix:results:uncertainty}.

\begin{figure*}[t]
\center
\begin{subfigure}[t]{0.4\textwidth}
    \includegraphics[width=\columnwidth]{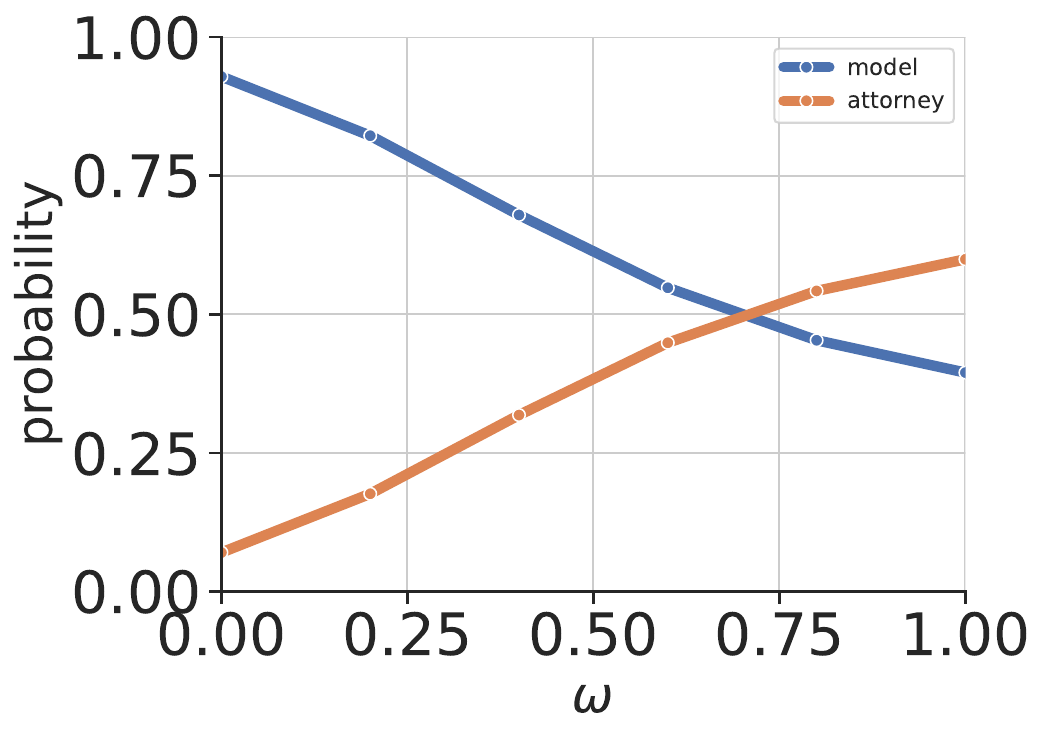}
    \caption{Male attorney}
\end{subfigure}
\hfill
\begin{subfigure}[t]{0.4\textwidth}
    \includegraphics[width=\columnwidth]{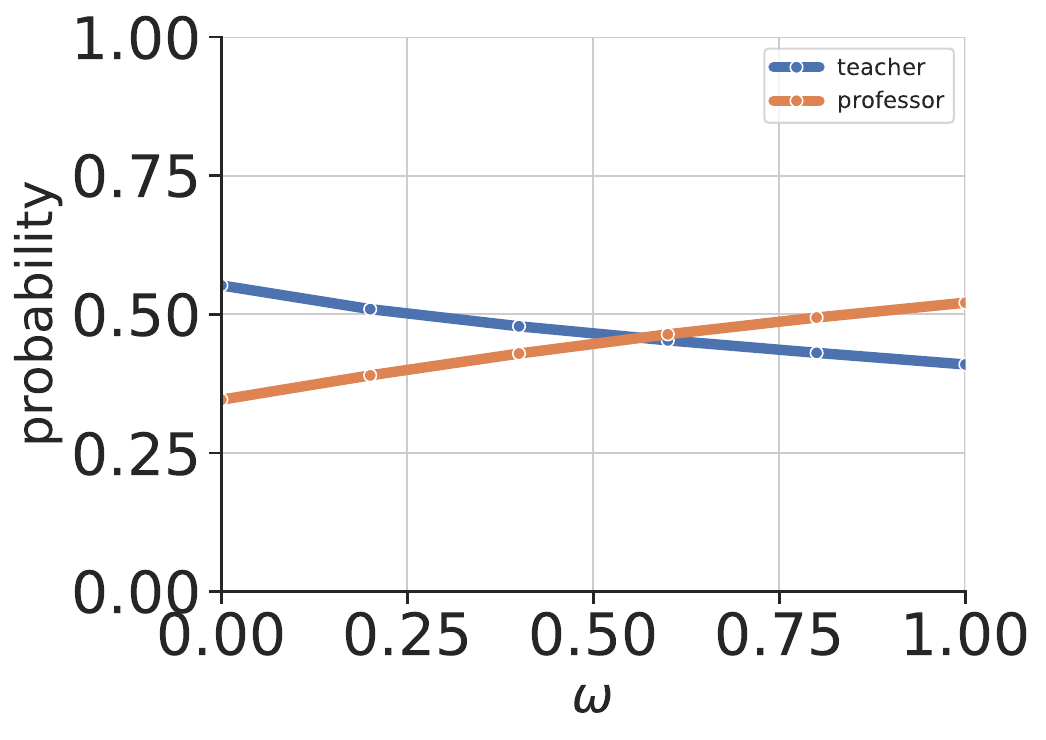}
    \caption{Female professor}
\end{subfigure}
\hfill
\caption{Figures show the \textbf{positive} effect of \modelours on the prediction probability for the labels as we increase $\omega$ value and remove information about gender in BIOS dataset (a) A male attorney predicted wrongly with the initial model ($\omega=0$) and switches to the attorney as we increase omega value. \textbf{Bio}: [he] has been stated previously senator conrad has complete confidence in s ability to enforce the law and serve the people of north dakota conrad wrote (b) A female professor predicted teacher with initial model and switches to correct label as we remove information about gender. \textbf{Bio}: [she] has twelve years of teaching experience for the undergraduate students and postgraduate supervised many undergraduate projects and several master theses in aastmt and other universities also supervised several phd students in cooperation with ein shams university egypt}
\label{fig:appendix:sample:positive:1}
\end{figure*}

\begin{figure*}[t]
\begin{subfigure}[t]{0.4\textwidth}
    \includegraphics[width=\columnwidth]{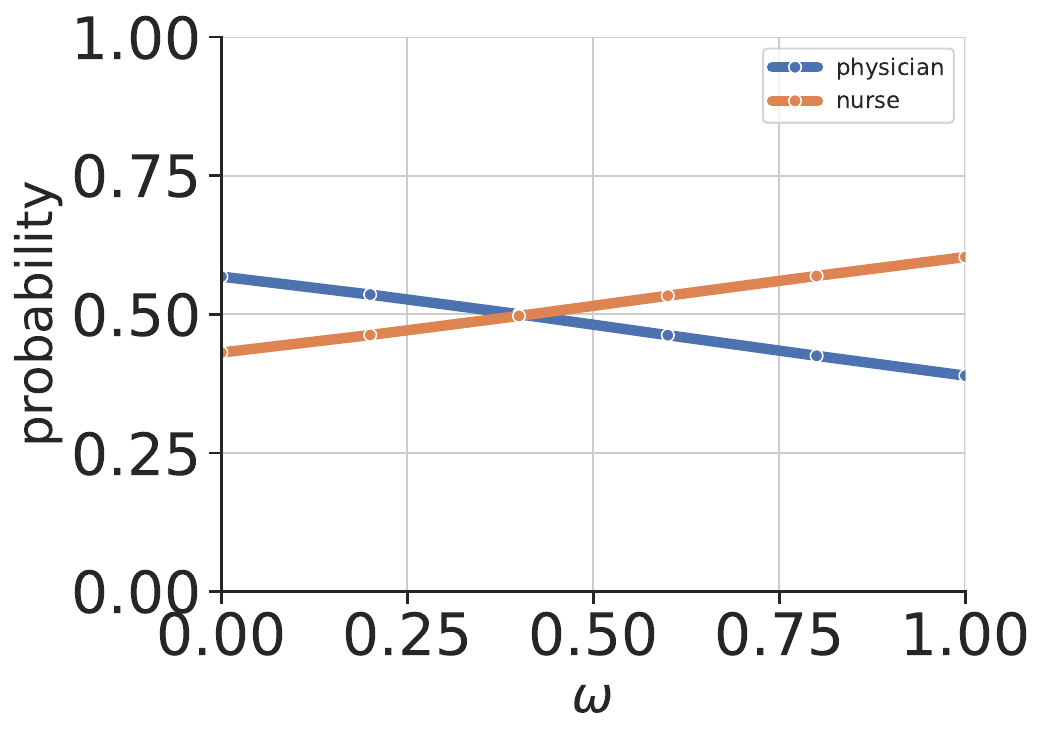}
    \caption{Female Physician}
\end{subfigure}
\hfill
\begin{subfigure}[t]{0.4\textwidth}
    \includegraphics[width=\columnwidth]{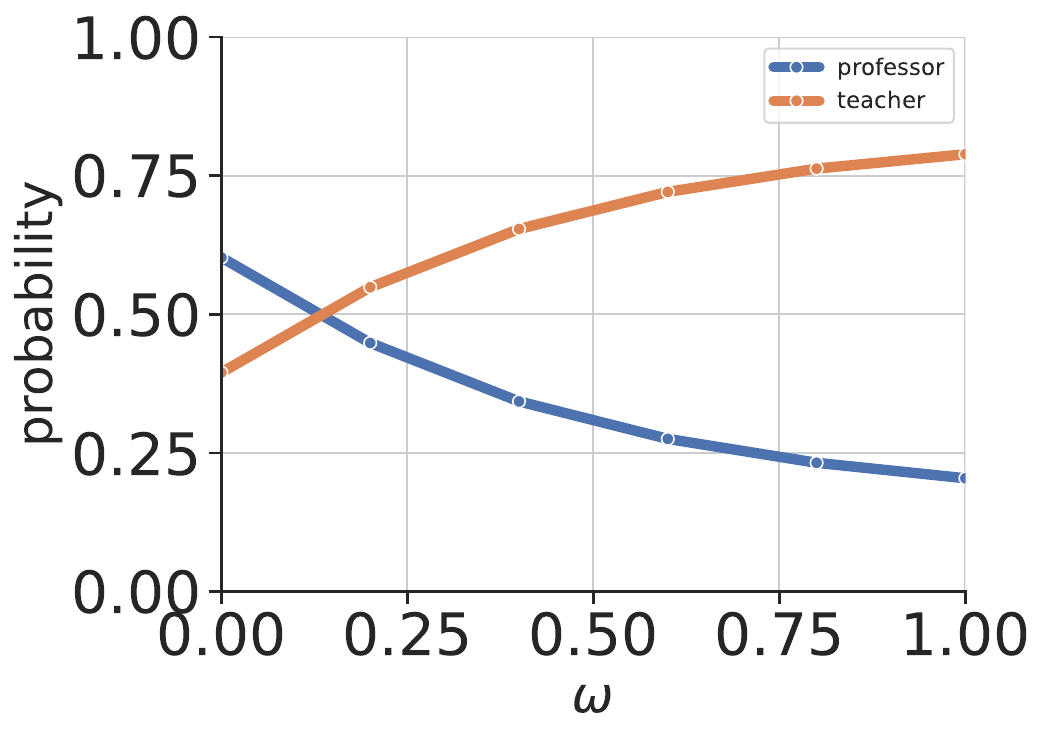}
    \caption{Male Professor}
\end{subfigure}
\hfill
\caption{Figures show the \textbf{negative} effect of \modelours on the prediction probability for the labels as we increase $\omega$ value and remove information about gender in BIOS dataset (a) A female physician predicted correctly with initial model switches to nurse as we increase omega value. \textbf{Bio}: [she] has worked in both hospital and outpatient clinical settings has experience in internal medicine preventative medicine and urgent care specializes in botox treatments that help people to look their very best without invasive cosmetic surgery (b) A male professor predicted correctly by the initial model misclassified as we increase $\omega$ value. \textbf{Bio}: has been an active member of nacta since serves on the nacta journal editorial review board and has reviewed numerous conference oral and poster presentation abstracts has submitted teaching tips and received the bob gough outstanding teaching tip award and the nacta educator award has made both oral and poster presentations at nacta conferences}
\label{fig:appendix:sample:positive:2}
\end{figure*}

\begin{figure*}[t]
\begin{subfigure}[t]{0.4\textwidth}
    \includegraphics[width=\columnwidth]{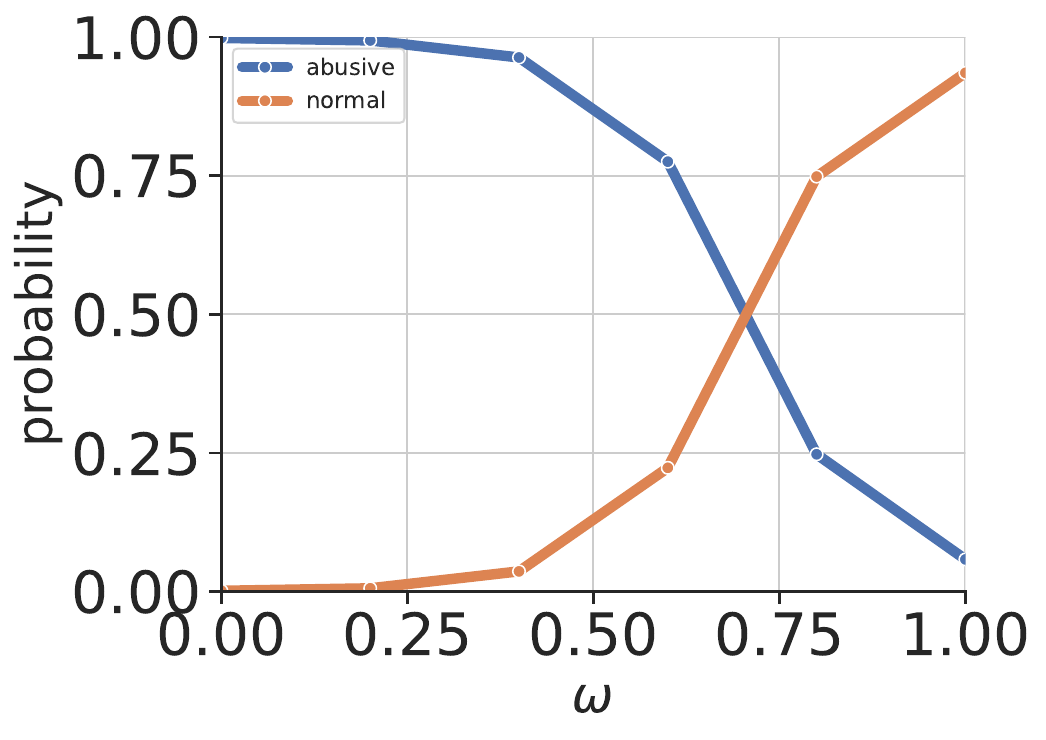}
    \caption{African American Normal tweet}
\end{subfigure}
\hfill
\begin{subfigure}[t]{0.4\textwidth}
    \includegraphics[width=\columnwidth]{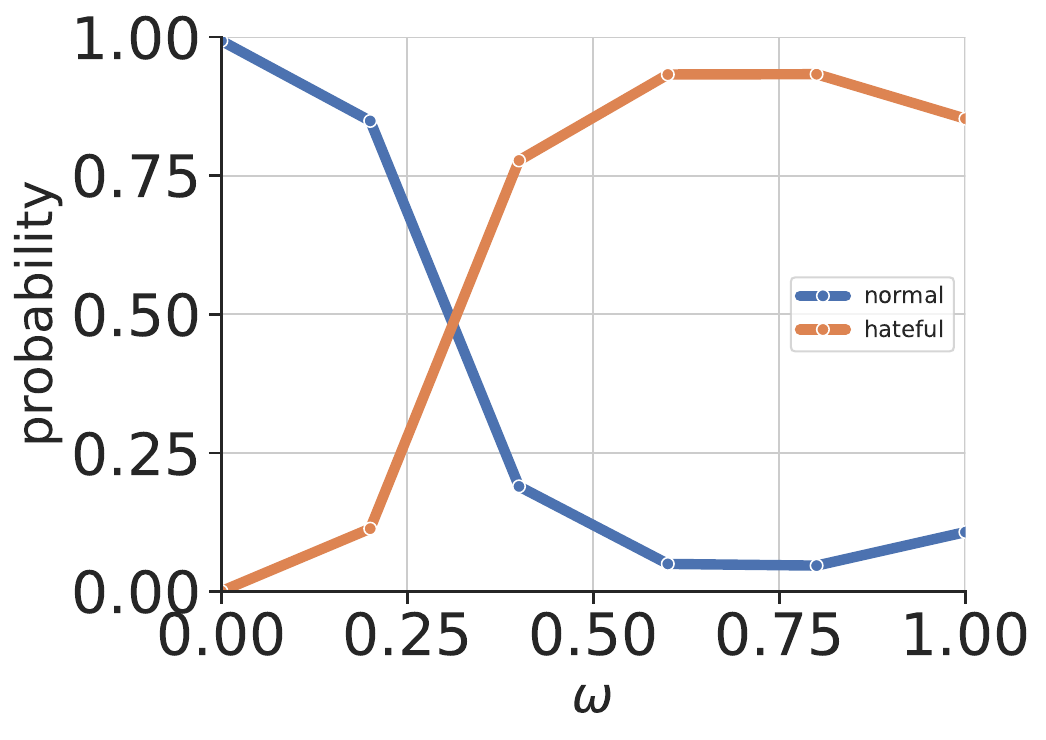}
    \caption{White person Hateful tweet}
\end{subfigure}
\hfill
\caption{Figures show \textbf{positive} effect of \modelours on the prediction probability for the labels as we increase $\omega$ value and remove information about gender in FCDL18 dataset (a) African American tweet labeled abusive at by the initial model, as we increase $\omega$ and remove dialect information labels flips to normal. \textbf{Tweet}: why is mother nature mad? somebody must have pissed her off! its rainingg hard af! (b) White person hateful tweet predicted normal by the initial model switches to the correct label as we increase $\omega$ value. \textbf{Tweet}: us is already doing something by backing some of the rebel groups. this is a proxy war afte…}
\label{fig:appendix:sample:positive:3}
\end{figure*}

\begin{figure*}[t]
\begin{subfigure}[t]{0.4\textwidth}
    \includegraphics[width=\columnwidth]{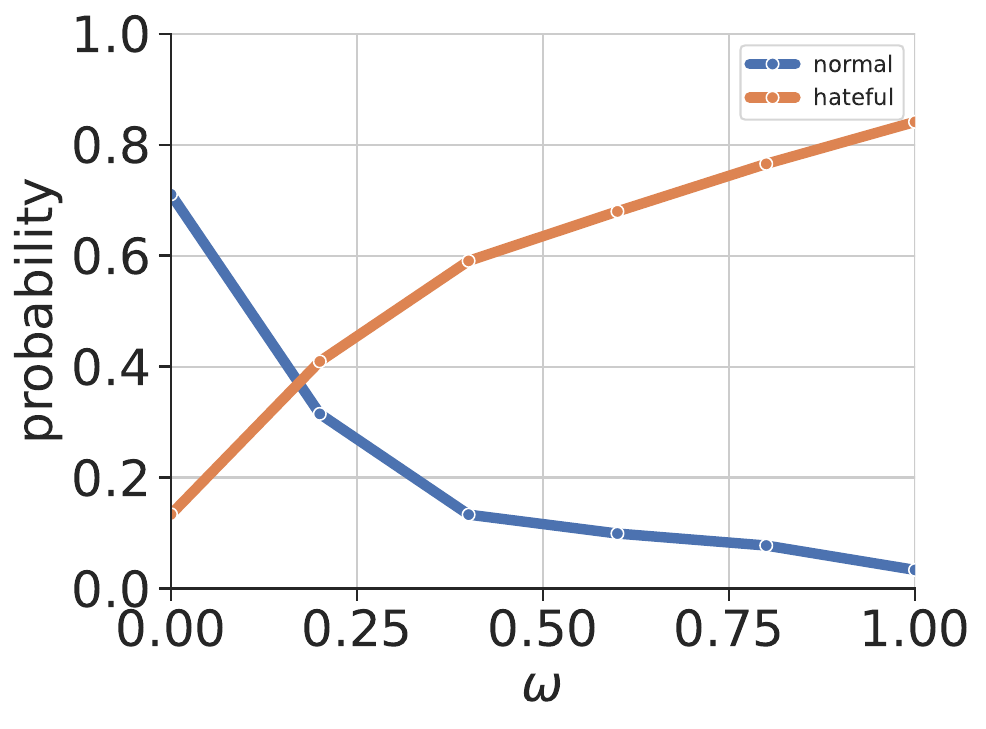}
    \caption{White Normal tweet}
\end{subfigure}
\hfill
\begin{subfigure}[t]{0.4\textwidth}
    \includegraphics[width=\columnwidth]{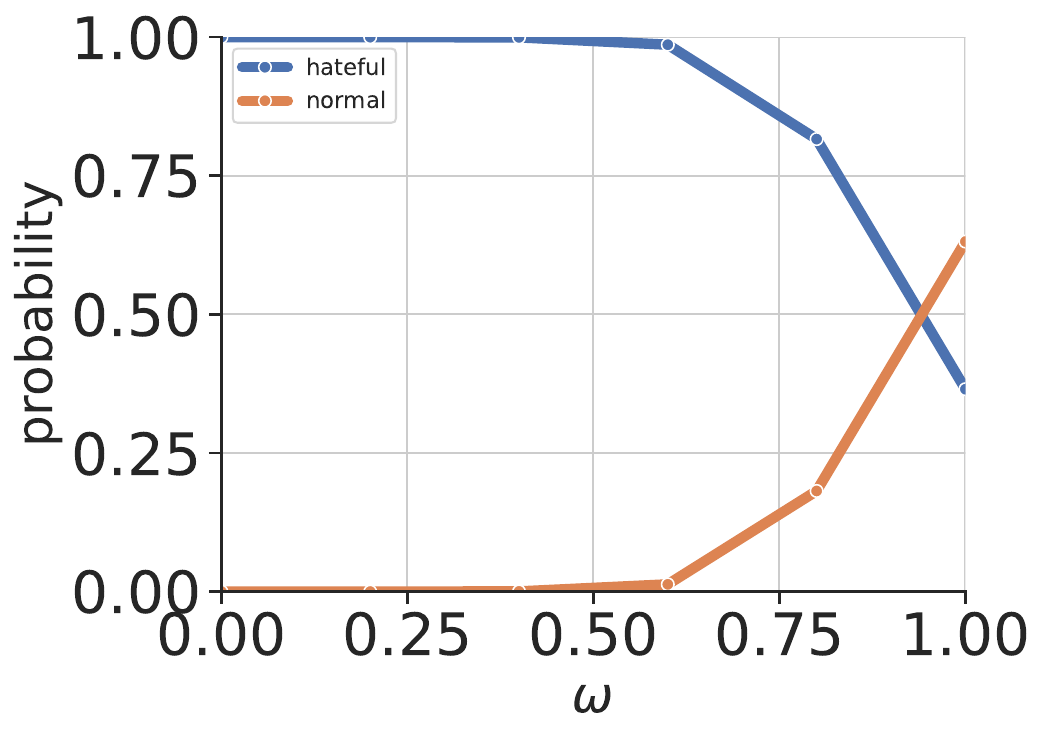}
    \caption{African American hateful tweet}
\end{subfigure}
\hfill
\caption{Figures show \textbf{negative} effect of \modelours on the prediction probability for the labels as we increase $\omega$ value and remove information about gender in the FCDL18 dataset (a) White normal tweet is correctly labels by the initial model. By increasing $\omega$ value and removing dialect information model predicts the tweet as hateful. \textbf{Tweet}: yes because the person we voted for is keeping his promises, in spite of the lefts resistance! maga. today and (b) African American hateful speech negatively switches to normal as we remove dialect information from the model. \textbf{Tweet}: y'all be wanting gifts from y'all [curse word]  when y'all mad just get me food ;  }
\label{fig:appendix:sample:positive:4}
\end{figure*}

\begin{figure*}[t]
\begin{subfigure}[t]{0.4\textwidth}
    \includegraphics[width=\columnwidth]{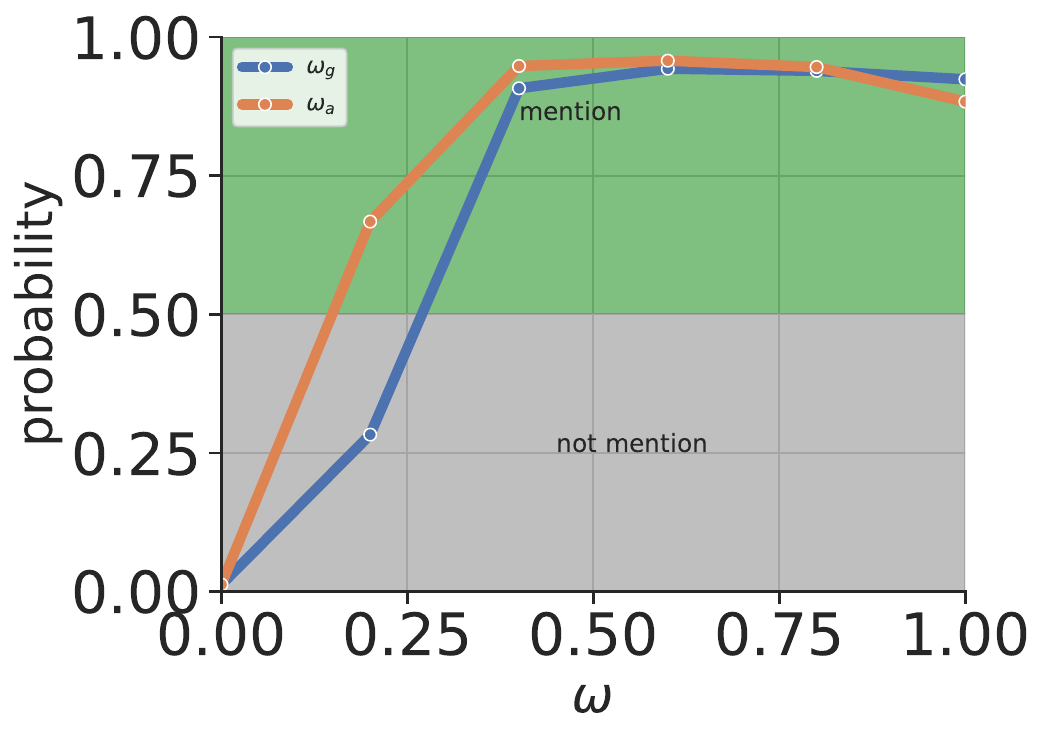}
    \caption{Female Mentioning someone }
\end{subfigure}
\hfill
\begin{subfigure}[t]{0.4\textwidth}
    \includegraphics[width=\columnwidth]{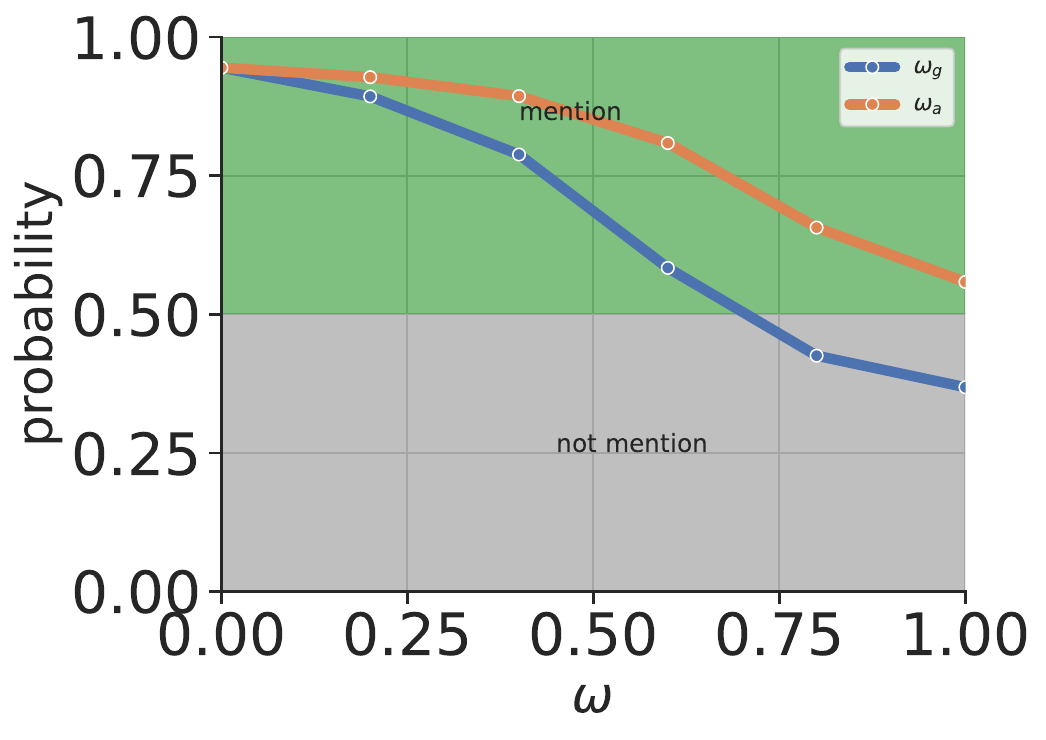}
    \caption{Male mentioning someone}
\end{subfigure}
\hfill
\caption{Figures show the prediction probability for the labels as we increase $\omega$ value and remove information about gender and age separately in PAN16 dataset. (a) Female Mentioning someone in the tweet but tagged as-not mentioned by the initial model \textbf{positively} switches to mention as we remove gender and age information separately. Removing any of the attributes alone results in the correct class of the prediction of this sample. \textbf{Tweet}: happy charlie. stolen from \#cats \#tuxedokitty \#blackandwhite (b) Male mentioning someone was correctly labeled by the initial model with gender and age information. increasing $\omega_{gender}$ \textbf{negatively} influences model until predicts the label wrongly. Although $\omega_{age}$ also influences the prediction negatively, but we can observe that at $\omega_{age}=1$ decision of the model is still correct. \textbf{Tweet}: (snort) yes.  there it is, and  there it is}
\vspace{-10mm}
\label{fig:appendix:sample:positive:5}
\end{figure*}

\section{NFaiRR}
\label{appendix:sec:nfairr}
We use Normalized Fairness of Retrieval Results (NFaiRR) to calculate the neutrality score which is fomulated in equation~\ref{eq:fairness}. In equation~\ref{eq:fairness} FaiRR is the fairness metric similar to other studies~\cite{Kulshrestha:fairness,fabris:fairness} is calculated by finding the importance of attribute in relation to its retrieved position. Also Ideal FaiRR is used to normalized the score which is considered as the best possible fairness result one can get by reordering the documents. For more details of the score please refer to~\cite{rekabsaz_2021_societal}. 

\begin{equation}
    NFaiRR(L,\hat{s}) = \sum_{q \in Q}\frac{FaiRR_q(L)}{IFaiRR_q(\hat{s})}
    \label{eq:fairness}
\end{equation}

\end{document}